\documentclass[12pt]{article}
\usepackage{amsmath,amssymb,amsthm,amsfonts,bm,mathrsfs,float,algorithm,algorithmic,caption,subcaption,setspace}
\usepackage{avant,array,geometry,fontenc,inputenc,natbib,hyperref,multirow,enumerate,rotating,booktabs,color,latexsym,times,stmaryrd,xr}
\usepackage{url}  
\newtheorem{lemma}{Lemma}
\newtheorem{theorem}{Theorem}

\newtheorem{prop}{Proposition}
\newtheorem{example}{Example}

\newcommand{\mubf}{\mu}
\newcommand{\omegabf}{\omega}
\newcommand{\thetabf}{\theta}
\newcommand{\Omegabf}{\Omega}
\newcommand{\bbf}{b}
\newcommand{\ubf}{u}
\newcommand{\Abf}{A}
\newcommand{\Ubf}{U}
\newcommand{\Wbf}{W}
\newcommand{\Xbf}{X}
\newcommand{\Xbm}{\mathbb{X}}

\newcommand{\AC}{\hbox{AC}}
\newcommand{\DS}{\hbox{DS}}
\newcommand{\PA}{\hbox{PA}}

\newcommand{\Mean}{{\mathbb{E}}}
\newcommand{\Var}{{\mbox{Var}}}
\newcommand{\Cov}{{\mbox{cov}}}

\newcommand{\prob}{{\mathbb{P}}}
\newcommand{\tsb}[1]{\textnormal{#1}}

\let\proglang=\textsf
\DeclareMathOperator*{\argmin}{arg\,min}
\DeclareMathOperator*{\argmax}{arg\,max}

\def\floor#1{\lfloor #1 \rfloor}

\newcommand{\blind}{1}

\def\eop{\hfill {\large $\Box$}}

\begin{document}

\if1\blind
{
\title{\Large{\textbf{Testing Directed Acyclic Graph via Structural, \\ 
Supervised and Generative Adversarial Learning}}}
\author{
\bigskip
Chengchun Shi$^\dag$, Yunzhe Zhou$^\ddag$, and Lexin Li$^\ddag$ \\
\normalsize{\textit{$^\dag$London School of Economics and Political Science }} \\
\normalsize{\textit{$^\ddag$University of California at Berkeley}}
}
\date{}
\maketitle
} \fi

\if0\blind
{
\title{\Large{\textbf{Testing Directed Acyclic Graph via Structural, \\ 
Supervised and Generative Adversarial Learning}}}
\author{
\bigskip
\vspace{0.5in}
}
\date{}
\maketitle
} \fi

\baselineskip=20pt
\begin{abstract}
In this article, we propose a new hypothesis testing method for directed acyclic graph (DAG). While there is a rich class of DAG estimation methods, there is a relative paucity of DAG inference solutions. Moreover, the existing methods often impose some specific model structures such as linear models or additive models, and assume independent data observations. Our proposed test instead allows the associations among the random variables to be nonlinear and the data to be time-dependent. We build the test based on some highly flexible neural networks learners. We establish the asymptotic guarantees of the test, while allowing either the number of subjects or the number of time points for each subject to diverge to infinity. We demonstrate the efficacy of the test through simulations and a brain connectivity network analysis.    
\end{abstract}

\noindent{\bf Key Words:} 
Brain connectivity networks; Directed acyclic graph; Hypothesis testing; Generative adversarial networks; Multilayer perceptron neural networks.

\baselineskip=22pt

\section{Introduction}

Directed acyclic graph (DAG) is an important tool to characterize pairwise associations among multivariate and high-dimensional random variables. It has been frequently used in a  wide range of scientific applications. One example is gene regulatory network analysis in genetics \citep{Sachs2005}, where the time-course expression data of multiple genes are measured over multiple cellular samples through microarray or RNA sequencing, and the goal is to understand the regulatory activation or repression relations among different genes. Another example is brain effective connectivity analysis in neuroscience \citep{Garg2011}, where the time-course neural activities are measured at multiple brain regions for multiple experimental subjects through functional magnetic resonance imaging, and the goal is to infer the influences of brain regions exerting over each other under the stimulus.

There is a large body of literature studying penalized estimation of DAG given the observational data \citep[see, e.g.,][among many others]{Spirtes2000, van2013, zheng2018dags, Yuan2019}. These works all impose some specific model structures, most often, linear models or additive models. There have recently emerged a number of proposals in the computer science  literature that used neural networks or reinforcement learning to tackle nonlinear models and to estimate the associated DAG \citep{yu2019dag, zheng2020learning, zhu2020causal}. While all these works have made crucial contributions, DAG model \emph{estimation} is an utterly different problem from DAG  {inference}. By inference, we mean hypothesis testing of individual edges throughout this article. The two problems are closely related, and both can, in effect, identify important links of a DAG. Besides, DAG inference usually relies on DAG estimation as a precedent step. Nevertheless, estimation does not produce an explicit quantification of statistical significance as inference does. Bayesian networks have been proposed for DAG estimation and inference. However, computationally, it is extremely difficult to search through all possible graph structures in a Bayesian network \citep{Chickering2004}, and as a result, the dimension of the Bayesian network is often small \citep{Friston2011}. There are very few frequentist inference solutions for inferring DAG structures. Only recently, \citet{Jana2019Inf} proposed a de-biased estimator to construct confidence intervals for the edge weights in a DAG, whereas \citet{li2019likelihood} developed a constrained likelihood ratio test to infer individual edges or some given directed paths of a DAG. These works are probably the most relevant to our proposal. However, both have focused on Gaussian linear DAG, and cannot be easily extended to more general nonlinear DAG models. Moreover, all the above works considered the setting where the data observations are independent and identically distributed (i.i.d.). Learning DAG from time-dependent data remains largely unexplored.

There is another body of literature studying conditional independence testing (CIT); see \citet{li2019nonparametric,shah2018hardness,shi2021double} and the references therein. CIT is closely related to DAG inference, and is to serve as a building block of our proposed testing procedure. On the other hand, naively performing CIT on two variables given the rest would fail to infer the directed edges of a DAG; see Section \ref{sec:equiv} for details. Besides, most CIT methods assume the data observations are independent, and are not suitable for the setting where the measurements are time-dependent.

In this article, we propose a novel statistical testing procedure for the inference of individual links or some given paths in a large and general DAG. The new test hinges upon some highly flexible neural networks-based machine learning techniques. The associations among the random variables can be either linear or nonlinear, the variables themselves can be either continuous or discrete-valued, and the observed data can be time-dependent. 

Methodologically, we employ a number of state-of-the-art deep learning techniques that are highly flexible and can capture nonlinear associations among high-dimensional variables. We begin with a new characterization of directed edges under the additive noise structure \citep{Peters2014causal}; see Theorem \ref{thm1}. Based on this characterization, we propose a new testing procedure that integrates three key deep learning ingredients: (a) a DAG structural learning method based on neural networks or reinforcement learning to estimate the DAG; (b) a supervised learning method based on neural networks to estimate the conditional mean; and (c) a distribution generator produced by generative adversarial networks \citep[GANs]{goodfellow2014generative} to approximate the conditional distribution of the variables in the DAG. We further couple these deep learning tools with some hypothesis testing strategies, including data splitting and cross-fitting to ensure a valid size control, and constructing a doubly robust test statistic as the maximum of multiple transformation functions to improve the power. 

Theoretically, we establish the asymptotic size and power guarantees for the proposed test. The data-splitting and cross-fitting strategy ensures that our test achieves a valid type-I error control asymptotically under minimal conditions on those learning methods. As a result, our test procedure can work with a wide range of nonparametric estimators. Next, our DAG testing procedure requires a DAG estimation solution as a precedent step, which is common for almost all graph inference approaches \citep{Cai2017review}. However, we do not assume the ordering of the nodes is known a priori, but instead estimate this DAG ordering from the data using some DAG structural learning method. To establish the consistency of the proposed test, we require this ordering is consistently estimated; see condition (C1). Nevertheless, this order consistency is much weaker than requiring the initial DAG estimator to be selection consistent, or to satisfy the sure screening property. In other words, we only require a reasonably good initial estimator of DAG, which is order consistent but not necessarily selection consistent. We then develop a testing procedure that produces an explicit quantification of statistical significance for each individual link, and we show the test has the desired size and power guarantees. We also prove that the estimator from the DAG structural learning method we employ is indeed order consistent. Meanwhile, we discuss the impact on our test when this order consistency condition is not satisfied. Finally, for our theoretical analysis, we introduce a bidirectional asymptotic framework that allows either the number of subjects, or the number of time points for each subject, to diverge to infinity. This is useful for different types of applications. There are plenty of studies where the interest is about the general population, and thus it is reasonable to let the number of subjects or samples to diverge. Meanwhile, there are plenty of other applications, e.g., neuroimaging-based brain networks studies, where the number of subjects is almost always limited, but the scanning time and the temporal resolution can greatly increase. For those applications, it is more suitable to let the number of time points to diverge.

Our proposal is innovative and makes useful contributions in several ways. 

First, rigorous inference of directed edges in DAG is a vital but also a long-standing open question. The existing solutions rely on particular model structures such as linear or additive models, and mostly deal with i.i.d.\ data. Such requirements can be restrictive in numerous applications, since the actual relations may be nonlinear and the data are correlated. By contrast, we only require an additive noise structure. To the best of our knowledge, our work is the first frequentist hypothesis testing solution for a general DAG with time-dependent data.

Second, we employ modern deep learning techniques such as neural networks and GANs to help address a classical statistical hypothesis testing problem. Such modern learning methods serve as nonparametric learners, and conceptually, play a similar role as splines and reproducing kernels. Meanwhile, they are often more flexible and can handle more complex data structures. With increasingly efficient implementations of these methods and improved understandings of their theoretical properties \citep[e.g.,][]{Bauer2019, farrell2018deep}, this family of deep learning methods offer a powerful set of tools for classical statistical problems. Our proposal can be viewed as one of the early examples of harnessing such power, as the use of these deep learning techniques allows us to accurately estimate the DAG structure, the conditional means, as well as the distribution functions, and to improve the power of the test. 

Third, even though the individual learning components such as neural networks, GANs and cross-fitting are not completely new, how to integrate them properly and effectively into a test with desired theoretical guarantees is highly nontrivial, and is one of the main contributions of this article. In effect, our proposed test achieves a parametric convergence rate and a parametric power guarantee while using nonparametric estimators. This is made possible mainly due to the innovative way we put together these learning components, which leads to a doubly robust test statistic \citep{tsiatis2007semiparametric}, in the sense that the proposed statistic is consistent, as long as either the conditional mean function in (b), or the distribution generator in (c) is correctly specified. In our solution, we propose to estimate both the conditional mean and the distribution generator fully nonparametrically. As such, the convergence rate of the two estimators, denoted by $\kappa_1$ and $\kappa_2$, respectively, may each be slower than the parametric rate. Nevertheless, we only require $\kappa_1 + \kappa_2 > 1/2$, which is totally achievable for the multilayer perceptron models and GANs; see the discussion after condition (C4). The key idea of our theoretically analysis is to show the bias of the estimating equation grows faster than the parametric rate. Thanks to the double robustness property of the test statistic, if we replace either estimator with its oracle value, the bias would be equal to zero. This observation, together with the Neyman orthogonality property of the estimating equation, ensures that the bias can be represented as a product of the difference between the two nonparametric estimators and their oracle values. Consequently, when $\kappa_1 + \kappa_2 > 1/2$, the test statistic converges at a parametric rate, the corresponding test controls the type-I error, and has a parametric power guarantee. We comment that, in their seminal work on double/debiased machine learning, \citet{chernozhukov2018double} proposed to combine two machine learning estimators to infer the average treatment effect, which they showed to achieve a parametric convergence rate, even though each of the machine learning estimator converges at a nonparametric rate. Our result is similar in spirit as theirs, but targets  a completely different problem, and thus is the first of its kind for DAG inference.  

The rest of the article is organized as follows. We formally define the hypotheses, along with the model and data structure, in Section \ref{sec:formulation}. We develop the testing procedure in Section \ref{sec:test}, and establish the theoretical properties in Section \ref{sec:theory}. We study the empirical performance of the test through simulations and a real data example in Sections \ref{sec:sim} and \ref{sec:realdata}. We relegate several extensions, additional results, and all technical proofs to the Supplementary Appendix.

\section{Problem Formulation}
\label{sec:formulation}

In this section, we first present the DAG model, based on which we formally define our hypotheses. We next propose an equivalent  characterization of the hypotheses, for which we develop our testing procedure. Finally, we detail the data structure.

\subsection{DAG model}
\label{sec:model}

Consider $d$ random variables $\Xbf = (X_1, \ldots, X_d)^\top$, each with a finite fourth moment. We use a directed graph to characterize the relationships among these variables, where a node of the graph corresponds to a variable in $\Xbf$. For two nodes $i,j\in \{1,\ldots,d\}$, if an arrow is drawn from $i$ to $j$, i.e., $i\rightarrow j$, then $X_i$ is called a parent of $X_j$, and $X_j$ a child of $X_i$. A directed path in the graph is a sequence of distinct nodes $i_1, \ldots, i_{d'}$, such that there is a directed edge $i_k \to i_{k+1}$ for all $k=1,\ldots,d'-1$. If there exists a directed path from $i$ to $j$, then $X_i$ is called an ancestor of $X_j$, and $X_j$ a descendant of $X_i$. For node $X_j$, let $\PA_j, \DS_j$ and $\AC_j$ denote the set of indices of the parents, descendants, and ancestors of $X_j$, respectively. Moreover, let $\Xbf_{\mathcal{M}}$ denote the sub-vector of $\Xbf$ formed by those whose indices are in a subset $\mathcal{M} \subseteq \{1, \ldots, d\}$.

To rigorously formulate our problem, we make two assumptions.
\vspace{-0.1in}
\begin{enumerate}[({A}1)]
\setlength\itemsep{-0.25em}
\item The directed graph is acyclic; i.e., no variable is an ancestor of itself.
\item The DAG is identifiable from the joint distribution of $\Xbf$.
\vspace{-0.1in}
\end{enumerate}

\noindent
Condition (A1) has been commonly imposed in directed graph analysis. It does not permit any variable to be its own ancestor. As a result, the relationship between any two variables is unidirectional. Condition (A2) helps simplify the problem, and avoids dealing with the equivalence class of DAG. This condition is again frequently imposed in the DAG estimation literature \citep{zheng2018dags,Yuan2019,li2019likelihood,zheng2020learning}. We discuss the extension to the equivalence class in Section \ref{sec:ext-markov} of the Appendix.

We consider a class of structural equation models that follow an additive noise structure, 
\begin{eqnarray}\label{eqn:model}
X_j=f_j(\Xbf_{\scriptsize{\PA}_j})+\varepsilon_j, \quad \textrm{ for any } j=1, \ldots, d,
\end{eqnarray}
where $\{f_j\}_{j=1}^d$ are a set of continuous functions, and $\{\varepsilon_j\}_{j=1}^{d}$ are a set of independent zero mean random errors. Model \eqref{eqn:model} permits a fairly flexible structure. For instance, if each $f_j$ is a linear function, then \eqref{eqn:model} reduces to a linear structural equation model. If each $f_j$ is an additive function, i.e., $f_j(\Xbf_{\scriptsize{\hbox{PA}}_j})=\sum_{k \in {\scriptsize{\hbox{PA}}_j}} f_{j,k}(X_k)$, then \eqref{eqn:model} becomes an additive model. In our test, we do \emph{not} impose linear or additive model structures. Moreover, we can easily extend the proposed test to the setting of generalized linear model, where the $X_j$ can be either continuous or discrete-valued. We discuss such an extension in Section \ref{sec:ext-glm} of the Appendix. 

Under model \eqref{eqn:model}, the corresponding DAG is identifiable under some reasonable conditions. We consider three examples to discuss explicitly those conditions. 

\begin{example}[Gaussian graphical model]
\tsb{Suppose $X_1,\ldots, X_d$ are jointly normal, and model \eqref{eqn:model} becomes $X_j = \Wbf_j^\top \Xbf_{\tsb{\scriptsize{\PA}}_j} + b_j + \varepsilon_j$, for some $\Wbf_j$ and $b_j$. Then the corresponding DAG is identifiable, if the variance of the random error $\varepsilon_j$ is the same for all $j=1, \ldots, d$ \citep[Theorem 1]{buhlmann2014}.}
\end{example}

\begin{example}[Nonlinear graphical model with Gaussian noise] \tsb{Suppose $\varepsilon_1, \ldots, \varepsilon_d$ are jointly normal, but  $X_1,\ldots, X_d$ are not. Then the corresponding DAG is identifiable, if each $f_j$ is three times differentiable and not linear in any of its arguments \citep[Corollary 31]{Peters2014causal}.}
\end{example}

\begin{example}[Nonlinear graphical model with general noise] \tsb{Suppose neither $X_j$ nor $\varepsilon_j$ is normal. Then the corresponding DAG is identifiable, if each $f_j$ is non-constant in each of its arguments, and \eqref{eqn:model} is a restricted additive noise model \citep[Definition 27]{Peters2014causal}.}
\end{example}

\subsection{Hypotheses and equivalent characterization}
\label{sec:equiv}

We next formally define the hypotheses we target, then give an equivalent characterization. For a given pair of nodes $(j,k)$, $j,k=1, \ldots, d, j \neq k$, we aim at the hypotheses:
\begin{eqnarray}\label{eqn:hypothesis}
H_0(j,k):k \notin \PA_j, \quad \textrm{versus} \quad H_1(j,k):k \in \PA_j.
\end{eqnarray}
When the alternative hypothesis holds, there is a link from $X_k$ to $X_j$. In the following, we mainly focus on testing an individual link $H_0(j,k)$. We discuss the extension of testing a directed pathway, or a union of links, in Section \ref{sec:ext-path} and Section \ref{sec:ext-union} of the Appendix. 

We next consider a pair of hypotheses that involve two variables that are \emph{conditionally independent} (CI). The new hypotheses are closely related to \eqref{eqn:hypothesis}, but are \emph{not} exactly the same.
\begin{eqnarray}\label{eqn:hypothesis1}
\begin{split}
\hspace{1cm} & H_0^*(j,k):X_k\,\,\textrm{and}\,\, X_j\,\,\textrm{are}\,\,\textrm{CI\,\,given the rest of variables}, \,\,\, \textrm{ versus } \\
& H_1^*(j,k):X_k\,\,\textrm{and}\,\, X_j\,\,\textrm{are}\,\,\textit{\hbox{not}}\,\,\textrm{CI\,\,given the rest of variables}.
\end{split}	
\end{eqnarray}
We point out that, testing for \eqref{eqn:hypothesis1} is generally \emph{not} the same as testing for \eqref{eqn:hypothesis}. To elaborate this, we consider a three-variable DAG with a v-structure.

\begin{figure}[t!]
\centering
\begin{tabular}{ccccc}
\includegraphics[width=2.75cm]{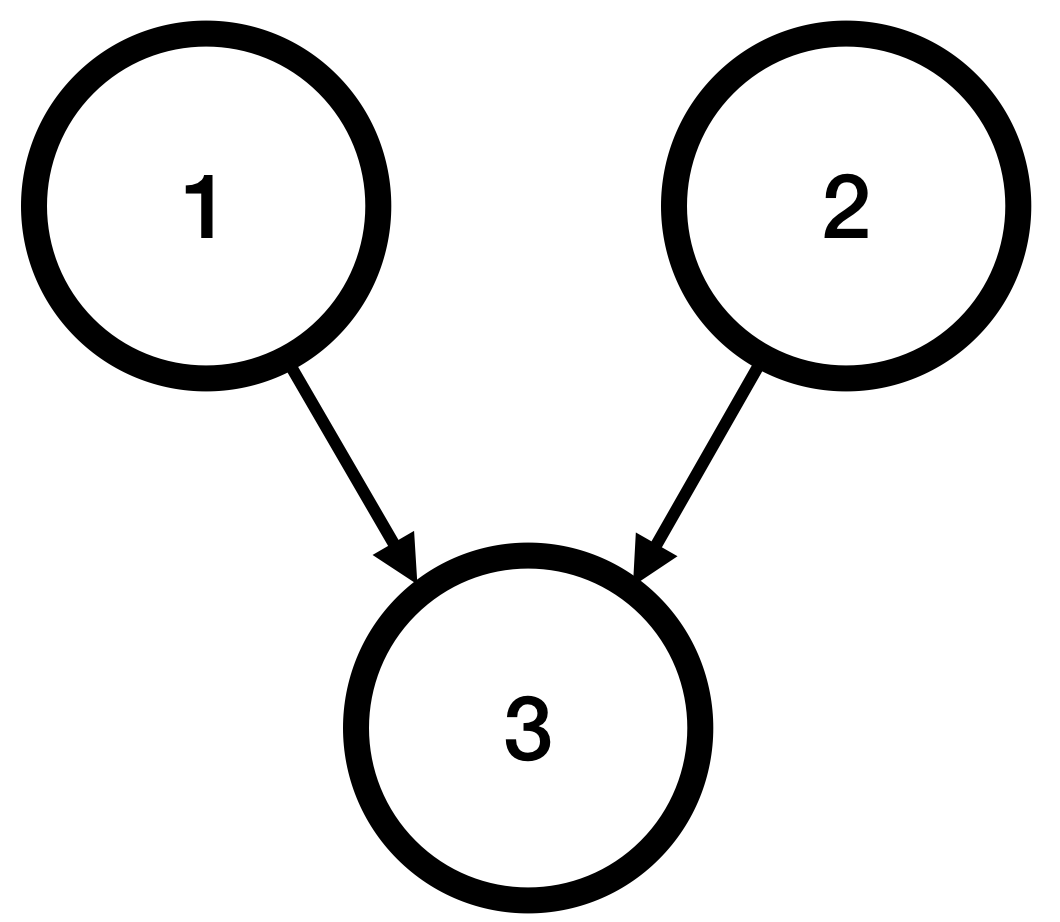} &
\hspace{0.25in}  &
\includegraphics[width=5.75cm]{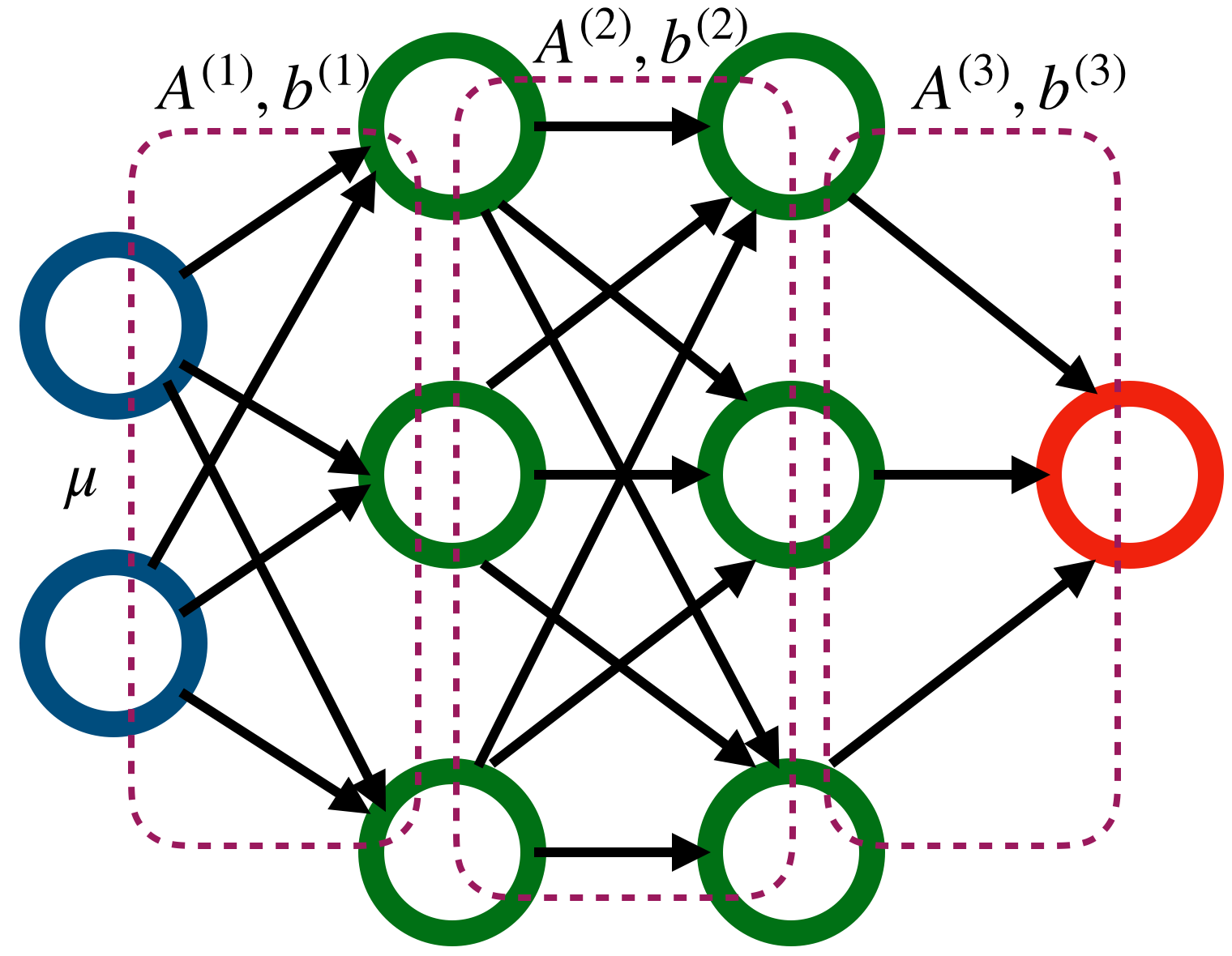} &
\hspace{0.25in}  &
\includegraphics[width=4cm]{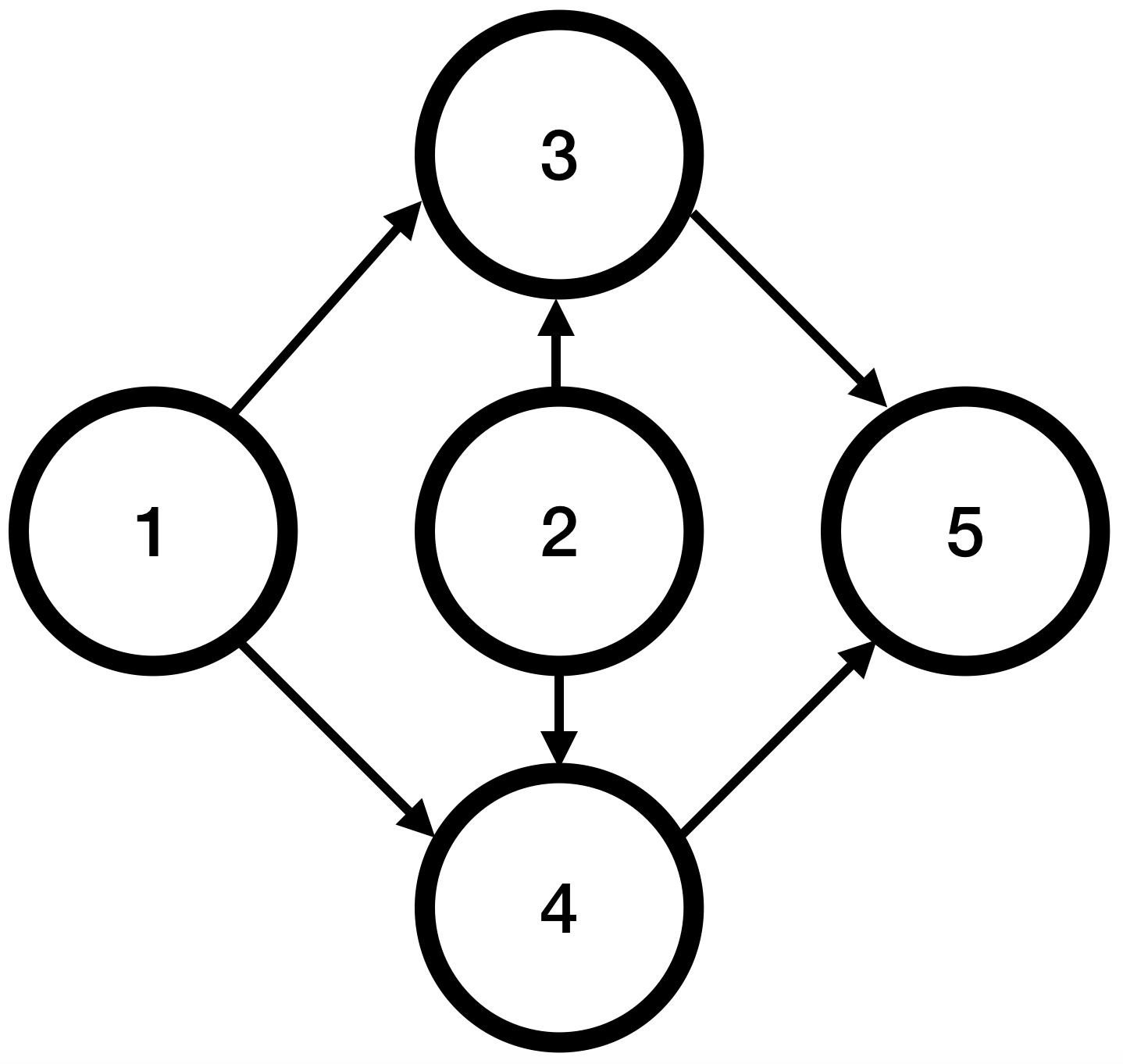} \\
(a) & & (b) & & (c) 
\end{tabular}
\caption{(a) A three-variable DAG with a v-structure; (b) A graphical illustration of a multilayer perceptron, with two hidden layers, $m_0=2$, $m_1=m_2=3$,  where $u$ is the input, $\Abf^{(\ell)}$ and $\bbf^{(\ell)}$ denote the corresponding parameters to produce the linear transformation for the $(\ell-1)$th layer; (c) A five-variable DAG.}
\label{fig:illustration}
\end{figure}

\begin{example}[v-structure]\label{examv}
\tsb{Consider three random variables $X_1, X_2, X_3$ that form a v-structure, as illustrated in Figure \ref{fig:illustration}(a), where $X_1$ and $X_2$ are the common parents of $X_3$. Even if $X_1$ and $X_2$ are \textit{marginally} independent, they can be \textit{conditionally} dependent given $X_3$. To better understand this, consider the following toy illustration. Either the ballgame or the rain could cause traffic jam, but they are uncorrelated. However, seeing traffic jam puts the ballgame and the rain in competition as a potential explanation. As such, these two events are conditionally dependent. Since $X_2$ is not a parent of $X_1$, both $H_0(1,2)$ and $H_1^*(1,2)$ hold. Consequently, testing for \eqref{eqn:hypothesis1} can have an inflated type-I error for testing \eqref{eqn:hypothesis}.}
\end{example}

In this example, we see the reason that testing for \eqref{eqn:hypothesis1} is not the same as for \eqref{eqn:hypothesis} is because the conditioning set of $X_1$ and $X_2$ contains their common descendant $X_3$. This key observation motivates us to consider a variant of \eqref{eqn:hypothesis1}, which we show is equivalent to \eqref{eqn:hypothesis} under certain conditions. We also remark that missing links in a DAG correspond to specific conditional independence between variables, but are not equivalent to marginal independence in general. 

Specifically, for a given set of indices $\mathcal{M}\subseteq \{1,\ldots,d\}$ such that $j\notin \mathcal{M}$, and letting $\Xbf_{\mathcal{M}-\{k\}}$ denote the set of variables in $\mathcal{M}-\{k\}$, we consider the hypotheses:
\begin{eqnarray}\label{eqn:hypothesis2}
\begin{split}
&H_0^*(j,k|\mathcal{M}):X_k\,\,\textrm{and}\,\, X_j\,\,\textrm{are}\,\,\textrm{CI\,\,given }\Xbf_{\mathcal{M}-\{k\}}, \,\,\,\,\,\textrm{versus}\,\,\,\,\\
&H_1^*(j,k|\mathcal{M}):X_k\,\,\textrm{and}\,\, X_j\,\,\textrm{are}\,\,\textit{\hbox{not}}\,\,\textrm{CI\,\,given\,\,}\Xbf_{\mathcal{M}-\{k\}},
\end{split}	
\end{eqnarray}

\begin{prop}\label{prop1}
For a given pair of nodes $(j,k)$ such that $j \in \tsb{\DS}_k$, $j, k = 1, \ldots, d$, and for any $\mathcal{M}$ such that $j \notin \mathcal{M}$,  $\tsb{\PA}_j\subseteq \mathcal{M}$ and $ \mathcal{M}\cap \tsb{\DS}_j = \emptyset$, testing \eqref{eqn:hypothesis2} is equivalent to testing \eqref{eqn:hypothesis}.
\end{prop}

\noindent
Proposition \ref{prop1} forms the basis for our test. That is, to infer the directed links, we first restrict our attention to the pairs $(j,k)$ such that $j\in \DS_k$. Apparently, $H_0(j,k)$ does not hold when $j \notin \DS_k$. Next, when devising a conditional independence test for $H_0(j,k)$, the conditioning set $\mathcal{M}$ is supposed to contain the parents of node $j$, but \emph{cannot} contain any common descendants of $j, k$. Under these conditions, we establish the equivalence between \eqref{eqn:hypothesis2} and \eqref{eqn:hypothesis}. A similar idea of using CI tests for DAG structural learning was employed in \citet{Spirtes2000} too. 

Next, we develop a test statistic for the hypotheses \eqref{eqn:hypothesis2}. We introduce a key quantity. Let $h$ denote a square-integrable function that takes $X_k$ and $\Xbf_{\mathcal{M}-\{k\}}$ as the input. Define
\begin{eqnarray*}
I(j,k|\mathcal{M};h)= \Mean \left\{ X_j - \Mean \left( X_j | \Xbf_{\mathcal{M}-\{k\}} \right) \right\} \left[ h\left( X_k,\Xbf_{\mathcal{M}-\{k\}} \right) - \Mean \left\{ h\left( X_k,\Xbf_{\mathcal{M}-\{k\}} \right) | \Xbf_{\mathcal{M}-\{k\}} \right\} \right].
\end{eqnarray*}

\noindent
Under the additive noise model \eqref{eqn:model}, the next theorem connects this quantity with the null hypothesis $H_0^*(j,k|\mathcal{M})$ in \eqref{eqn:hypothesis2}. Together with Proposition \ref{prop1}, it shows that $I(j,k|\mathcal{M};h)$ can serve as a test statistic for \eqref{eqn:hypothesis2}, and equivalently, for \eqref{eqn:hypothesis} that we target.

\begin{theorem}\label{thm1}
Suppose \eqref{eqn:model} holds. For a given pair of nodes $(j,k)$ such that $j\in \tsb{\hbox{DS}}_k$, $j,k=1, \ldots, d$, for any $\mathcal{M}$ such that $j\notin \mathcal{M}$,  $\tsb{\PA}_j \subseteq {\mathcal{M}}$ and ${\mathcal{M}} \cap \tsb{\DS}_j = \emptyset$, the null hypothesis $H_0^*(j,k|\mathcal{M})$ in \eqref{eqn:hypothesis2} is equivalent to $\sup_h |I(j,k|\mathcal{M};h)|=0$ where the supremum is taken over all square-integrable functions $h$.
\end{theorem}

\noindent
Theorem \ref{thm1} immediately suggests a possible testing procedure for \eqref{eqn:hypothesis2}. That is, we first employ a DAG estimator to learn the ancestors and descendants for node $j$. We then consider a natural choice for $h$, where $h\left( X_k, \Xbf_{\mathcal{M}-\{k\}} \right) = X_k$. Then $I(j,k|\mathcal{M};h)$ becomes
\begin{align} \label{eqn:DGT}
I(j,k|\mathcal{M};h) = \Mean \left\{ X_j - \Mean \left( X_j | \Xbf_{\mathcal{M}-\{k\}} \right) \right\} \left\{ X_k - \Mean \left( X_k|\Xbf_{\mathcal{M}-\{k\}} \right) \right\}.
\end{align}
By Theorem \ref{thm1}, under the null hypothesis $H_0^*(j,k|\mathcal{M})$, a consistent estimator for \eqref{eqn:DGT} should be close to zero. A Wald type test can then be devised with i.i.d.\ data. That is, we first obtain an estimator $\widehat{I}_{j,k}$ for $I(j,k|\mathcal{M};h)$, by plugging in the estimators of the conditional mean functions, $\widehat{\Mean} \left( X_j | \Xbf_{\mathcal{M}-\{k\}} \right)$ and $\widehat{\Mean} \left( X_k|\Xbf_{\mathcal{M}-\{k\}} \right)$. We then get an estimator of its asymptotic variance $\widehat{\sigma}^2_{j,k}$, and obtain the Wald type test statistic, $\sqrt{N} \widehat{\sigma}_{j,k}^{-1} \widehat{I}_{j,k}$, where $N$ is the number of samples. Such a test is similar in spirit as the tests of \citet{zhang2018measuring} and \cite{shah2018hardness}. Since it involves estimation of two conditional mean functions, we refer to it as the \emph{double regression-based test}. We later numerically compare our proposed test with this test. 

On the other hand, this double regression-based test has some limitations. One is that it requires the set $\mathcal{M}$ to be fixed. To meet the requirement in Proposition \ref{prop1}, $\mathcal{M}$ needs to be determined in a data-adaptive way. The resulting test may not control the type-I error due to the dependence between $\mathcal{M}$ and the estimator of the mean functions in $\widehat{I}_{j,k}$. Another limitation is that it may not have a sufficient power to detect $H_1(j,k)$. As an illustration, we revisit Example \ref{examv}. For this example, consider the structural equation model: $X_1=\varepsilon_1$, $X_2=\varepsilon_2$, and $X_3=X_1^2+X_2+\varepsilon_3$. Under this model, $H_1(1,3)$ holds. Meanwhile, $I(1,3) = \Mean (X_3-X_2)X_1 = \Mean \varepsilon_1^3$. When the distribution of $\varepsilon_1$ is symmetric, $I(1,3) = 0$, despite the fact that $X_1$ is a parent of $X_3$. As such, for this example, the double regression-based  test is to have no power at all.

To address the first limitation, we employ the sample splitting strategy to ensure its size control. To address the second limitation, we consider multiple transformation functions $h$, instead of a single $h$, to improve the power. We detail our idea in Section \ref{sec:test}.

\subsection{Time-dependent observational data}
\label{sec:time-dependency}

Throughout this article, we use $X$ to denote the population variables, and $\mathbb{X}$ to denote the data realizations. Suppose the data come from an observational study, and are of the form, $\{\Xbm_{i,t,j} : i=1, \ldots, N, t=1, \ldots, T_i, j=1, \ldots, d\}$, where $i$ indexes the $i$th subject,  $t$ indexes the $t$th time point, and $j$ indexes the $j$th random variable. Suppose there are totally $N$ subjects, with $T_i$ observations for the $i$th subject. Write $\Xbm_{i,t} = (\Xbm_{i,t,1}, \ldots, \Xbm_{i,t,d})^\top$, $i=1, \ldots, N, t=1, \ldots, T_i$. We consider the following data structure.
\vspace{-0.1in}
\begin{enumerate}[({B}1)]
\setlength\itemsep{-0.25em}
\item Across subjects, the measurements $\Xbm_{1,t}$, $\ldots$, $\Xbm_{N,t}$ are i.i.d.
	
\item Across time points, the random vectors $\Xbm_{i,1}$, $\ldots$, $\Xbm_{i,T_i}$ are stationary.
	
\item For any $i,t$, $\Xbm_{i,t,1}$, $\ldots$, $\Xbm_{i,t,d}$ are DAG-structured. In addition, their joint distribution is the same as that of $X_1,\ldots,X_d$.
\vspace{-0.1in}
\end{enumerate}

\noindent
Condition (B1) is reasonable, as the subjects are usually independent from each other. We do not study the scenario where the data come from the same families or clusters. Condition (B2) about the stationarity is common in numerous applications such as brain connectivity analysis \citep{Bullmore2009, QiuHan2016, WangKang2016}.  Condition (B3) brings the data into the DAG framework that we study. Note that (B3) does not allow directed edges from past to future observations. Meanwhile, we discuss the extensions of our test for non-stationary DAG, or for past to future edges, in Section \ref{sec:ext-nonstationary} of the Appendix.

\section{Testing Procedure}
\label{sec:test}

In this section, we develop an inferential procedure for the hypotheses in \eqref{eqn:hypothesis} for a given pair $(j,k)$, through \eqref{eqn:hypothesis2}, given the observational data $\Xbm_{i,t}$. We first present the main ideas and the complete procedure, then detail the major steps. As our test is based on \textbf{S}tructural learning, s\textbf{U}pervised learning, and \textbf{G}enerative \textbf{A}dve\textbf{R}sarial networks, we call our method SUGAR.

\subsection{The main algorithm}
\label{sec:algo}

Our main idea is to construct a series of measures $\{I(j,k|\mathcal{M};h_b) : b=1,\ldots,B\}$, for a large number of transformation functions $h_1, \ldots,$ $h_B$, then take the maximum of some standardized version of $I(j,k|\mathcal{M};h_b)$. Toward that goal, our test involves three key components: 

\vspace{-0.1in}
\begin{enumerate}[(a)]
\setlength\itemsep{-0.25em}
\item A DAG structural learning method to learn the set of indices $\mathcal{M}$ that satisfy Proposition \ref{prop1};
	
\item A supervised learning method to estimate the conditional mean function $\Mean \left( X_j|\Xbf_{\mathcal{M}-\{k\}} \right)$;
	
\item A distribution generator to approximate the conditional distribution of the variables.
\vspace{-0.1in}
\end{enumerate}

For (a), we apply a structural learning algorithm to learn the underlying DAG $\mathcal{G}$ corresponding to $\Xbf$. The input of this step is the observed data $\{\Xbm_{i,t,j} : i=1, \ldots, N, t=1, \ldots, T_i, j=1, \ldots, d\}$, and the output is the estimated DAG. We then set $\mathcal{M}$ as the estimated set of ancestors of $X_j$. To capture possible sparsity and nonlinear associations in $\mathcal{G}$, we employ the DAG estimation method of \cite{zheng2020learning}. See Section \ref{sec:CSL} for details. 

For (b), we employ a supervised learning algorithm. The input of this step is $\Xbf_{\mathcal{M}-\{k\}}$ that serves as the ``predictors", and $X_j$ that serves as the ``response", and the output is the estimated mean function $\widehat{\Mean} \left( X_j|\Xbf_{\mathcal{M}-\{k\}} \right)$. We employ a multilayer perceptron learner, which has a good capacity of estimating complex high-dimensional mean, and the estimator has the desired consistency guarantees \citep{farrell2018deep}. See Section \ref{sec:SL} for details.

For (c), we propose to use generative adversarial networks \citep[GANs]{goodfellow2014generative} to approximate the conditional distribution of $X_k$ given $\Xbf_{\mathcal{M}-\{k\}}$. The input of this step is $\Xbm_{i,t,\mathcal{M}-\{k\}}$ and multivariate Gaussian noise vectors, and the output is the learnt generator model, with a set of $M$ pseudo samples $\widetilde{\Xbm}_{i,t,k}^{(s,m)}$, $m = 1, \ldots, M$, that have a similar distribution as the training samples. We employ a generator model with the Sinkhorn divergence loss \citep{genevay2017learning} to mitigate the potential bias of GANs. See Section \ref{sec:gans} for details. 

Given the generated pseudo samples, we then proceed to estimate the conditional mean function $\Mean \big\{ h_b\big(X_k, \Xbf_{\mathcal{M}-\{k\}}\big) |$ $\Xbf_{\mathcal{M}-\{k\}} \big\}$ in \eqref{eqn:DGT}, and construct the corresponding test statistic. We also incorporate the data-splitting and cross-fitting strategy \citep{romano2019}, to ensure a valid type-I error control for the test under minimal conditions for the above three learners. Specifically, we randomly split the samples into two equal halves $\mathcal{I}_1 \cup \mathcal{I}_2$, where $\mathcal{I}_{s}$ denotes the set of subsample indices, $s=1,2$. We then compute the three learners in (a) to (c) using each half of the data separately. Based on these learners, we next use cross-fitting to estimate $\{I(j,k|\mathcal{M};h_b)\}_{b=1}^{B}$, and their associated standard deviations. We construct our test statistic as the largest standardized version of $I(j,k|\mathcal{M};h_b)$ in the absolute value. This leads to two Wald-type test statistics, one for each half of the data. Finally, we derive the $p$-values based on Gaussian approximation, and reject the null when either one of the $p$-value is smaller than $\alpha/2$. By Bonferroni's inequality, this yields a valid $\alpha$-level test. See Section \ref{sec:pvalue} for details. 

A summary of the proposed testing procedure is given in Algorithm \ref{alg:full}. 

\begin{algorithm}[t!]
\caption{Testing procedure for a given edge $(j,k)$.}
\label{alg:full}
\begin{algorithmic}
\item[]\normalsize
\begin{enumerate}[Step 1.]
\setlength\itemsep{-0.25em}
\item Randomly split the data into two equal halves, $\{\Xbm_{i,t,k}\}_{i \in \mathcal{I}_{s}, t=1, \ldots, T_i}$, $s=1,2$.
			
\item For each half of the data, $s=1,2$,
\vspace{-0.1in}
\begin{enumerate}[({2}a)]
\item Apply the structural learning method \eqref{eqn:DAG} to estimate the DAG $\mathcal{G}$. Denote the estimated set of ancestors of $X_j$ by $\widehat{\hbox{AC}}^{(s)}_j$. Set $\mathcal{M}^{(s)} = \widehat{\hbox{AC}}^{(s)}_j-\{k\}$.
\item If $k\notin \widehat{\hbox{AC}}^{(s)}_j$, return the $p$-value, $p^{(s)}(j,k)=1$.
\end{enumerate}

\vspace{-0.1in}			
\item For $s=1,2$, apply the supervised learning method \eqref{eqn:SL} to estimate the conditional mean function $\Mean \big( X_j | \Xbf_{\mathcal{M}^{(s)}} \big)$, and denote the estimator by $\widehat{g}^{(s)}$. 
			
\item For $s=1,2$, apply the GANs method to learn a generator model to approximate the conditional distribution of $X_k$ given $\Xbf_{\mathcal{M}^{(s)}-\{k\}}$. It returns the learnt  generator $\mathbb{G}^{(s)}$, and a set of pseudo samples $\big\{ \widetilde{\Xbm}_{i,t,k}^{(s,m)} \big\}_{i \in \mathcal{I}_{s}, t=1, \ldots, T_i, m=1, \ldots, M}$.
			
\item Construct the test statistic:
\vspace{-0.15in}
\begin{enumerate}[({5}a)]
\item Randomly generate $B$ functions $\big\{ h_b^{(s)} \big\}_{b=1}^B$ from the class $\mathbb{H}^{(s)}$ in \eqref{eqn:funclass}.		
\item For each $(s,b)$, construct two standardized measures, $\widehat{T}_{b,\textrm{CF}}^{(s)}$ and $\widehat{T}_{b,\textrm{NCF}}^{(s)}$, with and without cross-fitting, using \eqref{eqn:measure}.
\item Select the index, $\widehat{b}^{(s)}=\argmax_{b\in \{1,\ldots,B\}} \big| \widehat{T}_{b,\textrm{NCF}}^{(s)} \big|$, based on the measure without cross-fitting.			
\item Set the test statistic as $\widehat{T}_{\widehat{b}^{(s)},\textrm{CF}}^{(s)}$, based on the measure with cross-fitting.
\end{enumerate}

\vspace{-0.1in}			
\item Return the $p$-value:
\vspace{-0.1in}
\begin{enumerate}[({6}a)]
\item Compute the $p$-value, $p^{(s)}(j,k)=2\prob \big\{ Z_0 \ge \big| \widehat{T}_{\widehat{b}^{(s)},\textrm{CF}}^{(s)} \big| \big\}$, for each half of the data, $s = 1, 2$, where $Z_0$ is a standard normal random variable.
\item Return $p(j,k)=2\min\left\{ p^{(1)}(j,k),p^{(2)}(j,k) \right\}$.
\end{enumerate}
\end{enumerate}
\end{algorithmic}
\end{algorithm}

\subsection{Test statistic and $p$-value}
\label{sec:pvalue}

We begin with the presentation of our test, including the test statistic and the computation of the $p$-value, which are built on the three learners in (a) to (c) that we discuss in detail  later.

First, for each half of the data, $s=1,2$, we begin with a bounded function class $\mathbb{H}^{(s)} = \left\{ h_{\omegabf}^{(s)} : \omegabf \in \Omegabf^{(s)} \right\}$, indexed by some parameter $\omegabf$. In our implementation, we consider the class of characteristic functions of $X_k$,
\vspace{-0.05in}
\begin{eqnarray}\label{eqn:funclass}
\mathbb{H}^{(1)} = \mathbb{H}^{(2)} = \mathbb{H} = \big\{ \cos(\omega X_k), \; \sin(\omega X_k):\omega\in \mathbb{R} \big\}.
\end{eqnarray}
We note that \eqref{eqn:funclass} is not able to approximate the entire class of square integrable functions. Nevertheless, our numerical experiments have found that setting $\mathbb{H}^{(s)}$ according to \eqref{eqn:funclass} results in a good power empirically. Moreover, we note that one may set $\mathbb{H}^{(s)}$ to the class of characteristic functions of $(X_k,X_{\mathcal{M}^{(s)}})$. By the Fourier Theorem \citep{siebert1986circuits}, this alternative choice can approximate any square integrable function $h$, and the resulting test is consistent against all alternatives. We choose \eqref{eqn:funclass} for its simplicity as well as good empirical performance. Without loss of generality, we choose an even number for the total number of transformation functions $B$. We randomly generate i.i.d.\ standard normal variables $\omega_1,\ldots,\omega_{B/2}$, and set 
\begin{eqnarray*}
h_b^{(s)}\left( X_k,\Xbf_{\mathcal{M}^{(s)}} \right) =
\begin{cases}
\cos(\omega_b X_k), & \textrm{ for } b=1, \ldots, B/2, \\
\sin(\omega_b X_k), & \textrm{ for } b = B/2+1, \ldots, B.
\end{cases}
\end{eqnarray*}

Next, for each pair of $(s,b)$, $b=1, \ldots, B, s = 1, 2$, let $\widehat{\textrm{AC}}_j^{(s)}$, $\mathcal{M}^{(s)}$, $\widehat{g}^{(s)}$, and $\{\widetilde{\Xbm}_{i,t,k}^{(s,m)}\}$ denote the estimated set of ancestors of $X_j$, the estimated set of indices $\mathcal{M}$, the estimated conditional mean function, and the generated pseudo samples, obtain from the components (a) to (c), respectively. We compute two estimators $\widehat{I}_{b,\textrm{CF}}^{(s)}$ and $\widehat{I}_{b,\textrm{NCF}}^{(s)}$ for the measure $I\left( j,k | \widehat{\textrm{AC}}_j^{(s)},h_b^{(s)} \right)$, one \emph{with} cross-fitting, and the other \emph{without} cross-fitting. Specifically, we compute
\begin{eqnarray*}
\widehat{I}_{b,\textrm{CF}}^{(s)} = \left( \textstyle\sum_{i\in \mathcal{I}_{s}^c} T_i \right)^{-1} \left( \textstyle\sum_{i\in \mathcal{I}_{s}^c} I_{i,t,b}^{(s)} \right),
\quad
\widehat{I}_{b,\textrm{NCF}}^{(s)} & = & \left( \textstyle\sum_{i\in \mathcal{I}_{s}} T_i \right)^{-1} \left( \textstyle\sum_{i\in \mathcal{I}_{s}} I_{i,t,b}^{(s)} \right),
\end{eqnarray*}
where
\vspace{-0.1in}
\begin{align*} 
\begin{split}	
I_{i,t,b}^{(s)} =
\left\{ \Xbm_{i,t,j}-\widehat{g}^{(s)}\left( \Xbm_{i,t,\mathcal{M}^{(s)}} \right) \right\}
\left\{ h_b^{(s)}\left( \Xbm_{i,t,k},\Xbm_{i,t,\mathcal{M}^{(s)}} \right) - \frac{1}{M}\sum_{m=1}^M h_b^{(s)}\left( \widetilde{\Xbm}_{i,t,k}^{(s,m)}, \Xbm_{i,t,\mathcal{M}^{(s)}} \right) \right\},
\end{split}	
\end{align*}
and $M$ is the total number of pseudo samples. We note that, for $\widehat{I}_{b,\textrm{NCF}}^{(s)}$, we use the same subset of data to learn the graph, the generator, the condition mean function, and to construct $I_{i,t,b}^{(s)}$. By contrast, for $\widehat{I}_{b,\textrm{CF}}^{(s)}$, the data used for the DAG learner, the conditional mean learner and the generator are independent from the data used to construct $I_{i,t,b}^{(s)}$.

Next, we compute the corresponding standard errors $\widehat{\sigma}_{b,\textrm{CF}}^{(s)}$ and $\widehat{\sigma}_{b,\textrm{NCF}}^{(s)}$ for $\widehat{I}_{b,\textrm{CF}}^{(s)}$ and $\widehat{I}_{b,\textrm{NCF}}^{(s)}$, respectively. Since our data are time-dependent, the usual sample variance would not be a consistent estimator. Therefore, we employ the batched estimator common in time series analysis \citep{Carlstein1986}. That is, we divide the data associated with each subject into non-overlapping batches, with each batch containing at most $K$ observations. For simplicity, suppose $T_i$ is divisible by $K$ for all $i=1, \ldots, N$. We obtain the following standard error estimators,
\vspace{-0.01in}
\begin{align*} 
\begin{split}
\widehat{\sigma}_{b,\textrm{CF}}^{(s)} & = \left[\frac{K}{\sum_{i\in \mathcal{I}_{s}^c} T_i} \sum_{i\in \mathcal{I}_{s}^c}\sum_{k=1}^{T_i/K} \left\{ \frac{\sum_{t=(k-1)K+1}^{kK} \left( I_{i,t,b}^{(s)}-\widehat{I}_{b,\textrm{CF}}^{(s)} \right) }{\sqrt{K}} \right\}^2 \right]^{1/2},\\
\widehat{\sigma}_{b,\textrm{NCF}}^{(s)} & = \left[ \frac{K}{\sum_{i\in \mathcal{I}_{s}} T_i} \sum_{i\in \mathcal{I}_{s}}\sum_{k=1}^{T_i/K} \left\{ \frac{\sum_{t=(k-1)K+1}^{kK} \left( I_{i,t,b}^{(s)}-\widehat{I}_{b,\textrm{NCF}}^{(s)} \right) }{\sqrt{K}} \right\}^2 \right]^{1/2}.	
\end{split}
\end{align*}

Putting $\widehat{I}_{b,\textrm{CF}}^{(s)}$ and $\widehat{I}_{b,\textrm{NCF}}^{(s)}$ together with their standard error estimators, we obtain two standardized measures,
\begin{eqnarray}\label{eqn:measure}
\begin{split}
\quad\quad\quad 
\widehat{T}_{b,\textrm{CF}}^{(s)} = \sqrt{\textstyle\sum_{i\in \mathcal{I}_{s}^c} T_i} \, \left( \widehat{\sigma}_{b,\textrm{CF}}^{(s)} \right)^{-1} \widehat{I}_{b,\textrm{CF}}^{(s)}, \; \textrm{ and } \;
\widehat{T}_{b,\textrm{NCF}}^{(s)} = \sqrt{\textstyle\sum_{i\in \mathcal{I}_{s}} T_i} \, \left( \widehat{\sigma}_{b,\textrm{NCF}}^{(s)} \right)^{-1} \widehat{I}_{b,\textrm{NCF}}^{(s)}.
\end{split}	
\end{eqnarray}
We then select the index $\widehat{b}^{(s)}$ that maximizes the standardized measure without cross-fitting, $\widehat{T}_{b,\textrm{NCF}}^{(s)}$, in absolute value, i.e., $\widehat{b}^{(s)} = \argmax_{b\in \{1,\ldots,B\}} \left| \widehat{T}_{b,\textrm{NCF}}^{(s)} \right|$. We take the measure with cross-fitting, $\widehat{T}_{\widehat{b}^{(s)},\textrm{CF}}^{(s)}$, under the selected $\widehat{b}^{(s)}$, as our final test statistic.

We make a few remarks. First, we use the cross-fitting measure to construct the test statistic $\widehat{T}_{\widehat{b}^{(s)},\textrm{CF}}^{(s)}$. This enables us to derive its limiting distribution more easily. Specifically, conditional on the data in $\mathcal{I}_s$, for each $b = 1, \ldots, B$, $\widehat{T}_{b,\textrm{CF}}^{(s)}$ converges in distribution to standard normal under the null. Since $\widehat{b}^{(s)}$ is determined by $\widehat{T}_{b,\textrm{NCF}}^{(s)}$, the index $\widehat{b}^{(s)}$ depends solely on the data in $\mathcal{I}_s$.  Consequently, conditional on the data in $\mathcal{I}_s$, $\widehat{T}_{\widehat{b}^{(s)},\textrm{CF}}^{(s)}$ converges in distribution to standard normal under the null as well. By contrast, the limiting distribution of the no-cross-fitting measure $\widehat{T}_{\widehat{b}^{(s)},\textrm{NCF}}^{(s)}$ is unclear, due to the complicated dependence between $\widehat{b}^{(s)}$ and $\widehat{T}_{b,\textrm{NCF}}^{(s)}$.

Second, we use the no-cross-fitting measure to select the index $\widehat{b}^{(s)}$. As we show in Section \ref{sec:theory}, when the estimated conditional mean function and the distributional generator belong to the VC type class \citep[Definition 2.1]{Cherno2014}, the index $\widehat{b}^{(s)}$ that maximizes the no-cross-fitting measure $\{\widehat{T}_{b,\textrm{NCF}}^{(s)}\}$ asymptotically maximizes the cross-fitting measure $\{\widehat{T}_{b,\textrm{CF}}^{(s)}\}$ as well. This choice of the index $\widehat{b}^{(s)}$ is to maximize the power of the resulting test.

Finally, the random binary data splitting may introduce some sampling uncertainty. This issue is mitigated in our test, since we construct two test statistics based on both data subsets, then combine them to derive the final decision rule. One may also consider the multiple binary-splits idea of \citet{meinshausen2009}, or the multi-split idea of \citet{romano2019}. We discuss a multiple binary-splits version of our test in Section \ref{sec:more-splitting} of the Appendix.

\subsection{DAG structural learning}
\label{sec:CSL}

We next discuss the three key learning components (a) to (c) of our proposed test. The first is to estimate the DAG $\mathcal{G}$ associated with $X = (X_1, \ldots, X_d)^\top$, and to construct $\mathcal{M}$. In our implementation, we employ the neural structural learning method of \cite{zheng2020learning}. Other methods, e.g., \cite{yu2019dag, zhu2020causal}, can be used as well.

Consider a multilayer perceptron (MLP) with $L$ hidden layers and an activation function $\sigma$: 
\begin{align} \label{eqn:MLP}
\begin{split}
\textrm{MLP}\left( \ubf; \Abf^{(1)}, \bbf^{(1)}, \ldots, \Abf^{(L)}, \bbf^{(L)} \right) 
= \; \Abf^{(L)} \sigma\left\{ \ldots \Abf^{(2)} \sigma\left( \Abf^{(1)} \mubf + \bbf^{(1)} \right) \ldots + \bbf^{(L-1)} \right\} + \bbf^{(L)},
\end{split}	
\end{align}
where $\ubf \in \mathbb{R}^{m_0}$ is the input signal of the MLP, $\Abf^{(s)} \in \mathbb{R}^{m_{\ell} \times m_{\ell-1}}, \bbf^{(s)} \in \mathbb{R}^{m_{\ell}}$ are the parameters that produce the linear transformation of the $(\ell-1)$th layer, the output is a scalar with $m_L=1$, and there are $m_{\ell}$ nodes at layer $\ell$, $\ell = 0, \ldots, L$. See Figure \ref{fig:illustration}(b) for a graphical illustration. 

We employ MLP to approximate the functions $f_j$'s in our DAG model \eqref{eqn:model}. In our theoretical analysis, we focus on the setting where $f_j$'s are a set of continuous functions. Meanwhile, we may also consider a family of piecewise smooth functions \citep{imaizumi2019deep} for $f_j$'s. In both cases, neural networks models such as MLP can consistently estimate $f_j$'s. Let $\thetabf_j = \big\{ \Abf_j^{(\ell)}, \bbf_j^{(\ell)}: 1\le \ell\le L \big\}$ collect all the parameters for the $j$th MLP that approximates $f_j$, and let $\theta = \{ \thetabf_j \}_{j=1}^{d}$. Accordingly, $\theta$ uniquely determines a graph structure, i.e., how the variables are dependent to each other in the graph. We call this structure the graph induced by $\theta$, and denote it by $\mathcal{G}(\theta)$. For each half of the data, $s = 1, 2$, we estimate the DAG via
\begin{eqnarray*}
\min_{\thetabf} \sum_{i\in \mathcal{I}_{s}} \sum_{t,j} \big\{ \Xbm_{i,t,j}-\textrm{MLP}(\Xbm_{i,t};\thetabf_j) \big\}^2, \,\, \textrm{ subject to } \, \mathcal{G}(\thetabf) \, \textrm{ is a DAG}.
\end{eqnarray*}
This optimization, however, is challenging to solve, mainly due to the fact that the search space scales super-exponentially with the dimension $d$. To resolve this issue, \cite{zheng2020learning} proposed a novel characterization of the acyclic constraint, and showed that the DAG constraint can be represented by trace$[\exp\{ \Wbf(\thetabf) \circ \Wbf(\thetabf) \}]=d$, where $\circ$ denotes the Hadamard product, $\exp(\Wbf)$ is the matrix exponential of $\Wbf$, trace$(\Wbf)$ is the trace of $\Wbf$, and $\Wbf(\thetabf)$ is a $d\times d$ matrix whose $(k,j)$th entry equals the Euclidean norm of the $k$th column of $\Abf_j^{(1)}$. Based on this characterization, the above optimization problem becomes,
\begin{align}\label{eqn:DAG}
\begin{split}
\min_{\thetabf} \sum_{j=1}^{d} \left[\sum_{i\in \mathcal{I}_{s}} \sum_{t=1}^{T_i}  \left\{ \Xbm_{i,t,j}-\textrm{MLP}(\Xbm_{i,t};\thetabf_j)\right\}^2+\lambda  n_s\big\| \Abf_j^{(1)} \big\|_{1,1}\right], \\
\textrm{subject to}\,\,\,\,\textrm{trace}[\exp\{ W(\thetabf) \circ W(\thetabf) \}]=d,
\end{split}	
\end{align}
where $n_s=\sum_{i\in \mathcal{I}_s} T_i$ is the number of observations in $\mathcal{I}_s$, $\big\| \Abf_j^{(1)} \big\|_{1,1}$ is the sum of all elements in $\Abf_j^{(1)}$ in absolute values, and $\lambda>0$ is a sparsity tuning parameter. Note that the sparsity penalization is placed only on $\Abf_j^{(1)}$, since this is the only layer that determines the sparsity of the input variables $X_1, \ldots, X_d$. This new optimization problem in \eqref{eqn:DAG} can be efficiently solved using the augmented Lagrangian method \citep{zheng2020learning}. 

Let $\widehat{\mathcal{G}}^{(s)}$ denote the estimated graph, and $\widehat{\AC}_j$ and $\widehat{\PA}_j$ denote the corresponding estimated set of ancestors and parents of $X_j$, respectively. If $k \notin \widehat{\textrm{AC}}_j^{(s)}$, then it follows from $\textrm{PA}_j\subseteq \widehat{\textrm{AC}}_j^{(s)}$ that $k\notin \PA_j$. Consequently, we simply set the corresponding $p$-value $p^{(s)}(j,k)=1$. Our subsequent testing procedure is to focus on the case where $k \in \widehat{\textrm{AC}}_j^{(s)}$, and we set $\mathcal{M}^{(s)} = \widehat{\hbox{AC}}^{(s)}_j-\{k\}$. We also remark that, to establish the consistency of our test, we only require $\prob(\textrm{PA}_j\subseteq \widehat{\textrm{AC}}_j^{(s)} \subseteq \DS_j^c -\{j\})\to 1$, where $\DS_j^c$ denotes the complement of the set $\DS_j$. This essentially requires the order of the DAG to be consistently estimated. We later show in Section \ref{sec:thmcsl} that this condition is satisfied when using the method of \cite{zheng2020learning}. Meanwhile, this order consistency is much weaker than requiring the DAG estimator $\widehat{\mathcal{G}}^{(s)}$ to be selection consistent, i.e., $\prob(\PA_j= \widehat{\PA}_j) \to 1$, or to satisfy sure screening, i.e., $\prob(\PA_j \subseteq \widehat{\PA}_j) \to 1$.

\subsection{Supervised learning}
\label{sec:SL}

The second key component of our test is to learn the conditional mean $g^{(s)}(x)=\Mean \left( X_j|\Xbf_{\mathcal{M}^{(s)}}=x \right)$. This is essentially a regression problem, and there are many choices, e.g., boosting, random forests, or neural networks. In our implementation, we use the MLP again, by seeking 
\begin{eqnarray}\label{eqn:SL}
\min_{\thetabf_j} \sum_{i\in \mathcal{I}_{s}} \sum_{t=1}^{T_i} \left\{ \Xbm_{i,t,j}-\textrm{MLP}\left( \Xbm_{i,t,\mathcal{M}^{(s)}};\thetabf_j \right) \right\}^2,
\end{eqnarray}
where the learner MLP$(\cdot)$ is as defined in \eqref{eqn:MLP}.  The optimization problem in \eqref{eqn:SL} can be solved using a stochastic gradient descent algorithm, or the limited-memory Broyden-Fletcher-Goldfarb-Shanno algorithm \citep{byrd1995limited}.

\subsection{Generative adversarial learning}
\label{sec:gans}

The third key component of our test is to use GANs to learn a generator $\mathbb{G}^{(s)}(\cdot, \cdot)$, which generates a set of pseudo samples that have a similar distribution as the training samples. More accurately, in our setting, we learn the generator $\mathbb{G}(\cdot,\cdot)$ that takes $\Xbm_{i,t,\mathcal{M}-\{k\}}$ and a set of multivariate Gaussian noise vectors as the input, and the output are a set of pseudo samples $\widetilde{\Xbm}_{i,t,k}^{(s,m)}$. We train the generator such that the divergence between the conditional distribution of $\Xbm_{i,t,k}$ given $\Xbm_{i,t,\mathcal{M}-\{k\}}$ and that of $\widetilde{\Xbm}_{i,t,k}^{(s,m)}$ given $\Xbm_{i,t,\mathcal{M}-\{k\}}$ is minimized.

More specifically, we adopt \cite{genevay2017learning} to learn the generator $\mathbb{G}^{(s)}$, by optimizing
\begin{eqnarray}\label{eqn:GAN}
\min_{\mathbb{G}} \max_c \widetilde{\mathcal{D}}_{c,\rho}(\mu, \nu),
\end{eqnarray}
where $\mu$ and $\nu$ denote the joint distribution of $\left( \Xbm_{i,t,k},\Xbm_{i,t,\mathcal{M}^{(s)}} \right)$ and $\left( \widetilde{\Xbm}_{i,t,k}^{(s,m)},\Xbm_{i,t,\mathcal{M}^{(s)}} \right)$, respectively, and $\widetilde{\mathcal{D}}_{c,\rho}$ is the Sinkhorn loss function between two probability measures. The loss $\widetilde{\mathcal{D}}_{c,\rho}$ is with respect to a cost function $c$ and a regularization parameter $\rho>0$,
\vspace{-0.05in}
\begin{eqnarray*}
\mathcal{\widetilde{D}}_{c, \rho}(\mu, \nu) & = & 2\mathcal{D}_{c, \rho}(\mu, \nu) - \mathcal{D}_{c, \rho}(\mu, \mu) - \mathcal{D}_{c, \rho}(\nu, \nu), \\
\mathcal{D}_{c, \rho}(\mu, \nu) & = & \inf_{\pi \in \Pi(\mu, \nu)} \int_{x,y} \big\{ c(x, y)-\rho H(\pi|\mu  \otimes \nu) \big\} \pi(dx,dy),
\end{eqnarray*}
where $\Pi(\mu, \nu)$ is a set containing all probability measures $\pi$ whose marginal distributions correspond to $\mu$ and $\nu$, $H$ is the Kullback-Leibler divergence, and $\mu \otimes \nu$ is the product measure of $\mu$ and $\nu$. When $\rho = 0$, $\mathcal{D}_{c, 0}(\mu, \nu)$ measures the optimal transport of $\mu$ into $\nu$ with respect to the cost function $c(\cdot, \cdot)$ \citep{cuturi2013sinkhorn}. When $\rho \neq 0$, an entropic regularization is added to this optimal transport. As such, the objective function $\mathcal{\widetilde{D}}_{c, \rho}$ in \eqref{eqn:GAN} is a regularized optimal transport metric, where the regularization is to facilitate the computation, so that $\mathcal{\widetilde{D}}_{c, \rho}$ can be efficiently evaluated. Intuitively, the closer the two conditional distributions, the smaller the Sinkhorn loss. Therefore, maximizing $\mathcal{\widetilde{D}}_{c, \rho}$ with respect to the cost $c$ learns a discriminator that can better discriminate $\mu$ and $\nu$. On the other hand, minimizing the maximum cost with respect to the generator $\mathbb{G}$ makes the conditional distribution of $\widetilde{\Xbm}_{i,t,k}^{(s,m)}$ given $\Xbm_{i,t,\mathcal{M}^{(s)}}$ closer to that of $\Xbm_{i,t,k}$ given $\Xbm_{i,t,\mathcal{M}^{(s)}}$. This yields the minimax formulation in \eqref{eqn:GAN}.  In our implementation, we approximate the cost function $c$ and the generator based on MLP \eqref{eqn:MLP}. We approximate the distributions $\mu_{j,k}$ and $\nu_{j,k}$ in \eqref{eqn:GAN} by the empirical distributions of the data samples. We update the parameters in GANs by the Adam algorithm \citep{Kingma2015}. 

We again make a few remarks. First, we choose the Gaussian noise as the input for GANs. We have found the performance of the generator is not overly sensitive to the choice of the distribution of the input noise. We present more discussion and some additional numerical results in Section \ref{sec:more-noise} of the Appendix. Besides, we choose GANs based on the Sinkhorn divergence loss to mitigate the potential bias of traditional GANs. Moreover, in addition to GANs, other deep generative learning approaches such as variational auto-encoders \citep{kingma2013auto} are equally applicable here. Second, we note that, based on the estimated conditional distribution from GANs, one can derive the joint distribution of all variables, then infer the corresponding DAG structure. However, this may be computational inefficient, due to the huge number of conditional dependence relations that must be learnt. Finally, we note that, an alternative approach for this step is to separately apply a supervised learning method $B$ times to estimate $\Mean \big\{h_b\big( X_k,\Xbf_{\mathcal{M}-\{k\}} \big) | \Xbf_{\mathcal{M}-\{k\}} \big\}$, for $b=1,\ldots,B$. Nevertheless, when $B$ is large, and in our implementation, $B = 2000$, this approach is computationally very expensive. Therefore, we choose the generative learning approach for this step.

\section{Bidirectional Theory}
\label{sec:theory}

In this section, we establish the asymptotic size and power of the proposed test. As a by-product, we also derive the oracle property of the DAG estimator produced by \eqref{eqn:DAG}, which is needed to guarantee the validity of the test. In the interest of space, we report that result in Section \ref{sec:thmcsl} of the Appendix. To simplify the theoretical analysis, we assume $T_1=\ldots=T_n=T$. All the asymptotic results are derived when either the number of subjects $N$, or the number of time points $T$, diverges to infinity. Such results are new, provide useful theoretical guarantees for different types of applications, and are referred as the bidirectional theory.

We begin with a set of regularity conditions needed for the asymptotic consistency.

\vspace{-0.1in}
\begin{enumerate}[({C}1)]
\setlength\itemsep{-0.25em}
\item With probability approaching one, $\textrm{PA}_j\subseteq \widehat{\textrm{AC}}^{(s)}_j\subseteq \DS_j^c-\{j\}$. 
	
\item Suppose $\Mean \Big | g^{(s)} \Big( \Xbf_{\mathcal{M}^{(s)}} \Big) - \widehat{g}^{(s)} \Big( \Xbf_{\mathcal{M}^{(s)}} \Big) \Big |^2 = O\left\{ (NT)^{-2\kappa_1} \right\}$ for some constant $\kappa_1>0$, and $\widehat{g}^{(s)}$ is uniformly bounded almost surely. Suppose $\Mean \sup_{\widetilde{B} \in \mathcal{B}} \Big| \prob\Big\{ X_k \in \widetilde{B} | \Xbf_{\mathcal{M}^{(s)}} \Big\} - \prob\Big\{ \mathbb{G}^{(s)}\Big( \Xbf_{\mathcal{M}^{(s)}},Z_{j,k}^{(m)} \Big) \in \widetilde{B} |$ $\Xbf_{\mathcal{M}^{(s)}} \Big\} \Big|^2 = O\left\{ (NT)^{-2\kappa_2} \right\}$ for some constant $\kappa_2>0$, where $\mathcal{B}$ denotes the Borel algebra on $\mathbb{R}$. Suppose $\kappa_1+\kappa_2>1/2$. 
	
\item The random process $\{\Xbm_{i,t}\}_{t\ge 0}$ is  $\beta$-mixing if $T$ diverges to infinity. The $\beta$-mixing coefficients $\{\beta(q)\}_{q}$ satisfy that $\sum_q q^{\kappa_3} \beta(q)<+\infty$ for some constant $\kappa_3>0$. Here, $\beta(q)$ denotes the $\beta$-mixing coefficient at lag $q$, which measures the time dependence between the set of variables $\{\Xbm_{i,j}\}_{j\le t}$ and $\{\Xbm_{i,j}\}_{j\ge t+q}$.
	
\item Suppose the number of observations $K$ in the batched standard error estimators $\widehat{\sigma}_{b,\textrm{CF}}^{(s)}$ and $\widehat{\sigma}_{b,\textrm{NCF}}^{(s)}$ satisfies that, $K=T$ if $T$ is bounded, and $T^{(1+\kappa_3)^{-1}} \ll K\ll NT$ otherwise.
\vspace{-0.1in}
\end{enumerate}

Condition (C1) concerns about the step of structural learning of DAG, which essentially requires the order of the DAG can be consistently estimated. We first remark that, this order consistency is much weaker than the selection consistency. In other words, we only require a reasonably good initial DAG estimator that is order consistent, which is much easier to obtain than a DAG estimator that is selection consistent. In Section \ref{sec:thmcsl}, we show that (C1) holds when \eqref{eqn:DAG} is employed to estimate the DAG. Second, (C1) may not be a necessary condition to ensure the type-I error control. We next give two examples, where (C1) does not hold, but our proposed test can still control the type-I error. Moreover, in our simulation examples in Section \ref{sec:sim}, (C1) does not alway hold either. We report the percentage of times out of 500 data replications when (C1) holds for some selected nodes in Section \ref{sec:more-table} of the Appendix. Nevertheless, our test still manages to achieve a competitive empirical performance. On the other hand, we keep (C1) in its current form, as it helps simplify the proof considerably. 

\begin{example}[missing parents]
\tsb{We first consider an example where $\widehat{\textrm{AC}}^{(s)}_j$ misses some nodes in $\PA_j$. The proposed test remains valid as long as these nodes have weak effects on $X_j$ and $X_k$. More specifically, consider the five-variable example as illustrated in Figure \ref{fig:illustration}(c). Our goal is to test whether there is a directed link from $X_3$ to $X_4$. Then $\PA_j\subseteq \widehat{\textrm{AC}}^{(s)}_j$ requires that $\{1,2\}\subseteq \widehat{\textrm{AC}}^{(s)}_4$. Suppose $X_1$ has a weak effect on $X_4$, so that $X_1$ is not included in $\widehat{\textrm{AC}}^{(s)}_4$. Suppose $|\Mean (X_4|X_1,X_2)-\Mean (X_4|X_2)|^2=O\{(NT)^{-2\kappa_1^*}\}$, for some $\kappa_1^*\ge \kappa_1$. When $\Mean \sup_{\widetilde{B} \in \mathcal{B}} |\prob(X_3\in \widetilde{B}|X_2)-\prob(X_3\in \widetilde{B}|X_1,X_2)|=O\{(NT)^{-2\kappa_2^*}\}$, for some $\kappa_2^*\ge \kappa_2$, under (C2)-(C4), the estimated conditional mean function and the distributional generator would converge to $\Mean (X_4|X_1,X_2)$ and $\prob_{X_3|X_1,X_2}$ at the rate of $(NT)^{-\kappa_1}$ and $(NT)^{-\kappa_2}$, respectively. As such, the proposed test still works as if $X_1$ were included in $\widehat{\textrm{AC}}^{(s)}_4$.}
\end{example}	
	
\begin{example}[including descendants]
\tsb{We next consider an example where $\widehat{\textrm{AC}}^{(s)}_j$ includes some nodes in $\DS_j$. The proposed test remains valid as long as none of these nodes is a descendant of $X_k$, or has a common descendant with $X_k$. In this case, $X_k$ and $X_j$ are d-separated given $\widehat{\textrm{AC}}^{(s)}_j$, as none of those falsely included nodes is a collider on any path between $X_j$ and $X_k$; see the definition of d-separation and collider in \citet{pearl2009causal}. As d-separation implies conditional independence, the proposed test is still able to control the type-I error. For the example in Figure \ref{fig:illustration}(c), when $\{5\} \in \widehat{\textrm{AC}}^{(s)}_4$, (C1) is violated. However, when $X_3$ does not have affect $X_5$, the proposed test remains valid.} 
\end{example}		
 
Condition (C2) concerns about the steps of learning the conditional mean function and the distribution generator. It requires the squared prediction loss of the supervised learner of the conditional mean, and the squared total variation norm between the conditional distributions of the observed and pseudo samples to satisfy some convergence rate, $\kappa_1$ and $\kappa_2$, respectively. We note that both estimators are nonparametric, and as such, both $\kappa_1$ and $\kappa_2$ can be slower than the parametric rate of $1/2$. However, (C2) only requires that $\kappa_1 + \kappa_2 > 1/2$. This is relatively easy to achieve when using the multilayer perceptron models and GANs, whose convergence rates have been established \citep[see e.g.,][]{schmidt2017nonparametric, farrell2018deep, liang2018well, Bauer2019, chen2020statistical}. Moreover, we remark that, it is possible to further relax the requirement of $\kappa_1 + \kappa_2 > 1/2$ to $\kappa_1,\kappa_2>0$, by using the theory of higher order influence functions \citep{robins2017minimax}. However, the corresponding estimators would be considerably much more complicated, and thus we do not pursue those in this article.

Condition (C3) characterizes the dependence of the data observations over time, and is commonly imposed in the time series literature \citep{Bradley2005}. We also note that, (C3) is \emph{not} needed when $T$ is bounded but $N$ diverges to infinity. Condition (C4) guarantees the consistency of the batched standard error estimators $\widehat{\sigma}_{b,\textrm{CF}}^{(s)}$ and $\widehat{\sigma}_{b,\textrm{NCF}}^{(s)}$, and is easily satisfied, since $K$ is a parameter we specify. When $T$ is bounded and is relatively small compared to a large sample size $N$, we can simply set $K=T$, i.e., treating the entire time series as one batch.

We next establish the asymptotic size of the propose testing procedure. 

\begin{theorem}[Size]\label{thm2}
Suppose model \eqref{eqn:model}, and conditions (C1)-(C4) hold. Suppose $\min_b NT$ $\Var\Big( \widehat{I}^{(s)}_{b,\textrm{CF}} | \{\Xbm_{i,t}\}_{i\in \mathcal{I}_s,1\le t\le T} \Big)$ $\ge \kappa_4$ for some constant $\kappa_4>0$. If the constants $\kappa_1$, $\kappa_2$, $\kappa_3$ satisfy that  $\kappa_3>\max[\{2\min(\kappa_1,\kappa_2)\}^{-1}-1,2]$, then, as either $N$ or $T\to \infty$,
\begin{enumerate}[(a)]
\item The test statistic $\widehat{T}^{(s)}_{\widehat{b}^{(s)},\tsb{\rm{CF}}}\stackrel{d}{\to} \textrm{Normal}(0,1)$ under $H_0(j,k)$.
\item The $p$-value satisfies that $\prob\{p(j,k)\le \alpha\}\le \alpha+o(1)$, for any nominal level $0<\alpha<1$.
\end{enumerate}
\end{theorem}

\noindent
To establish the asymptotic size of the test, we require $\beta(q)$ to decay at a polynomial rate with respect to $q$. Such a condition holds for many common time series models \citep[see, e.g.,][]{McDonald2015est}. We also require a minimum variance condition, which automatically holds when the conditional variance of $h_b^{(s)}\big( X_k, X_{\mathcal{M}^{(s)}} \big) - \Mean \big\{h_b^{(s)}\big( X_k, X_{\mathcal{M}^{(s)}} \big) | X_{\mathcal{M}^{(s)}} \big\}$ given $X_{\mathcal{M}^{(s)}}$ is bounded away from zero. Under these conditions, we establish the asymptotic normality of the test statistic $\widehat{T}^{(s)}_{\widehat{b}^{(s)},\tsb{\rm{CF}}}$, which further implies that the $p$-value $p^{(s)}(j,k)$ converges to a uniform distribution on $[0,1]$. By Bonferroni's inequality, $p(j,k)$ is a valid $p$-value, and consequently, the proposed test achieves a valid control of type-I error.

Next, we study the asymptotic power of the test. We introduce a quantity to characterize the degree to which the alternative hypothesis deviates from the null for a given function class $\mathbb{H}$: $\Delta(\mathbb{H}) = \min_{\mathcal{M}}\sup_{h\in \mathbb{H}} |I(j,k|\mathcal{M};h)|$, where the minimum is taken over all subsets $\mathcal{M}$ that satisfy the conditions in Proposition \ref{prop1}. When $\mathbb{H}$ is taken over the class of characteristic functions of $(X_k,X_{\mathcal{M}})$, we have $\Delta(\mathbb{H})>0$. We also need the concept of the VC type class \citep[][Definition 2.1]{Cherno2014}; see Section \ref{sec:proofthm3} of the Appendix. To simplify the analysis, we suppose $X_j$ is bounded, and without loss of generality, its support is $[0,1]$.

\begin{theorem}[Power]\label{thm3}
Suppose the conditions in Theorem \ref{thm2} hold, and the $\beta$-mixing coefficient $\beta(q)$ in (C3) satisfies that $\beta(q)=O(\kappa_5^q)$ for some constant $0<\kappa_5<1$ when $T$ diverges. Suppose $\Delta(\mathbb{H})\gg (NT)^{-1/2}\log (NT)$ under $H_1(j,k)$. Suppose, with probability tending to one, $\widehat{g}^{(s)}$ and $\mathbb{G}^{(s)}$ belong to the class of VC type functions with bounded envelope functions and the bounded VC indices no greater than $O\{(NT)^{\min(2\kappa_1,2\kappa_2,1/2)}\}$, $s=1,2$. If the number of transformation functions $B=\kappa_6 (NT)^{\kappa_7}$ for some constants $\kappa_6 > 0, \kappa_7 \ge 1/2$, then, as either $N$ or $T\to \infty$, $p(j,k)\stackrel{p}{\to} 0$ under $H_1(j,k)$.
\end{theorem}

\noindent
To establish the asymptotic power of the test, we require the function $\widehat{g}^{(s)}$ and the generator $\mathbb{G}^{(s)}$ to both belong to the VC type class. This is to help establish the concentration inequalities for the measure $\widehat{I}^{(s)}_{b,\textrm{NCF}}$ without cross-fitting. This condition automatically holds in our implementation where the MLP is used to model both \citep{farrell2018deep}. We have also strengthened the requirement on $\beta(q)$, so that it decays exponentially with respect to $q$. This is to ensure the $\sqrt{NT}$-consistency of the proposed test when $T\to \infty$. This condition holds when the process $\{\Xbm_{i,t}\}_{t\ge 0}$ forms a recurrent Markov chain with a finite state space. It also holds for more general state space Markov chains \citep[see, e.g.,][Section 3]{Bradley2005}. Under these conditions, Theorem \ref{thm3} shows that our proposed test is consistent against some local alternatives that are $\sqrt{NT}$-consistent to the null up to some logarithmic term.

We remark that, Theorems \ref{thm2} and \ref{thm3} show that the proposed test controls the type-I error and achieves a parametric power guarantee, even though we estimate the three key components, the DAG structure, the conditional mean, and the distribution generator, all using fully nonparametric methods. This is achieved mainly due to the fact that our test statistic $\widehat{T}_{\widehat{b}^{(s)},\textrm{CF}}^{(s)}$ is doubly robust, in that it is consistent as long as either the conditional mean or the distribution generator is correctly specified. Together with the Neyman orthogonality of the estimating equation, we show that the bias can be represented as a product of the difference between the two nonparametric estimators and their oracle values; see Step 3 of the proof of Theorem \ref{thm2} in Section \ref{sec:proof-thm2} of the Appendix. Consequently, as long as $\kappa_1 + \kappa_2 > 1/2$, the test statistic converges at a parametric rate, and the test has a parametric power guarantee.

We also remark that, in our theory, the dimension $d$ of the DAG is allowed to diverge to infinity with the sample size. Note that there is no explicit specification on $d$ in the statements of Theorems \ref{thm2} and \ref{thm3}. It is implicitly imposed due to the requirement that $\kappa_1+\kappa_2>1/2$, as the convergence rates would become slower as the dimension $d$ increases.

\section{Simulations}
\label{sec:sim}

In this section, we examine the finite-sample performance of the proposed testing procedure. 

We begin with a discussion of some implementation details. Our test employs three neural networks-based learners, which involve numerous tuning parameters. Many of these parameters are common, e.g., the number of hidden layers and hidden nodes, the activation function, batch size, and epoch size, and we set them at the typical values recommended in the literature. For the DAG learning step, one tuning parameter is the sparsity parameter $\lambda$ in \eqref{eqn:DAG}. Following \cite{zheng2020learning}, we fix $\lambda=0.025$ in our implementation to speed up the computation. We have also experimented with a number of values of $\lambda$ and find the results are not overly sensitive. It can also be tuned via cross-validation. For the supervised learning step, we employ the multilayer perceptron regressor implementation of \citet{SCIKit2011}. For the GANs training step, we follow the implementation of \citet{genevay2017learning}. There are three additional parameters associated with our test, including the number of transformation functions $B$, the number of pseudo samples $M$, and the number of observations $K$ in the batched standard error estimators. We have found that the results are not sensitive to the choice of $M$ and $K$, and we fix $M=100$ and $K=20$. For $B$, a larger value generally improves the power of the test, but also increases the computational cost. In our implementation, we set $B=2000$, which achieves a reasonable balance between the test accuracy and the  computational cost.

We compare the proposed test with two alternative solutions, the double regression-based  test (DRT) as outlined in Section \ref{sec:equiv}, and the constrained likelihood ratio test (LRT) proposed by \citet{li2019likelihood} for linear DAGs. The implementation of DRT is similar to our proposed method. The main difference lies in that DRT uses the MLP regressor to first estimate the conditional mean function $\Mean (X_k|X_{\mathcal{M}^{(j)}_{j,k}})$ in Step 4, then plugs in this estimate to construct the test statistic in Step 5, with $B=1$ and $h_1^{(s)}(X_k,X_{\mathcal{M}^{(j)}_{j,k}})=X_k$.

We consider the following nonlinear DAG model,
\begin{align} \label{eqn:sim-nonlinear}
X_{t,j} & = \sum_{\substack{k_1,k_2 \in \scriptsize{\PA}_{j}\\ k_1 \leq k_2}}c_{j,k_1,k_2}f^{(1)}_{j,k_1,k_2}(X_{t,k_1})f^{(2)}_{j,k_1,k_2}(X_{t,k_2}) + \sum_{k_3 \in \scriptsize{\PA}_{j}}c_{j,k_3}f_{j,k_3}^{(3)}(X_{t,k_3}) + \varepsilon_{t,j}.
\end{align}
The data generation follows that of \cite{zhu2020causal}. Specifically, $f^{(1)}_{j,k_1,k_2}$, $f^{(2)}_{j,k_1,k_2}$, and $f_{j,k_3}^{(3)}$ in \eqref{eqn:sim-nonlinear} are randomly set to be sine or cosine function with equal probability, whereas $c_{j,k_1,k_2}$ and $c_{j,k_3}$ are randomly generated from uniform $[0.5 \delta,1.5\delta]$ or $[-1.5\delta,-0.5\delta]$ with an equal probability, where $\delta>0$ denotes some constant that controls the signal strength. The error $\varepsilon_{t,j}$ is an AR(1) process with the autoregressive coefficient equal to 0.5 and a standard normal white noise. The DAG structure is determined by a $d \times d$ lower triangular binary adjacency matrix, in which each entry is randomly sampled from a Bernoulli distribution with probability $\zeta$. We vary four sets of key parameters in our simulations: (a) the number of subjects $N$ from $\{ 10, 20, 40 \}$; (b) the number of time points $T$ from $\{ 50, 100, 200 \}$; (c) the signal strength $\delta$ from $\{ 0.5, 1, 2 \}$, and (d) the dimension $d$ and the Bernoulli probability $\zeta$ from $(d, \zeta) = \{(50, 0.10), (100, 0.04), (150, 0.02) \}$. When we vary one set of the parameters, we keep the rest fixed at their default values of $N=20, T=100, \delta=1, d=50, \zeta=0.10$. 

\begin{figure}[!t]
\centering
\includegraphics[width=15.25cm,height=15.75cm]{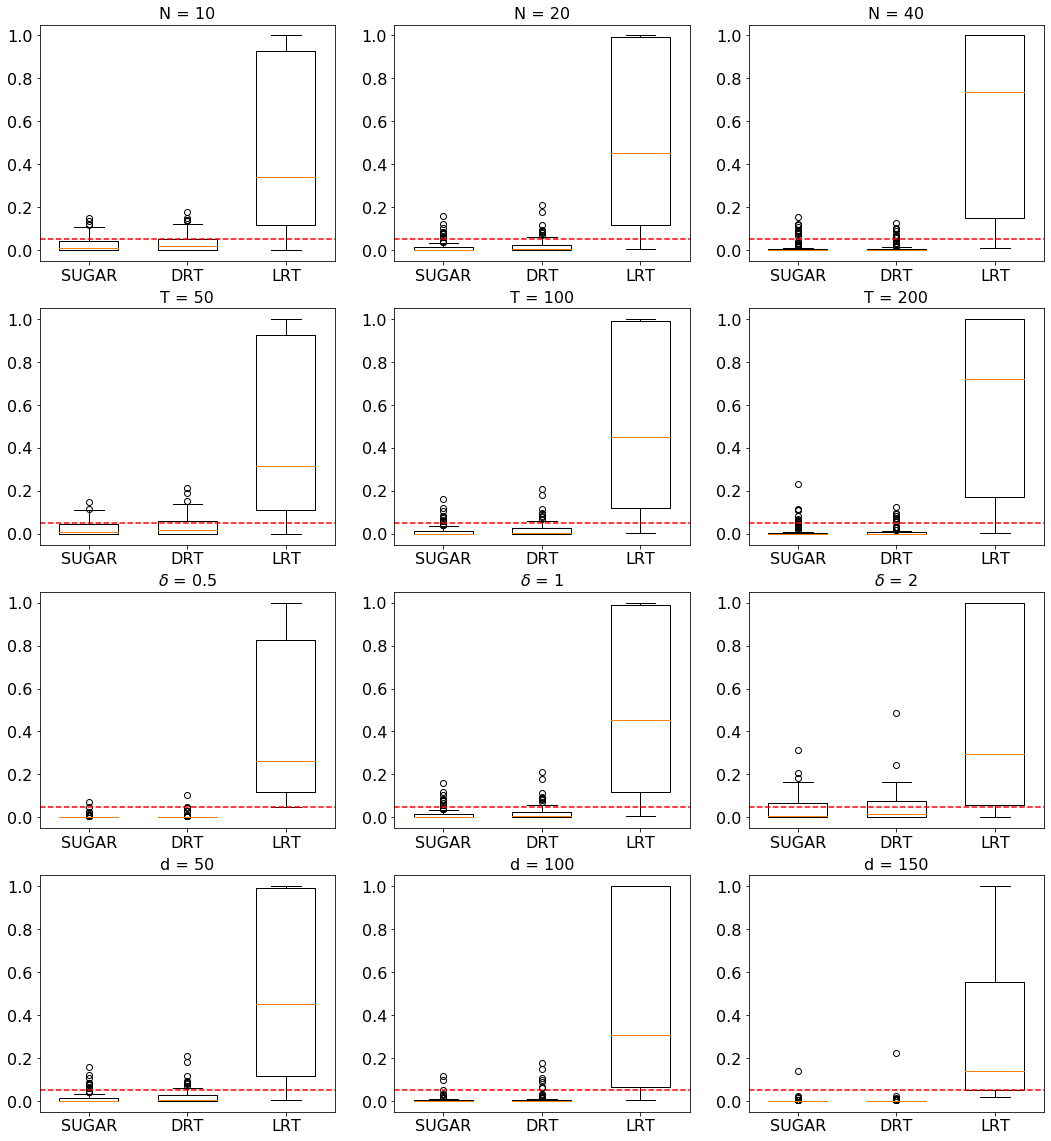}
\caption{The boxplots of the empirical size of three methods: our proposed test (SUGAR), the double regression-based test (DRT), and the constrained likelihood ratio test (LRT), under four sets of varying parameters: first row $N = \{ 10, 20, 40 \}$, second row $T = \{ 50, 100, 200 \}$, third row $\delta = \{ 0.5, 1, 2 \}$, and fourth row $(d, \zeta) = \{(50, 0.10), (100, 0.04), (150, 0.02) \}$.}
\label{fig:sim_null}
\end{figure}

\begin{figure}[!t]
\centering
\includegraphics[width=15.25cm,height=15.75cm]{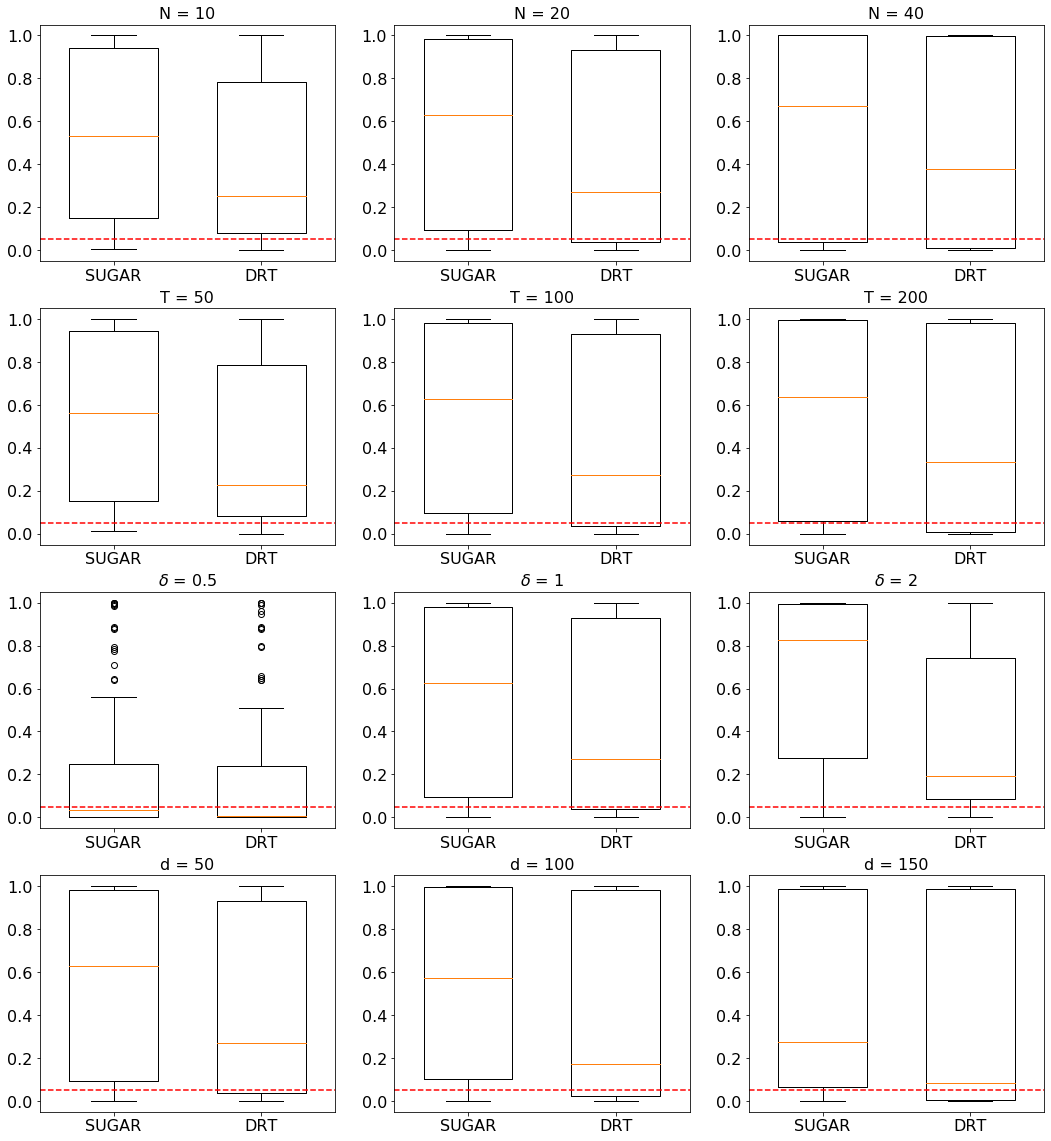}
\caption{The boxplots of the empirical power of two methods: our proposed test (SUGAR), and the double regression-based test (DRT), under four sets of varying parameters: first row $N = \{ 10, 20, 40 \}$, second row $T = \{ 50, 100, 200 \}$, third row $\delta = \{ 0.5, 1, 2 \}$, and fourth row $(d, \zeta) = \{(50, 0.10), (100, 0.04), (150, 0.02) \}$.}
\label{fig:sim_alter}
\end{figure}

For each scenario, we randomly sample 100 pairs of nodes where the null hypothesis holds, and another 100 pairs of nodes where the alternative hypothesis holds. We then apply the proposed test to these pairs, and record the empirical size and power of the test, i.e., the percentage of the times out of 200 data replications when the $p$-value is smaller than the nominal level $\alpha=0.05$. Figure \ref{fig:sim_null} shows the boxplots of the empirical size for the pairs when the null holds, and Figure \ref{fig:sim_alter} shows the boxplots of the empirical power for the pairs when the alternative holds. We further report the difference of the powers of SUGAR and DRT in Figure \ref{fig:sim_diff} in Section \ref{sec:more-power} of the Appendix. We do not report the power of LRT, because it fails to control the type-I error, and thus its empirical power becomes meaningless. We make the following observations from these plots. In terms of the empirical size, both SUGAR and DRT manage to control the type-I error, but LRT does not. The reason is that LRT requires the graph to have a linear structure and the samples to be independent, but none is satisfied in our simulation model. On the other hand, in terms of the empirical power, SUGAR achieves generally a higher power than DRT, over 75\% of the times in all scenarios as seen from Figure \ref{fig:sim_diff}. Finally, as the key model parameters vary, the power of both SUGAR and DRT increases as the number of subjects $N$, or the number of time points $T$ increases, since more data information becomes available, and the power of both tests decreases as the dimension $d$ increases, since the graph becomes bigger and the problem more challenging. Meanwhile, the power of SUGAR increases as the signal strength $\delta$ increases, but that of DRT is not monotonic with respect to $\delta$, because DRT is not guaranteed to be consistent in general, as we have commented earlier. 

In terms of the computational time, our testing procedure consists of two main parts: the DAG estimation in Step 2 of Algorithm \ref{alg:full}, and the rest in Steps 3 to 6. The DAG estimation is the most time consuming step, but it only needs to be learnt once for all pairs of edges in the graph. We implemented the DAG estimation step on the NVIDIA Tesla T4 GPU, and it took about 5 to 20 minutes when $d$ ranges from 50 to 150 for one data replication. We implemented the rest of the testing procedure on the N1 standard CPU, and it took about 2 minutes for one data replication. A \proglang{Python} implementation of our method is available at \url{https://github.com/yunzhe-zhou/SUGAR}.

\section{Brain Effective Connectivity Analysis}
\label{sec:realdata}

We next illustrate our method with a brain effective connectivity analysis of task-evoked functional magnetic resonance imaging (fMRI) data. The brain is a highly interconnected dynamic system, and it is of great interest to understand the relations among different brain regions through fMRI, which measures synchronized blood oxygen level dependent brain signals. The dataset we analyze is part of the Human Connectome Project \citep[HCP,][]{van2013wu}, whose overarching objective is to understand brain connectivity patterns of healthy adults. We study the fMRI scans of a group of individuals who undertook a story-math task. The task consisted of blocks of auditory stories and addition-subtraction calculations, and required the participant to answer a series of questions. An accuracy score was given at the end. We analyze two subsets of individuals with matching age and sex. One set consists of $N = 28$ individuals who scored below 65 out of 100, and the other set consists of $N = 28$ individuals who achieved the perfect score of 100.  All fMRI scans have been preprocessed following the pipeline of \citet{glasser2013minimal} that summarized each fMRI scan as a matrix of time series. Each row is a time series with length $T = 316$, and there are 264 rows corresponding to 264 brain regions \citep{power2011functional}. Those brain regions are further grouped into 14 functional modules \citep{Smith2009}. Each module possesses a relatively autonomous functionality, and complex tasks are believed to perform through coordinated collaborations among the modules. In our analysis, we concentrate on $d = 127$ brain regions from four functional modules: auditory, visual, frontoparietal task control, and default mode, which are generally believed to be involved in language processing and problem solving domains \citep{Barch2013}. 

\begin{table}[t!]
\centering
\caption{The number of identified significant within-module and between-module connections of the four functional modules for the low-performance and high-performance groups. The number of brain regions of each functional module is reported in the parenthesis.}
\label{tab:HCP}
\renewcommand\arraystretch{1.8}
\setlength{\tabcolsep}{2.1mm}{
\begin{tabular}{|c|c|c|c|c|c|c|c|c|c|} \hline
\multicolumn{2}{|c|}{\parbox{2.0cm}{\centering  \quad}} & \multicolumn{2}{c|}{\parbox{2.5cm}{\centering Auditory \quad (13)}} & \multicolumn{2}{c|}{\parbox{2.5cm}{\centering Default mode \quad (58)}} & \multicolumn{2}{c|}{\parbox{2.5cm}{\centering Visual \qquad (31)}} & \multicolumn{2}{c|}{\parbox{2.6cm}{\centering Fronto-parietal\quad (25)}}\\ \hline
\multicolumn{2}{|c|}{\parbox{2.0cm}{}}  & low & high & low & high & low & high & low & high \\ \hline
\multicolumn{2}{|c|}{\parbox{2.5cm}{\centering Auditory \quad (13)}} & \parbox{0.9cm}{\centering 20}  & 17 & \parbox{0.9cm}{\centering 0} & 0 & \parbox{0.9cm}{\centering 0} & 1 & \parbox{0.9cm}{\centering 2} & 0 \\ \hline
\multicolumn{2}{|c|}{\parbox{2.5cm}{\centering Default mode \quad (58)}} & 0 & 0 & 68 & 46 & 3 & 2 & 11 & 23 \\ \hline
\multicolumn{2}{|c|}{\parbox{2.5cm}{\centering Visual \qquad (31)}} & 0 & 0 & 3 & 2 & 56 & 46 & 0 & 1 \\ \hline
\multicolumn{2}{|c|}{\parbox{2.6cm}{\centering Fronto-parietal\quad (25)}} & 2 & 1 & 11 & 23 & 0 & 1 & 22 & 27 \\ \hline
\end{tabular}}
\end{table}

We apply the proposed test to the two datasets separately. We control the false discovery at $0.05$ using the standard Benjamini-Hochberg procedure \citep{BH1995}. Table \ref{tab:HCP} reports the number of identified significant within-module and between-module connections. We first note that, we identify many more within-module connections than the between-module connections. The partition of the brain regions into the functional modules has been fully based on the biological knowledge, and our finding lends some numerical support to this partition. In addition, we identify more within-module connections for the frontoparietal task control module for the high-performance subjects than the low-performance subjects, while we have identified fewer within-module connections for the default mode and visual modules for the high-performance subjects. These findings generally agree with the neuroscience literature. Particularly, the frontoparietal network is known to be involved in sustained attention, complex problem solving and working memory \citep{Menon2011}, and the high-performance group exhibits more active connections for this module. Meanwhile, the default mode network is more active during passive rest and mind-wandering, which usually involves remembering the past or envisioning the future rather than the task being performed \citep{van2017mind}, and the high-performance group exhibits fewer active connections for this module.

\section*{Acknowledgement}
Li's research was partially supported by NSF grant CIF-2102227, and NIH grants R01AG061303, and R01AG062542. Shi's research was partially supported by EPSRC grant EP/W014971/1.

\baselineskip=21pt
\bibliographystyle{apa}
\bibliography{ref-sugar}

\newpage
\appendix
In this appendix, Section \ref{sec:extensions} discusses several extensions of the proposed test. Section \ref{sec:add-results} presents additional theoretical and numerical results. Section \ref{sec:proofs} gives the detailed proofs.

\section{Extensions}
\label{sec:extensions}

In the article, we have primarily focused on testing a particular pair of nodes $(j,k)$ in the DAG model, $j, k = 1, \ldots, d$. Next, we discuss the extensions to test a directed pathway, a union of directed edges, and the categorical $X_j$ following a generalized linear model. We also outline the extensions to the Markov equivalence class, and non-stationary and time-varying DAG.

\subsection{Extension to a directed pathway}
\label{sec:ext-path}

Suppose our goal is to test a given directed pathway, $j_1\rightarrow j_2 \rightarrow \ldots \rightarrow j_K$, where $j_1,j_2,\ldots,j_K$ are a sequence of nodes in the DAG. The problem can be formulated as the pair of hypotheses:
\begin{eqnarray}\label{eqn:hypo1}
	\begin{split}
		H_{p0}: \;\; & H_0(j_k,j_{k+1})\,\,\textrm{holds~for~some}~k,\,\,\,\,\,\textrm{versus}\\
		H_{p1}: \;\; & H_0(j_k,j_{k+1})\,\,\textrm{does~\textit{not}~hold~for~any}~k = 1, \ldots, d.
	\end{split}	
\end{eqnarray}
Under the alternative, each individual null hypothesis $H_0(j_k,j_{k+1})$ does not hold, and thus there exists such a directed pathway. The hypotheses in \eqref{eqn:hypo1} can be tested using the union-intersection principle. Specifically, let $p(j_k,j_{k+1})$ denote the $p$-value for $H_0(j_k,j_{k+1})$ from the proposed test. Then it is straightforward to show that $\max_k p(j_k,j_{k+1})$ is a valid $p$-value for \eqref{eqn:hypo1}. Based on Theorems \ref{thm2} and \ref{thm3}, we can also show that such a test is consistent.

\subsection{Extension to a union of directed edges}
\label{sec:ext-union}

Suppose our goal is to test a union of the hypotheses $\cup_{l \in \mathcal{L}} H_0(j_l,k_l)$. We first apply the proposed test to construct two standardized measures, $\widehat{T}^{(s)}_{b,\textrm{CF}}(j_l,k_l)$ and $\widehat{T}^{(s)}_{b,\textrm{NCF}}(j_l,k_l)$, with and without cross-validation, for each $b=1,\ldots,B, s=1,2$, and $l \in \mathcal{L}$. Then for each $s$, we select the indices $\widehat{b}^{(s)}$ and $\widehat{l}^{(s)}$ that yield the largest measure $\max_{b,l}|\widehat{T}^{(s)}_{b,\textrm{NCF}}(j_l,k_l)|$ in the absolute value. We then construct the Wald type test statistic $\widehat{T}^{(s)}_{\widehat{b}^{(s)},\textrm{CF}}\left( j_{\widehat{l}^{(s)}},k_{\widehat{l}^{(s)}} \right)$. Based on Theorems \ref{thm2} and \ref{thm3}, we can establish the consistency of this test.

\subsection{Extension to generalized linear model}
\label{sec:ext-glm}

We can further extend the proposed test to the following class of models:
\begin{eqnarray*}
	\Mean (X_j|\Xbf_{\scriptsize{\PA}_j}) = \phi_j\left\{ f_j(\Xbf_{\scriptsize{\PA}_j}) \right\}, \quad \textrm{ for any } j=1, \ldots, d,
\end{eqnarray*}
where the link function $\phi_j$ is pre-specified while the function $f_j$ is unspecified. For instance, when $X_j$ is binary, we may set $\phi_j$ as the logistic function. Similar to Theorem \ref{thm1}, we can show that the null hypothesis in \eqref{eqn:hypothesis2} is equivalent to $I(j,k|\mathcal{M};h)=0$, for all square-integrable function $h$. Therefore, the proposed test can be applied to this class of models as well.

\subsection{Extension to Markov equivalence class}
\label{sec:ext-markov}

In the article, we have mainly focused on the case when the underlying DAG is identifiable. In this section, we discuss the extension to the Markov equivalence class. We first outline the key steps of the extension, then consider a way to expedite the computation. We further discuss the relation between our test and the DAGs in the equivalence class. Meanwhile, we leave the full investigation of the inference for the equivalence class as future research.

\bigskip
\noindent
\textbf{Outline of the extension}:
Suppose there exists an equivalence class of DAGs that could generate the same joint distribution of the variables. Such a class can be uniquely represented by a completed partially directed acyclic graph (CPDAG). For each DAG $\mathcal{G}$ that belongs to the equivalence class, we define $\PA_j(\mathcal{G})$ as the set of parents of node $j$ in $\mathcal{G}$. Then, we aim to test the hypotheses:
\begin{eqnarray} \label{eqn:hypo-equivclass}
	\begin{split}
		H_{e0}(j,k): \;\; & k \notin \PA_j(\mathcal{G}), \,\,\,\,\,\, \textrm{ versus }\\
		H_{e1}(j,k): \;\; & k \in \PA_j(\mathcal{G}'),\,\,\,\, \textrm{ for some }\mathcal{G}'~\textrm{that~belongs~to~the~equivalence~class}.
	\end{split}
\end{eqnarray}

To test the hypotheses in \eqref{eqn:hypo-equivclass}, we first estimate the equivalence class given each half of the data. Next, for each DAG $\mathcal{G}$ that belongs to the estimated equivalence class, we employ supervised learning and generative adversarial learning to compute the standardized measures, $\{\widehat{T}_{b,\textrm{CF}}^{(s)}(\mathcal{G})\}_{b=1}^{B}$, and $\{\widehat{T}_{b,\textrm{NCF}}^{(s)}(\mathcal{G})\}_{b=1}^{B}$. We then select the index $(\widehat{b}^{(s)},\widehat{\mathcal{G}}^{(s)})$ that maximizes $|\widehat{T}_{b,\textrm{NCF}}^{(s)}(\mathcal{G})|$, and take $|\widehat{T}_{\widehat{b}^{(s)},\textrm{NCF}}^{(s)}(\widehat{\mathcal{G}}^{(s)})|$ as the final test statistic. Finally, we compute the $p$-value as $p(j,k)=2\min \Big[$ $\Phi\left\{ Z_0>|\widehat{T}_{\widehat{b}^{(1)},\textrm{NCF}}^{(1)}(\widehat{\mathcal{G}}^{(1)})| \right\}, \Phi\left\{ Z_0>|\widehat{T}_{\widehat{b}^{(2)},\textrm{NCF}}^{(2)}(\widehat{\mathcal{G}}^{(2)})| \right\} \Big]$, where $Z_0$ is a standard normal variable. This testing procedure is similar as Algorithm \ref{alg:full}, except that the index is now selected among all possible pairs of $(b, \mathcal{G})$, whereas the index is selected among $b$ only in Algorithm \ref{alg:full}.

We can show the above test is consistent, following a similar approach as the test for an identifiable DAG in Section \ref{sec:theory}. We remark that, to establish the type-I error control, we only require each DAG estimator in the estimated equivalence class to be order consistent to some DAG in the true equivalence class. By contrast, to establish the power guarantee, we further require a one-to-one correspondence between the estimated and the true equivalence class.

\bigskip
\noindent
\textbf{Computation acceleration}:
When the graph is large, we recognize that it is computationally intensive to enumerate \emph{all} the DAGs within the equivalence class. To accelerate the computation, we propose to focus on those DAGs that are only ``locally" different. 

Specifically, we first observe that our proposed algorithm depends on the estimated DAG $\mathcal{G}$ only through the index set $\mathcal{M}=\widehat{\AC}_j(\mathcal{G})-\{k\}$. As such, we can speed up the computation by directly calculating the multi-set of the ancestor sets,
\begin{eqnarray*}
	\widetilde{\AC}_{j,k} = \left\{ \widehat{\AC}_j(\mathcal{G}) : \mathcal{G} \; \textrm{that belongs to the equivalence class and} \; k \in \widehat{\AC}_j(\mathcal{G}) \right\}.
\end{eqnarray*}

Moreover, for a graph $\mathcal{G}$, denote a subset of its estimated ancestor set $\widehat{\AC}_j(\mathcal{G})$ up to $G$ generations by $\widehat{\AC}_j^{(G)}(\mathcal{G})$. For instance, $\widehat{\AC}_j^{(1)}(\mathcal{G})$ denotes all the estimated parent nodes, and $\widehat{\AC}_j^{(2)}(\mathcal{G})$ denotes all the estimated parent and grandparent nodes. Along with some other mild conditions, if the following condition holds, 
\begin{equation} \label{eqn:require-consistency}
	\textrm{PA}_j(\mathcal{G}) \subseteq \widehat{\AC}_j^{(G)}(\mathcal{G}), 
\end{equation}
then the corresponding test remains to be consistent. On the other hand, while the ancestor sets of two DAGs may not be completely the same, their ancestor sets up to certain  generations, e.g., the parent sets or the grandparent sets, may be the same. This motivates us to consider the following multi-set to further speed up the computation, 
\begin{eqnarray*}
	\widetilde{\AC}_{j,k}^{(G)} = \left\{ \widehat{\AC}_j^{(G)}(\mathcal{G}): \mathcal{G} \; \textrm{that belongs to the equivalence class and} \; k \in \widehat{\AC}_j^{(G)}(\mathcal{G}) \right\}.
\end{eqnarray*}
Correspondingly, the number of elements in $\widetilde{\AC}_{j,k}^{(G)}$ can potentially be much smaller than that of $\widetilde{\AC}_{j,k}$. In other words, we focus on the ancestors of node $j$ for the graphs in the equivalence class up to $G$ generations only, instead of all the generations. Here $G$ represents a trade-off between the computational cost and the sufficient condition to ensure the consistency of the test. When $G$ is large, it is easier for the condition \eqref{eqn:require-consistency} to hold, but it is computationally more expensive. When $G$ is small, it is harder for \eqref{eqn:require-consistency} to hold, but it allows us to focus on the DAGs that are only ``locally" different around the link $(j,k)$, and thus accelerates the computation.

To implement the above idea, we first use each half of the data to obtain a 
CPDAG. This can be achieved by directly applying some existing structural learning method, e.g., the PC algorithm \citep{Spirtes2000}, or by first applying the method in Section \ref{sec:CSL}, then converting the learnt DAG to a CPDAG \citep{kalisch2021pcalg}. Next, based on the estimated CPDAG, we select those nodes that are ancestors of $j$ up to $G$ generations. Let $\mathcal{N}^{(G)}$ denote these nodes. We then apply Algorithm 3 of \citet{nandy2017estimating} to obtain the multi-set of the parent sets of $\mathcal{N}^{(G)}\cup \{j\}$, 
\begin{eqnarray*} 
	\bigg\{ \left\{ \widehat{\PA}_l(\mathcal{G}) : l \in \mathcal{N}^{(G)}\cup \{j\} \right\} : \mathcal{G} \ \textrm{that belongs to the equivalence class and} \ k \in \widehat{\AC}_j^{(G)}(\mathcal{G}) \bigg\}.
\end{eqnarray*}
For each $\mathcal{G}$, the parent set of $\mathcal{N}^{(G)}\cup \{j\}$, i.e., $\left\{ \widehat{\PA}_l(\mathcal{G}) : l \in \mathcal{N}^{(G)} \right\}$ essentially contains all parents for each node in $\mathcal{N}^{(G)}\cup \{j\}$, based on which we can derive $\widehat{\AC}_j^{(G)}(\mathcal{G})$, and subsequently $\widetilde{\AC}_{j,k}^{(G)}$. \citet{nandy2017estimating} and \citet{chakrabortty2018inference} noted that it is much more computationally efficient to obtain the multi-set than to enumerate all DAGs.

\bigskip
\noindent
\textbf{Equivalence class}:
We remark that our proposed test is built upon testing the conditional independence, and can test if a link exists in a DAG in an equivalence class. However, our test is generally not able to distinguish different DAGs in an equivalence class. We consider the following example to further elaborate. 
	
\begin{figure}[t!]
		\centering
		\begin{tabular}{ccccc}
			\includegraphics[width=3.5cm]{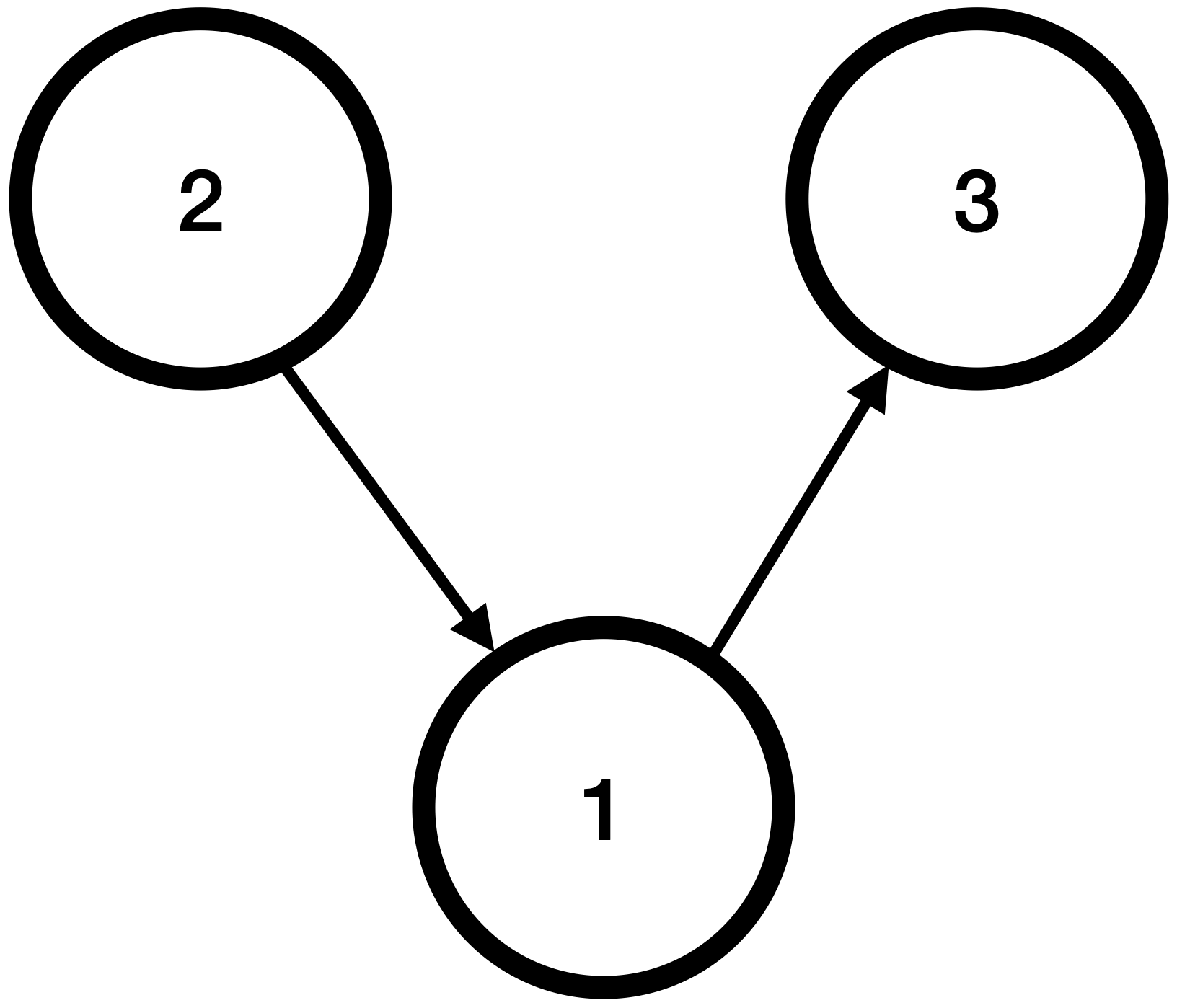} & \hspace{0.25in} &
			\includegraphics[width=3.5cm]{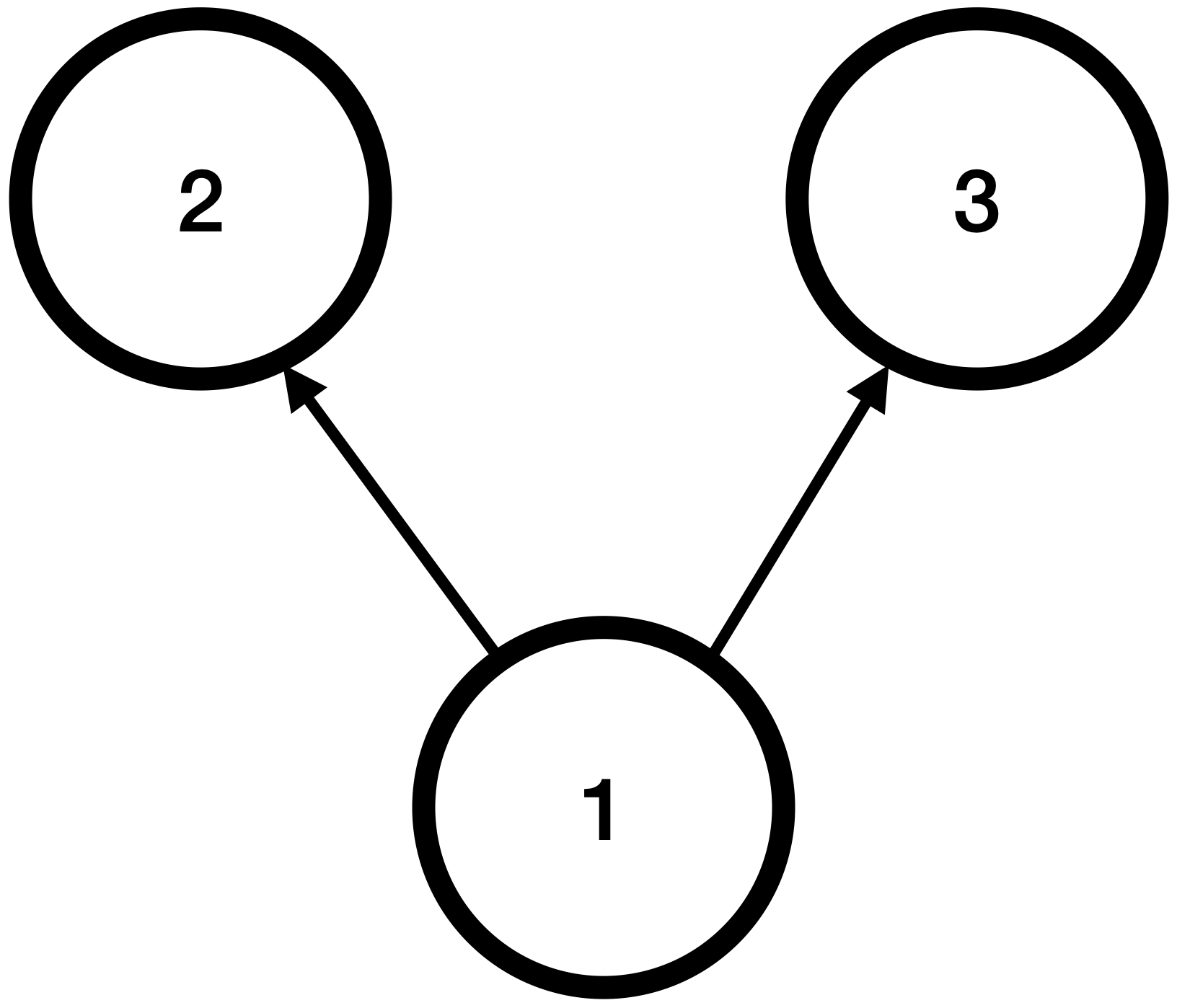} & \hspace{0.25in} &
			\includegraphics[width=3.5cm]{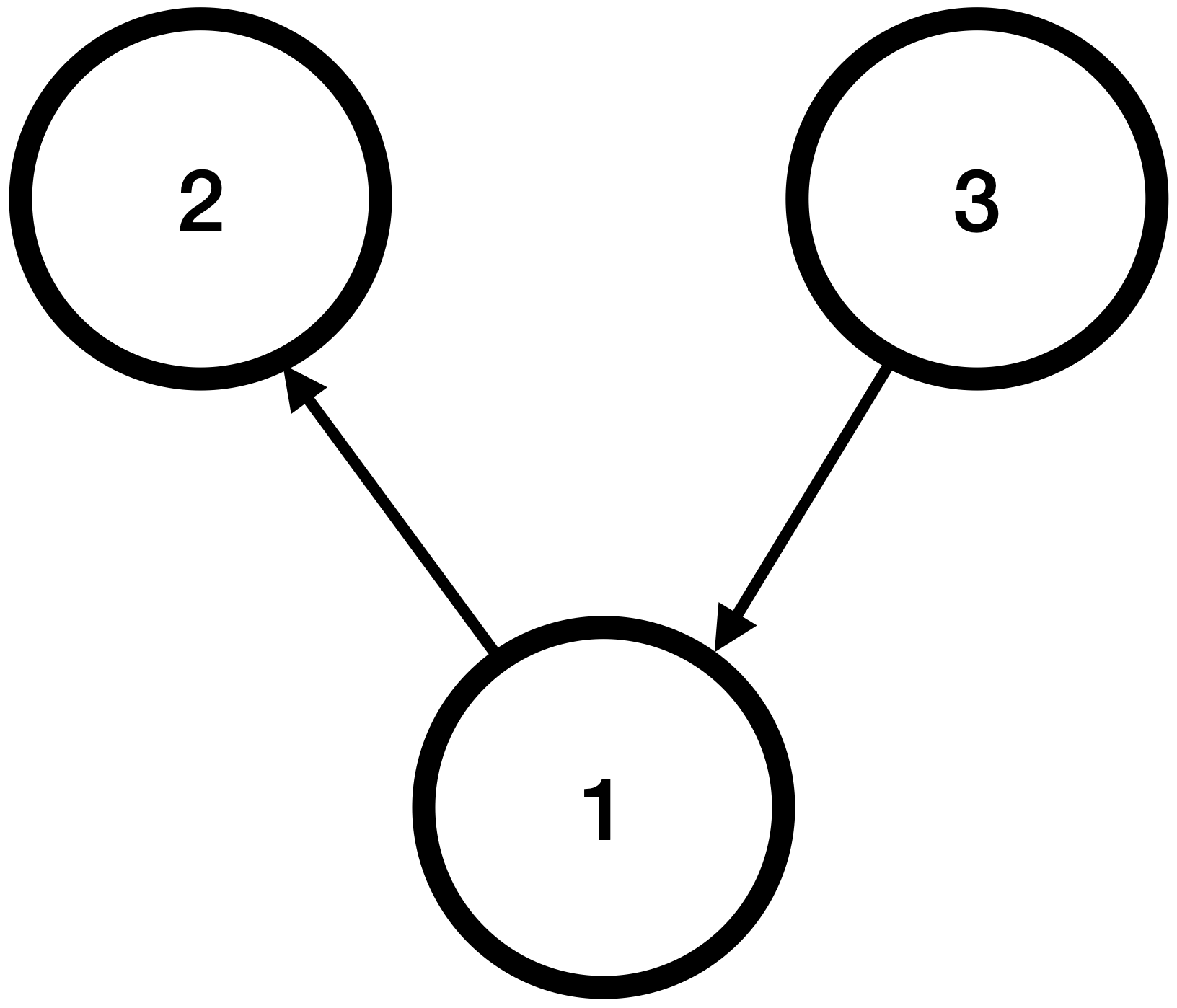}\\
			$\mathcal{G}_1$ & & $\mathcal{G}_2$ & & $\mathcal{G}_3$
		\end{tabular}
		\caption{Three DAGs that belong to the same equivalence class, and each with three variables.}
		\label{fig:equi}
	\end{figure}

	\begin{example}[Equivalence class]
		\tsb{Consider three DAGs depicted in Figure \ref{fig:equi}. All three DAGs have the same skeleton, none has colliders, and thus they belong to the same equivalence class following \citet{verma1990equivalence}. Each DAG has three variables, which are all binary, and are generated in the following three ways for the three DAGs, respectively:}
		\begin{align*}
			\mathcal{G}_1: \quad & \prob(X_2=1)=p_0, \quad \prob(X_1=X_2|X_2)=p_1, \quad \prob(X_3=X_1|X_1)=p_2; \\
			\mathcal{G}_2: \quad & \prob(X_1=1)=p_0 p_1 + (1-p_0) (1-p_1), \quad \prob(X_3=X_1|X_1)=p_2, \\
			& \prob(X_2=X_1|X_1)=\left\{ \begin{array}{ll}
				\displaystyle \frac{p_0 p_1}{p_0 p_1 + (1-p_0) (1-p_1)}, & \hbox{if}~X_1=1,\\
				\displaystyle \frac{(1 - p_0) p_1}{(1-p_0) p_1 + p_0 (1-p_1)}, & \hbox{otherwise;}
			\end{array}
			\right. \\
			\mathcal{G}_3: \quad & \prob(X_3=1)=p_0 p_2 + (1-p_0) (1-p_2), \\
			&  \prob(X_1=X_3|X_3)=\left\{ \begin{array}{ll}
				\displaystyle \frac{p_0 p_2}{p_0 p_2 + (1-p_0) (1-p_2)}, & \hbox{if}~X_1=1,\\
				\displaystyle \frac{(1 - p_0) p_2}{(1-p_0) p_1 + p_0 (1-p_2)}, & \hbox{otherwise,}
			\end{array}
			\right. \\
			& \prob(X_2=X_1|X_1)=\left\{ \begin{array}{ll}
				\displaystyle \frac{p_0 p_1}{p_0 p_1 + (1-p_0) (1-p_1)}, & \hbox{if}~X_1=1,\\
				\displaystyle \frac{(1 - p_0) p_1}{(1-p_0) p_1 + p_0 (1-p_1)}, & \hbox{otherwise;}
			\end{array}
			\right. 
		\end{align*}
		\tsb{for some $p_0\in (0,1)$, $p_1,p_2\in (0,0.5)\cup (0.5, 1)$. It can be shown that $(X_1,X_2,X_3)$ has the same likelihood function, and the three DAGs are not identifiable.} 
		
		\tsb{Suppose we test whether there is an edge from $X_1$ to $X_2$, i.e., we test the hypotheses in \eqref{eqn:hypothesis} with $j=2, k=1$. We first apply the structural learning to estimate the DAG. When the estimated DAG equals $\mathcal{G}_1$, since $X_1$ is not in the ancestor set of $X_2$, following Step 2b of Algorithm \ref{alg:full}, our test returns the $p$-value of 1 directly, and thus would not reject the null hypothesis. When the estimated DAG equals $\mathcal{G}_2$, since the ancestor set of $X_2$ contains $X_1$, while $\mathcal{M} = \widehat{\hbox{AC}}_j - \{k\} = \emptyset$, the problem becomes testing the marginal independence between $X_1$ and $X_2$. Following Steps 3 to 6 of Algorithm \ref{alg:full}, our test would reject the null, as there is a link from $X_1$ to $X_2$. When the estimated DAG equals $\mathcal{G}_3$, since the ancestor set of $X_2$ contains both $X_1$ and $X_3$, and $\mathcal{M} = \widehat{\hbox{AC}}_j - \{k\} = \{3\}$, the problem becomes testing the conditional independence between $X_1$ and $X_2$ given $X_3$. Again, following Steps 3 to 6 of Algorithm \ref{alg:full}, our test would reject the null. In this example, we are not able to differentiate $\mathcal{G}_2$ and $\mathcal{G}_3$ in the equivalence class from our testing result alone. Even though the testing result is different when the estimated DAG equals $\mathcal{G}_1$, we still do not know if the estimated DAG corresponds to the true DAG in the equivalence class where the data is generated from.}
	\end{example}
	
	Therefore, without specific distributional assumptions, it is generally impossible to distinguish DAGs in an equivalence class, and our test alone cannot either. The main reason is that there is no way to tell if the estimated DAG actually corresponds to the true DAG. Although our test result depends on the estimated DAG, or say, the estimated ordering of the nodes, it is independent of the true DAG that generates the data.

\subsection{Extension to non-stationary and time-varying DAG}
\label{sec:ext-nonstationary}

In Section \ref{sec:time-dependency}, we have focused on the case when DAG is stationary, as imposed by condition (B2). We have also excluded the case when DAG is time-varying, as implied by condition (B3). In this section, we again outline the key steps of extensions, first to non-stationary DAG, then to time-varying DAG. We leave the full investigation as possible future research. To simplify the presentation, we assume $T_1=T_2=\cdots=T_N=T$. In addition, we denote the random variable $X_j$ at time $t$ as $X_{j,t}$, for $j=1,\ldots,d, t=1,\ldots,T$. 

We first consider a non-stationary DAG, and relax the stationarity condition (B2). Toward that end, suppose the DAG structure is piecewise constant over time. That is, there exist some change points, $1 = \tau_1 < \tau_2 < \ldots < \tau_M = T$, such that the random vectors $\mathbb{X}_{i,\tau_m}, \mathbb{X}_{i,\tau_m+1}, \ldots, \mathbb{X}_{i,\tau_{m+1}-1}$ are stationary for any $m = 1, \ldots, M-1$. Then, our goal is to test if there exists a directed edge from $X_{k,t}$ to $X_{j,t}$, for some $\tau_m \le t < \tau_{m+1}$. 

To test the hypotheses, we first estimate the change point locations and the graph structures given each half of the data. We consider the following optimization, 
\begin{align}\label{eqn:dyn-DAG}
	\begin{split}
		\min_{\thetabf} \sum_{j=1}^{d} \left[\sum_{i\in \mathcal{I}_{s}} \sum_{m=1}^M \sum_{t=\tau_{m}}^{\tau_{m+1}-1}  \big\{ \Xbm_{i,t,j}-\textrm{MLP}(\Xbm_{i,t};\thetabf_{j,m}) \big\}^2 +\sum_m \frac{\lambda  n_s (\tau_{m}-\tau_{m-1})}{T}\big\| \Abf_{j,m}^{(1)} \big\|_{1,1}\right] \\ 
		+ \; \gamma n_s M, \quad \textrm{subject to } \; \textrm{trace}[\exp\{ W_{m}(\thetabf) \circ W_{m}(\thetabf) \}] = d,
	\end{split}	
\end{align}
for all $m = 1, \ldots, M$, where $\theta_{j,m} = \{A^{(l)}_{j,m},b^{(l)}_{j,m}\}_{l}$ denotes the parameters in MLP that models the conditional mean function of $\mathbb{X}_{i,t,j}$ when $t$ belongs to the time interval $[\tau_{m},\tau_{m+1})$, and $W_{m}(\theta)$ is a $d\times d$ matrix whose $(k,j)$th entry equals the Euclidean norm of the $k$th column of $\Abf_{j,m}^{(1)}$. The first penalty in \eqref{eqn:dyn-DAG} is placed on $\| \Abf_{j,m}^{(1)} \big\|_{1,1}$ and is to impose the sparsity structure on the estimated DAG. The second penalty in \eqref{eqn:dyn-DAG} is placed on $M$, and is to penalize the total number of change points. Dynamic programming method such as \citet{friedrich2008complexity} can be employed to solve the optimization problem \eqref{eqn:dyn-DAG}.  Let $\widehat{\tau}_m$ denote the estimated change point locations, and $\widehat{\mathcal{G}}^{(s)}_{\widehat{\tau}_m}$ denote the estimated graphs, $m = 1,\ldots, \widehat{M}$, where $\widehat{M}$ denotes the corresponding estimator for $M$. Let $\widehat{\textrm{AC}}_j^{(s)}$ denote the set of ancestors of $j$ based on $\widehat{\mathcal{G}}^{(s)}_{\widehat{\tau}_m}$, and $\mathcal{M}^{(s)}=\widehat{\textrm{AC}}_j^{(s)}-\{k\}$. We apply Steps 3 to 6 of Algorithm \ref{alg:full} to $\{ \mathbb{X}_{i,t} \}_{ 1\le i\le N, \widehat{\tau}_m\le t<\widehat{\tau}_{m+1} }$, and derive the corresponding $p$-value. 

We can again show that the above test is consistent. This is based on the following key observation. Under the piecewise stationary structure, the number of change points can be consistently estimated, and the estimated change point locations converge at a faster rate than the estimated DAG. This phenomenon is well-known in the time series literature \citep[see e.g.,][]{boysen2009consistencies}, where the estimated change point converges at a rate of $O_p(n^{-1}\log n)$, and this rate is much faster than the parametric rate. As a consequence, our test is to behave as well as if the true change point locations were known in advance.  

Next, we briefly consider a time-varying DAG, which allows to test directed links from past to future observations. Suppose at time $t$, a given node not only depends on other nodes at the same time, but also on past variables at time $t-1,t-2,\cdots,t-Q$ as well. Our goal is to test if there exists a directed edge from $X_{k, t-q}$ to $X_{j, t}$, for some $0 \le q \le Q$. We can essentially apply Algorithm \ref{alg:full} to this problem, and can establish the consistency of the test similarly.

\section{Additional Results}
\label{sec:add-results}

\subsection{Oracle property of  the DAG learner}
\label{sec:thmcsl}

As a by-product of our theoretical analysis, we derive the oracle property of the DAG estimator produced by \eqref{eqn:DAG}. This result is to guarantee $\prob\left( \cap_{j\in \{1,\cdots,d\}}\{\textrm{PA}_j\subseteq \widehat{\textrm{AC}}^{(s)}_j\} \right) \to 1$, which was not available in \cite{zheng2020learning}. It implies that the ordering of the true DAG can be consistently estimated, which in turn ensures the validity of (C1). In this section, for simplicity, we assume the DAG dimension $d$ is fixed. Nevertheless, we can extend our proof to the high-dimensional setting in a relatively straightforward fashion, by imposing a certain H{\"o}lder smoothness assumption on $\{f_j\}_j$; see, e.g.,  \citet[Assumption 2]{farrell2018deep}.

We first define the oracle estimator. For an ordering $\pi=(\pi_1,\ldots,\pi_d)$ for a given DAG, consider the estimator $\widetilde{\thetabf}^{(s)}(\pi) = \big\{ \widetilde{\thetabf}^{(s)}_1(\pi), \ldots,$ $\widetilde{\thetabf}^{(s)}_d(\pi) \big\}$, where each $\widetilde{\thetabf}^{(s)}_j(\pi)$ is obtained by
\begin{eqnarray*}
	\argmin_{\substack{\thetabf_j = \left\{ \Abf_j^{(1)}, \bbf^{(1)}, \ldots, \Abf_j^{(L)}, \bbf^{(L)} \right\} 
			\scriptsize{\textrm{supp}}\left( \Abf_j^{(1)} \right) \in \{\pi_1,\ldots,\pi_{j-1}\} } }
	\sum_{i \in \mathcal{I}_{s}} \sum_{t=1}^{T} \left\{ \Xbm_{i,t,j}-\textrm{MLP}(\Xbm_{i,t};\thetabf_j) \right\}^2 +  \frac{\lambda NT}{2} \big\|\Abf_j^{(1)} \big\|_{1,1},
\end{eqnarray*}
where $\textrm{supp}\left( \Abf_j^{(1)} \right) \in \{\pi_1,\ldots,\pi_{j-1}\}$ means that, for any $l$ that does not belong to this set, the $l$th column of $\Abf_j^{(1)}$ equals zero. In other words, the estimator $\widetilde{\thetabf}^{(s)}_j(\pi)$ is computed as if the order $\pi$ were known in advance.

Next, let $\Pi^*$ denote the set of all true orderings. This means, for any true ordering $\pi^*\in \Pi^*$, $\textrm{PA}_j\subseteq \{\pi_1^*,\ldots,\pi_{j-1}^*\}$, for any $j=1, \ldots, d$. In other words, the parents of each node should appear before the occurrence of this node under $\pi^*$. It is also worth mentioning that, the true ordering is \emph{not} necessarily unique, even though the underlying DAG is unique. For instance, consider Example \ref{examv} with a v-structure as shown in Figure \ref{fig:illustration}(a). In this example, both $(1,2,3)$ and $(1,3,2)$ are the true orderings, as there are no directional edges between nodes $X_2$ and $X_3$.

Next, we introduce some additional conditions. For any ordering $\pi$, define a least squares loss function, $\mathcal{L}(\pi)=\sum_{j=0}^{d-1} \Mean \left\{ X_{j+1}-\Mean \left( X_{j+1}|\Xbf_{\{\pi_1,\ldots,\pi_{j} \}} \right) \right\}^2$. Moreover, we focus on neural networks with a ReLU activation function, $\sigma(x)=\max(0,x)$.

\vspace{-0.05in}
\begin{enumerate}[({C}1)]
	\addtocounter{enumi}{4}
	\item All minimizers of $\mathcal{L}(\pi)$ are contained in $\Pi^*$.
	\item The widths of all layers in the MLP share a common asymptotic order $H$. Besides, the number of layers $L$ and the asymptotic order $H$ diverge with $NT$, in that $HL=O\{(NT)^{\kappa_8}\}$, for some constant $\kappa_8<1/2$.
	\item Suppose \tsb{MLP}$\big\{ \cdot;\widetilde{\thetabf}^{(s)}(\pi) \big\}$ is bounded for any $\pi$.
	\vspace{-0.05in}
\end{enumerate}

\noindent
Condition (C5) is reasonable and holds in numerous scenarios. One example is when all the random errors $\{\varepsilon_j\}_{j=1}^{d}$ in model \eqref{eqn:model} are normally distributed with equal variance. In that case, the least squares loss $\mathcal{L}$ is proportional to the expected value of the log-likelihood of $\Xbf$. Since the underlying DAG is identifiable, any ordering that minimizes the expected log-likelihood belongs to $\Pi^*$. Condition (C6) is also mild, as both $H$ and $L$ are the parameters that we specify. The part that $HL=O\{(NT)^{\kappa_8}\}$ ensures that the stochastic error resulting from the parameter estimation in the MLP is negligible. Condition (C7) ensures that the optimizer would not diverge in the $\ell_{\infty}$ sense. Similar assumptions are common in the literature to derive the convergence rates of deep learning estimators \citep[see e.g.][]{farrell2018deep}.

Now we show that the estimator $\widehat{\thetabf}^{(s)}$ obtained from \eqref{eqn:DAG} satisfies the oracle property, i.e., $\widehat{\thetabf}^{(s)}=\widetilde{\thetabf}^{(s)}(\pi^*)$, for some $\pi^* \in \Pi^*$. In other words, $\widehat{\thetabf}^{(s)}$ is computed as if one of the true ordering were known in advance. By the definition of $\Pi^*$, Condition (C1) holds for our estimated DAG. Moreover, we note that the oracle property does \emph{not} imply the selection consistency, i.e., $\textrm{PA}_j = \widehat{\textrm{PA}}_j$, nor the sure screening property, in that $\textrm{PA}_j \subseteq \widehat{\textrm{PA}}_j$, for any $j=1,\ldots,d$.

\begin{theorem}\label{thm4}
	Suppose $\{f_j\}_j$ in model \eqref{eqn:model} are a set of continuous functions, (C5)-(C7) hold, the $\beta$-mixing coefficient $\beta(q)$ in (C4) decays exponentially with $q$, and $\lambda \to 0$. Then, with probability approaching one, $\widehat{\thetabf}^{(s)}=\widetilde{\thetabf}^{(s)}(\pi^*)$, for some $\pi^* \in \Pi^*$, as either $N$ or $T \to \infty$.
\end{theorem}

\subsection{Sample splitting}
\label{sec:more-splitting}

We employ the data splitting and cross-fitting strategy for our test, and use a binary-split in Section \ref{sec:test}. To mitigate sample randomization arising from a single binary-split, in this section, we develop a version of our test based on multiple binary-splits. The main idea is to apply the binary-split in Algorithm \ref{alg:full} multiple times, then combine the $p$-values from all splits. In addition, we may also adopt the multi-split  strategy of \citet{romano2019}. These modifications may help reduce the sampling randomization, and may potentially improve the power of the test, but also come with a price of increased computations.
Specifically, we carry out the binary-split $R$ times. For the $r$th binary-split, we randomly split all samples $\{1,\ldots,N\}$ into two disjoint subsets $\mathcal{I}_{r,1}\cup \mathcal{I}_{r,2}$ of equal sizes. We then apply Algorithm \ref{alg:full}  to compute the $p$-values, $\widehat{p}^{(r,1)}$ and $\widehat{p}^{(r,2)}$, respectively, for each half of the data. We next combine these $p$-values by,
\begin{eqnarray*}
	\widehat{p} =  \min\Big(1,q_{\gamma}\left[ \left\{\gamma^{-1}\widehat{p}^{(r,s)}(0,q), r=1,\ldots,R, s=1,2 \right\} \right] \Big),
\end{eqnarray*}
where $0 < \gamma < 1$ is a constant, and $q_{\gamma}$ is the empirical $\gamma$-quantile. We recommend to set $\gamma$ to a small value, such as $0.1$ or $0.2$. This follows a similar idea as \cite{meinshausen2009}.

\subsection{Gaussian versus non-Gaussian input noise for GANs}	
\label{sec:more-noise}

When learning the distribution generator in Section \ref{sec:gans}, we take the Gaussian noise as the input of GANs. One may also use other non-Gaussian noises, e.g., uniformly distributed random vectors over a unit hypercube. In general, the performance of the generator computed via GANs is not overly sensitive to the choice of the distribution of the input noise. This is partly because, the objective of the GAN step is to learn a generator $\mathbb{G}$, such that the conditional distribution of $X_k$ given $X_{\mathcal{M}^{(s)}}$ can be well approximated by that of $\mathbb{G}(X_{\mathcal{M}^{(s)}},Z_{j,k})$ given  $X_{\mathcal{M}^{(s)}}$, where $Z_{j,k}$ is the Gaussian noise. Suppose we use some non-Gaussian noise $V_{j,k}$ with the same dimension. Under some regularity conditions, there exists a transformation function $\phi$, such that $\phi(V_{j,k})$ has the same distribution as $Z_{j,k}$. Define $\mathbb{G}_{\phi}(X_{\mathcal{M}^{(s)}}, V_{j,k})=\mathbb{G}(X_{\mathcal{M}^{(s)}}, \phi(V_{j,k}))$. Then, $\mathbb{G}_{\phi}$ has the same smoothness properties as $\mathbb{G}$. As such, the estimated distribution generator for $\mathbb{G}_{\phi}$ is expected to have similar statistical properties as that for $\mathbb{G}$ \citep{chen2020statistical}. 	
\begin{table}[b!]
	\centering
	\caption{The empirical size and power of the proposed testing method SUGAR under two distributions, Gaussian and uniform, for the input noise in GANs.}
	\label{tab:noise-dist}
	\setlength{\tabcolsep}{1.8mm}{
		\begin{tabular}{|c|c|c|c|c|c|c|c|c|} \hline
			\multicolumn{3}{|c|}{\parbox{1.5cm}{\centering Edge}} &
			\multicolumn{2}{c|}{\parbox{2.75cm}{\centering $j=35, k=5$}} &
			\multicolumn{2}{c|}{\parbox{2.75cm}{\centering $j=35, k=31$}}  &
			\multicolumn{2}{c|}{\parbox{2.75cm}{\centering $j=40, k=16$}} \\
			\hline
			\multicolumn{3}{|c|}{Hypothesis}&
			\multicolumn{2}{c|}{$\mathcal{H}_0$} &
			\multicolumn{2}{c|}{$\mathcal{H}_0$} &
			\multicolumn{2}{c|}{$\mathcal{H}_0$} \\
			\hline
			\multicolumn{3}{|c|}{Input Noise} & Normal & Uniform & Normal & Uniform & Normal & Uniform \\
			\hline
			\multicolumn{3}{|c|}{$\alpha=0.05$} & 0.050 & 0.046 & 0.012 & 0.022 & 0.016 & 0.016\\
			\hline
			\multicolumn{3}{|c|}{$\alpha=0.10$} & 0.078 & 0.078 & 0.032 & 0.046 & 0.032 & 0.022\\
			\hline
			\multicolumn{3}{|c|}{\parbox{1.5cm}{\centering Edge}} &
			\multicolumn{2}{c|}{\parbox{2.75cm}{\centering $j=45, k=14$}} &
			\multicolumn{2}{c|}{\parbox{2.75cm}{\centering $j=45, k=15$}}  &
			\multicolumn{2}{c|}{\parbox{2.75cm}{\centering $j=50, k=14$}} \\
			\hline
			\multicolumn{3}{|c|}{Hypothesis}&
			\multicolumn{2}{c|}{$\mathcal{H}_0$} &
			\multicolumn{2}{c|}{$\mathcal{H}_0$} &
			\multicolumn{2}{c|}{$\mathcal{H}_0$} \\
			\hline
			\multicolumn{3}{|c|}{Input Noise} & Normal & Uniform & Normal & Uniform & Normal & Uniform \\
			\hline
			\multicolumn{3}{|c|}{$\alpha=0.05$} & 0.014 & 0.020 & 0.032  & 0.030 & 0.030 & 0.034\\
			\hline
			\multicolumn{3}{|c|}{$\alpha=0.10$} & 0.030 & 0.032 & 0.058 & 0.052 & 0.046 & 0.052\\
			\hline
			\multicolumn{3}{|c|}{\parbox{1.5cm}{\centering Edge}} &
			\multicolumn{2}{c|}{\parbox{2.75cm}{\centering $j=35, k=4$}} &
			\multicolumn{2}{c|}{\parbox{2.75cm}{\centering $j=35, k=30$}}  &
			\multicolumn{2}{c|}{\parbox{2.75cm}{\centering $j=40, k=15$}} \\
			\hline
			\multicolumn{3}{|c|}{Hypothesis}&
			\multicolumn{2}{c|}{$\mathcal{H}_1$} &
			\multicolumn{2}{c|}{$\mathcal{H}_1$} &
			\multicolumn{2}{c|}{$\mathcal{H}_1$} \\
			\hline
			\multicolumn{3}{|c|}{Input Noise} & Normal & Uniform & Normal & Uniform & Normal & Uniform \\
			\hline
			\multicolumn{3}{|c|}{$\alpha=0.05$} & 0.534 & 0.524 & 0.992 & 0.992 & 0.550 & 0.550\\
			\hline
			\multicolumn{3}{|c|}{$\alpha=0.10$} & 0.546 & 0.552 & 0.992 & 0.992 & 0.550 & 0.550\\
			\hline
			\multicolumn{3}{|c|}{\parbox{1.5cm}{\centering Edge}} &
			\multicolumn{2}{c|}{\parbox{2.75cm}{\centering $j=45, k=12$}} &
			\multicolumn{2}{c|}{\parbox{2.75cm}{\centering $j=45, k=13$}}  &
			\multicolumn{2}{c|}{\parbox{2.75cm}{\centering $j=50, k=13$}} \\
			\hline
			\multicolumn{3}{|c|}{Hypothesis}&
			\multicolumn{2}{c|}{$\mathcal{H}_1$} &
			\multicolumn{2}{c|}{$\mathcal{H}_1$} &
			\multicolumn{2}{c|}{$\mathcal{H}_1$} \\
			\hline
			\multicolumn{3}{|c|}{Input Noise} & Normal & Uniform & Normal & Uniform & Normal & Uniform \\
			\hline
			\multicolumn{3}{|c|}{$\alpha=0.05$} & 0.946 & 0.952 & 0.808 & 0.824 & 0.670 & 0.670\\
			\hline
			\multicolumn{3}{|c|}{$\alpha=0.10$} & 0.948 & 0.954 & 0.816 & 0.832 & 0.672 & 0.670\\
			\hline
		\end{tabular}
	}
\end{table}

We also conduct a simulation to examine the empirical performance of our test under two distributions, Gaussian and uniform, for the input noise. We adopt the nonlinear model \eqref{eqn:sim-nonlinear} in Section \ref{sec:sim}, with $N = 20, T=100, \delta=1, d=50, \zeta=0.1$. Table \ref{tab:noise-dist} reports the empirical size and power, i.e., the percentage of times out of 500 data replications when the $p$-value is smaller than the nominal level $\alpha=0.05$ and $\alpha=0.10$, respectively, for some pairs of nodes. It is clearly seen from the table that the results are very similar for two input noise distributions.

\subsection{Condition (C1)}
\label{sec:more-table}

To establish the consistency of the proposed test, we require the initial DAG estimator can estimate the ordering consistently; see condition (C1) in Section \ref{sec:theory}. However, even when (C1) does not hold, our proposed test may still control the type-I error. Actually, in our simulation examples in Section \ref{sec:sim}, (C1) does not alway hold. 
Table \ref{tab:c1} reports the percentage of times out of 500 data replications when (C1) holds for those selected nodes reported in Table \ref{tab:noise-dist} for the nonlinear model \eqref{eqn:sim-nonlinear}. It is seen that, for numerous nodes, (C1) only holds for a small fraction of times. 

\begin{table}[h!]
	\centering
	\caption{The percentage of times out of 500 data replications when (C1) holds for selected nodes for four simulation models.}
	\label{tab:c1}
	\setlength{\tabcolsep}{1.5mm}{
		\begin{tabular}{|cl|c|c|c|c|} \hline
			\multicolumn{6}{|l|}{Nonlinear model \eqref{eqn:sim-nonlinear} with $d=50, \zeta=0.10$} \\ \hline
			\hspace{0.25in} & Node $j$ &  35 & 40 & 45 &  50 \\ \hline
			\hspace{0.25in}  & Percentage & 11.6\% & 44.0\% & 16.4 \% & 2.2\%  \\ \hline
			\multicolumn{6}{|l|}{Nonlinear model \eqref{eqn:sim-nonlinear} with $d=100, \zeta=0.04$} \\ \hline
			\hspace{0.25in} & Node $j$ & 80 & 85 & 90 & \\ \hline
			\hspace{0.25in} & Percentage  & 48 \% & 1.9 \% & 0 \% & \\ \hline
			\multicolumn{6}{|l|}{Nonlinear model \eqref{eqn:sim-nonlinear} with $d=150, \zeta=0.02$} \\ \hline
			\hspace{0.25in} & Node $j$ & 132 & 135 & 137 & 140  \\ \hline
			\hspace{0.25in} & Percentage & 37.1\% & 20.0\% & 46.5 \% & 91.8\% \\ \hline
		\end{tabular}
	}
\end{table}

\subsection{Power comparison}
\label{sec:more-power}

\begin{figure}[!t]
	\centering
	\includegraphics[width=15.25cm,height=15.75cm]{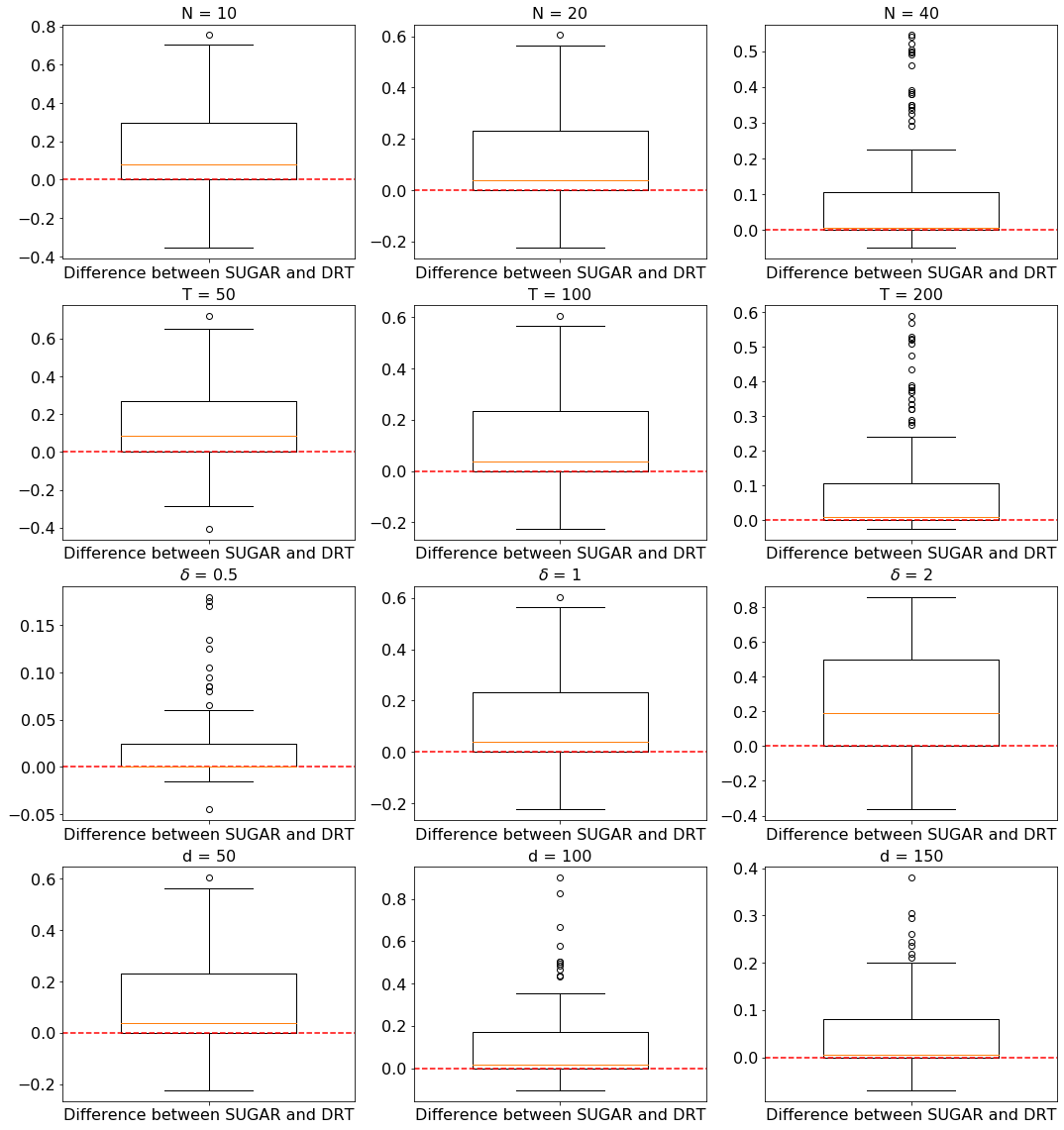}
	\caption{The boxplots of the difference of the empirical power of our proposed test (SUGAR) and that of the double regression-based test (DRT), under four sets of varying parameters: first row $N = \{ 10, 20, 40 \}$, second row $T = \{ 50, 100, 200 \}$, third row $\delta = \{ 0.5, 1, 2 \}$, and fourth row $(d, \zeta) = \{(50, 0.10), (100, 0.04), (150, 0.02) \}$.}
	\label{fig:sim_diff}
\end{figure}

To compare the power of the two testing methods, we further report the empirical power of our SUGAR method minus that of DRT in Figure \ref{fig:sim_diff}. It is seen that SUGAR achieves generally a higher power than DRT, over 75\% of the times in all scenarios. 

\section{Proofs}
\label{sec:proofs}

We present the technical proofs of Proposition \ref{prop1}, Theorems \ref{thm1}, \ref{thm2} and \ref{thm3}, followed by an auxiliary lemma needed for the proof of Theorem \ref{thm3}. To simplify the notation, we use $O_s$ to denote the data subset $\{\Xbm_{i,t}\}_{i\in \mathcal{I}_{s}, 1\le t\le T}$ throughout this section.

\subsection{Proof of Proposition \ref{prop1}}

We first show that $\mathcal{H}_0(j,k)$ implies $ \mathcal{H}_0^*(j,k |$ $\mathcal{M})$. Under model \eqref{eqn:model}, it follows from Theorem 1.4.1 of \cite{pearl2009causal} that the joint distribution of $(X_1,\ldots,X_d)$ is Markov with respect to the graph. This suggests that the $d$-separation implies the conditional independence \citep{pearl2009causal}. Under $\mathcal{H}_0(j,k)$, $X_j$ and $X_k$ are $d$-separated by $\Xbf_{\scriptsize{\hbox{PA}}_j}$. Under the given conditions on $\mathcal{M}$, we obtain that $X_j$ and $X_k$ are $d$-separated by $\Xbf_{\mathcal{M}-\{k\}}$ as well. Consequently, $\mathcal{H}_0^*(j,k|\mathcal{M})$ holds.

We next show that $\mathcal{H}_0^*(j,k|\mathcal{M})$ implies $\mathcal{H}_0(j,k)$. Under $ \mathcal{H}_0^*(j,k|\mathcal{M})$, we have $\Mean (X_j|\Xbf_{\mathcal{M}},X_k) = \Mean \left( X_j|\Xbf_{\mathcal{M}-\{k\}} \right)$. Since $j\in \tsb{\hbox{DS}}_k$ and $\mathcal{M}\cap \tsb{\hbox{DS}}_j=\emptyset$, the additive noise $\varepsilon_j$ is independent of $X_k$ and $\Xbf_{\mathcal{M}}$. Under model \eqref{eqn:model}, we obtain that $\Mean \{f_j(\Xbf_{\scriptsize{\hbox{PA}}_j})| \Xbf_{\mathcal{M}},X_k \} = \Mean \left[ f_j(\Xbf_{\scriptsize{\hbox{PA}}_j})|\Xbf_{\mathcal{M}-\{k\}} \right]$. Since ${\hbox{PA}}_j\subseteq \mathcal{M}$, we have $\Mean \{f_j(\Xbf_{\scriptsize{\hbox{PA}}_j})| \Xbf_{\mathcal{M}},X_k \}=f_j(\Xbf_{\scriptsize{\hbox{PA}}_j})$. Consequently, we have $f_j(\Xbf_{\scriptsize{\hbox{PA}}_j}) = \Mean \left[ f_j(\Xbf_{\scriptsize{\hbox{PA}}_j})|\Xbf_{\mathcal{M}-\{k\}} \right]$. As such, we have $k \notin \hbox{PA}_j$. Otherwise, there would exist two structural equation models with different graphs that lead to the same joint distribution of $(X_1,\ldots,X_d)$, and the identifiability condition would have been violated. Therefore, $\mathcal{H}_0(j,k)$ holds.

This completes the proof of Proposition \ref{prop1}.
\eop

\subsection{Proof of Theorem \ref{thm1}}	

It suffices to show that the null hypothesis in \eqref{eqn:hypothesis} is sufficient and necessary to $I(j,k|\mathcal{M};h)=0$ for all square integrable functions $h$.

The sufficiency follows immediately from Proposition \ref{prop1} and the definition of the conditional independence.

To prove the necessity, it suffices to show there exists some function $h$ such that $I(j,k|\mathcal{M},h)\neq 0$ under $\mathcal{H}_1(j,k)$. Since $X_j$ has a finite second moment, it follows from model \eqref{eqn:model} and Jensen's inequality that $\Mean \left\{ f_j^2(X_k, \Xbf_{\scriptsize{\hbox{PA}}_j}) \right\}$ is also finite. Define the function, $h^*(X_k, \Xbf_{\mathcal{M}-\{k\}}) = f_j(X_k, \Xbf_{\scriptsize{\hbox{PA}}_j}) - \Mean \big\{ f_j(X_k,$ $\Xbf_{\scriptsize{\hbox{PA}}_j}) | \Xbf_{\mathcal{M}-\{k\}} \big\}$. It follows that $h^*$ is square integrable. Also by definition,
\begin{eqnarray*}
	I(j,k|\mathcal{M},h^*) = \Mean \left[ f_j(X_k, \Xbf_{\scriptsize{\hbox{PA}}_j}) - \Mean \left\{f_j(X_k, \Xbf_{\scriptsize{\hbox{PA}}_j})|\Xbf_{\mathcal{M}-\{k\}} \right\} \right]^2.
\end{eqnarray*}
This measure is not zero. Otherwise, we would have $f_j(X_k, \Xbf_{\scriptsize{\hbox{PA}}_j}) = \Mean \big\{f_j(X_k,$ $ \Xbf_{\scriptsize{\hbox{PA}}_j}) | \Xbf_{\mathcal{M}-\{k\}} \big\}$, which would further imply that the data can be generated by another structural equation model such that $X_k$ is not a direct cause of $X_j$. This would have violated the identifiability condition.

This completes the proof of Theorem \ref{thm1}.
\eop

\subsection{Proof of Theorem \ref{thm2}}
\label{sec:proof-thm2}

We begin with a definition. Define
\begin{eqnarray*}
	\widehat{I}_{b,\textrm{CF}}^{(s)*} & = & \frac{2}{NT} \sum_{i\in \mathcal{I}_{\ell}^c} \sum_{1\le t\le T}I_{i,t,b}^{(s)*}, \;\; \textrm{ where } \\
	I_{i,t,b}^{(s)*} & = & \left\{ \Xbm_{i,t,j}-g^{(s)}\left( \Xbm_{i,t,\mathcal{M}^{(s)}} \right) \right\} \left[h_b^{(s)}\left( \Xbm_{i,t,k},\Xbm_{i,t,\mathcal{M}^{(s)}} \right) \right. \\
	& & \quad\quad\quad\quad \left. - \Mean \left\{ h_b^{(s)}\left( \Xbm_{i,t,k},\Xbm_{i,t,\mathcal{M}^{(s)}} \right) | \Xbm_{i,t,\mathcal{M}^{(s)}} \right\}\right].
\end{eqnarray*}
Note that $\left| \widehat{I}_{b,\textrm{CF}}^{(s)}-\widehat{I}_{b,\textrm{CF}}^{(s)*} \right| \le \sum_{l=1}^3 \left| \eta_{b,l}^{(s)} \right|$, where 
\vspace{-0.01in}
\begin{eqnarray*}
	\eta_{b,1}^{(s)} & = & \frac{2}{NT}\sum_{i\in \mathcal{I}_{\ell}^c}\sum_{1\le t\le T} \left\{ \Xbm_{i,t,j}-g^{(s)}\left( \Xbm_{i,t,\mathcal{M}^{(s)}} \right) \right\} \\
	& & \times \left[\frac{1}{M}\sum_{m=1}^M h_b^{(s)}\left( \widetilde{\Xbm}_{i,t,k}^{(s,m)},\Xbm_{i,t,\mathcal{M}^{(s)}} \right) - \Mean \left\{h_b^{(s)}\left( \Xbm_{i,t,k},\Xbm_{i,t,\mathcal{M}^{(s)}} \right) | \; \Xbm_{i,t,\mathcal{M}^{(s)}} \right\}\right], \\
	\eta_{b,2}^{(s)} & = & \frac{2}{NT}\sum_{i\in \mathcal{I}_{\ell}^c}\sum_{1\le t\le T} \left\{ g^{(s)}\left( \Xbm_{i,t,\mathcal{M}^{(s)}} \right) - \widehat{g}^{(s)}\left( \Xbm_{i,t,\mathcal{M}^{(s)}} \right) \right\} \\
	& & \times \left[ h_b^{(s)}\left( \Xbm_{i,t,k},\Xbm_{i,t,\mathcal{M}^{(s)}} \right) - \Mean \left\{ h_b^{(s)}\left( \Xbm_{i,t,k},\Xbm_{i,t,\mathcal{M}^{(s)}} \right) | \; \Xbm_{i,t,\mathcal{M}^{(s)}} \right\}\right], 
\end{eqnarray*}
\begin{eqnarray*}	
	\eta_{b,3}^{(s)} & = & \frac{2}{NT}\sum_{i\in \mathcal{I}_{\ell}^c}\sum_{1\le t\le T} \left\{ g^{(s)}\left( \Xbm_{i,t,\mathcal{M}^{(s)}} \right) - \widehat{g}^{(s)}\left( \Xbm_{i,t,\mathcal{M}^{(s)}} \right) \right\} \\
	& & \times \left[\frac{1}{M}\sum_{m=1}^M h_b^{(s)}\left( \widetilde{\Xbm}_{i,t,k}^{(s,m)},\Xbm_{i,t,\mathcal{M}^{(s)}} \right) - \Mean \left\{h_b^{(s)}\left( \Xbm_{i,t,k},\Xbm_{i,t,\mathcal{M}^{(s)}} \right) | \; \Xbm_{i,t,\mathcal{M}^{(s)}} \right\}\right].
\end{eqnarray*}
Condition (C1) implies that the set $\widehat{\textrm{AC}}_j^{(s)}$ meets the conditions of Proposition \ref{prop1}.  

We next divide the proof of this theorem into 6 steps. In Steps 1 to 3, we show that $\eta_{\widehat{b}^{(s)},l}^{(s)} = o_p\{(NT)^{-1/2}\}$, for $l=1,2,3$, respectively. In Step 4, we show that, conditional on $O_s$,
\begin{eqnarray}\label{eqn:step4}
	\frac{ \widehat{I}_{\widehat{b}^{(s)},\textrm{CF}}^{(s)*}}{\sqrt{\Var\left( \widehat{I}_{\widehat{b}^{(s)},\textrm{CF}}^{(s)*} \; | \; O_s \right)}}\stackrel{d}{\to} N(0,1).
\end{eqnarray}
In Step 5, we show that the batched mean estimator $\widehat{\sigma}_{\widehat{b}^{(s)},\textrm{CF}}^{(s)}$ converges to the standard deviation of $\sqrt{(NT)/2} \widehat{I}_{\widehat{b}^{(s)},\textrm{CF}}^{(s)}$ given $O_s$ and the indices of the data subsets $\mathcal{I}_s,\mathcal{I}_s^c$. This together with Step 4 yields that $\sqrt{(NT)/2} \widehat{I}_{\widehat{b}^{(s)},\textrm{CF}}^{(s)}/\widehat{\sigma}_{\widehat{b}^{(s)},\textrm{CF}}^{(s)}\stackrel{d}{\to} N(0,1)$ given $O_s$, $\mathcal{I}_s$ and $\mathcal{I}_s^c$. Hence, $\sqrt{(NT)/2} \widehat{I}_{\widehat{b}^{(s)},\textrm{CF}}^{(s)}/\widehat{\sigma}_{\widehat{b}^{(s)},\textrm{CF}}^{(s)}$ converges to a standard normal distribution unconditionally as well. In Step 6, we put all the above results together to complete the proof. In the following, we assume the data $O_s$ is fixed. The expectation and variance are taken with respect to the data $\{\Xbm_{i,t}\}_{i\in \mathcal{I}_s^c,1\le t\le T}$ conditional on $O_s$.

\smallskip
\noindent
\textbf{Step 1}.  We first use Berbee's coupling lemma \citep[Lemma 4.1]{Dedecker2002} to approximate $\eta^{(s)}_{\widehat{b}^{(s)},1}$ by a sum of independent random variables. We then derive the convergence rate of $\eta_{\widehat{b}^{(s)},1}^{(s)}$. Since we assume the data $O_s$ is fixed, the index $\widehat{b}^{(s)}$ is fixed as well.

Denote $\mathcal{I}_{\ell}^c=\{\ell_1,\ell_2,\ldots,\ell_{N/2}\}$ and $Q=NT/2$. Consider the sequence $\{\Xbm_{(n)}\}_{1\le n\le Q}$ formed by $\{\Xbm_{\ell_i,t}\}_{1\le i\le N/2, 1\le t\le T}$, such that $\Xbm_{\ell_i,t}=\Xbm_{((\ell_i-1)T+t)}$ for any $i,t$.  
By Condition (C3), each sequence $\{\Xbm_{i,t}\}_{t}$ is exponentially $\beta$-mixing, and so is $\{\Xbm_{(n)}\}_n$. Following the discussion after Lemma 4.1 of \cite{Dedecker2002}, we can construct a sequence of random vectors $\{\Xbm_{(n)}^0\}_n$, such that, with probability at least $1-Q\beta(q)/q$,
\begin{align*}
	\eta_{b,1}^{(s)} = \frac{1}{Q}\sum_{n=1}^{Q} \left\{ \Xbm_{(n),j}^0-g^{(s)}\left( \Xbm_{(n),\mathcal{M}^{(s)}}^0 \right) \right\} \times \left[\frac{1}{M}\sum_{m=1}^M h_b^{(s)}\left( \widetilde{\Xbm}_{(n),k}^{(m)},\Xbm_{(n),\mathcal{M}^{(s)}}^0 \right) \right. \\
	\left. - \Mean \left\{h_b^{(s)}\left( \Xbm_{(n),k}^0,\Xbm_{(n),\mathcal{M}^{(s)}}^0 \right) | \; \Xbm_{(n),\mathcal{M}^{(s)}}^0 \right\} \right],
\end{align*}
for any $b$, where we use $\widetilde{\Xbm}_{((\ell_i-1)T+t),k}^{(m)}$ to denote $\widetilde{\Xbm}_{\ell_i,t,k}^{(m)}$, and that the sequences $\{\Ubf_{2n}^0:n\ge 0\}$ and $\{\Ubf_{2n+1}^0:n\ge 0\}$ are i.i.d., with $\Ubf_{2n+1}^0=(\Xbm_{(nq)}^0,\Xbm_{(nq+1)}^0,\ldots,\Xbm_{(nq+q-1)}^0)$.

Let $\mathcal{I}_r=\{q\floor{Q/q}+1, q\floor{Q/q}+2,\ldots,Q\}$, we have
\begin{eqnarray*} \label{eqn:etab}
	\left| \eta_{\widehat{b}^{(s)},1}^{(s)} \right| \le \left|\frac{1}{Q}\sum_{\tau=1}^{\floor{Q/q}}\eta_{\widehat{b}^{(s)},1,\tau}^{(s)}\right|+\left|\frac{1}{Q}\sum_{\tau \in \mathcal{I}_r} \eta_{\widehat{b}^{(s)},1,\tau}^{(s)}\right| \equiv \delta_1 + \delta_2,
\end{eqnarray*}
with probability $1-Q\beta(q)/q$, where
\begin{align*}
	\eta_{b,1,\tau}^{(s)} = \sum_{n=(\tau-1)q+1}^{\tau q} \left\{ \Xbm_{(n),j}^0-g^{(s)}\left( \Xbm_{(n),\mathcal{M}^{(s)}}^0 \right) \right\} \times \left[\frac{1}{M}\sum_{m=1}^M h_b^{(s)}\left( \widetilde{\Xbm}_{(n),k}^{(m)},\Xbm_{(n),\mathcal{M}^{(s)}}^0 \right) \right. \\ 
	\left. - \Mean \left\{h_b^{(s)}\left( \Xbm_{(n),k}^0,\Xbm_{(n),\mathcal{M}^{(s)}}^0 \right) | \; \Xbm_{(n),\mathcal{M}^{(s)}}^0 \right\}\right],
\end{align*}
for $b=1,\ldots,B$. We next bound $\delta_1$ and $\delta_2$, respectively.

For $\delta_2$, since $\mathbb{H}^{(s)}$ is bounded, we have that,
\begin{eqnarray*}
	\delta_2 \le \frac{1}{Q} \sum_{n=(\tau-1)q+1}^{\tau q} \left| \Xbm_{(n),j}^0 - g^{(s)}\left( \Xbm_{(n),\mathcal{M}^{(s)}}^0 \right) \right|.
\end{eqnarray*}
The expectation of the above random variable is of the order $O(q N^{-1} T^{-1})$. Consequently, $\delta_2 = O_p(q N^{-1} T^{-1})$.

For $\delta_1$, without loss of generality, suppose $\floor{Q/q}$ is divisible by two. By construction, 
\begin{eqnarray*}
	\delta_1 \le \left|\frac{1}{Q}\sum_{\tau=1}^{\floor{Q/q}/2}\eta_{\widehat{b}^{(s)},1,2\tau-1}^{(s)}\right|+\left|\frac{1}{Q}\sum_{\tau=1}^{\floor{Q/q}/2}\eta_{\widehat{b}^{(s)},1,2\tau}^{(s)}\right|,
\end{eqnarray*}
where each of the above two terms corresponds to a sum of independent random variables. Since the data observations are stationary, it follows from Chebyshev's inequality that these two terms can be upper bounded by $O\left\{ (NTq)^{-1/2} \Var^{1/2}\left( \eta_{\widehat{b}^{(s)},1,\tau}^{(s)} \right) \right\}$. Next, it suffices to bound the variance term $\Var\left( \eta_{\widehat{b}^{(s)},1,\tau}^{(s)} \right)$.

By Cauchy-Schwarz inequality, we have
\begin{align*}
	\Var\left( \eta_{\widehat{b}^{(s)},1,\tau}^{(s)} \right) & \le q^2 \Mean \left\{ \Xbm_{(n),j}^0 - g^{(s)}\left( \Xbm_{(n),\mathcal{M}^{(s)}}^0 \right)\right\}^2 \\
	\times \left[\frac{1}{M}\sum_{m=1}^M \right. & \left. h_{\widehat{b}^{(s)}}^{(s)}\left( \widetilde{\Xbm}_{(n),k}^{(m)},\Xbm_{(n),\mathcal{M}^{(s)}}^0 \right) - \Mean \left\{ h_{\widehat{b}^{(s)}}^{(s)}\left( \Xbm_{(n),k}^0,\Xbm_{(n),\mathcal{M}^{(s)}}^0 \right) | \; \Xbm_{(n),\mathcal{M}^{(s)}}^0 \right\}\right]^2.
\end{align*}
Under $\mathcal{H}_0(j,k)$ and model \eqref{eqn:model}, the residual $\Xbm_{(n),j}^0 - g^{(s)}\left( \Xbm_{(n),\mathcal{M}^{(s)}}^0 \right)$ is independent of the variables on the second line. Consequently,
\begin{eqnarray*}
	\Var\left( \eta_{\widehat{b}^{(s)},1,\tau}^{(s)} \right) & \le &
	O(1)q^2 \Mean \left[\frac{1}{M}\sum_{m=1}^M h_{\widehat{b}^{(s)}}^{(s)}\left( \widetilde{\Xbm}_{(n),k}^{(m)},\Xbm_{(n),\mathcal{M}^{(s)}}^0 \right) \right. \\
	& & \quad\quad \left. - \Mean \left\{ h_{\widehat{b}^{(s)}}^{(s)}\left( \Xbm_{(n),k}^0,\Xbm_{(n),\mathcal{M}^{(s)}}^0 \right) | \; \Xbm_{(n),\mathcal{M}^{(s)}}^0 \right\}\right]^2,
\end{eqnarray*}
where $O(1)$ denotes some positive constant. Since
\begin{align*}
	& \Mean \left[\frac{1}{M}\sum_{m=1}^M h_{\widehat{b}^{(s)}}^{(s)}\left( \widetilde{\Xbm}_{(n),k,m}^{(s)},\Xbm_{(n),\mathcal{M}^{(s)}}^0 \right) - \Mean \left\{h_{\widehat{b}^{(s)}}^{(s)}\left( X_{(n),k}^0,\Xbm_{(n),\mathcal{M}^{(s)}}^0 \right) | \Xbm_{(n),\mathcal{M}^{(s)}}^0 \right\}\right]^2 \\ 
	= \; & \Mean \left[ \Var\left\{ \left.\frac{1}{M}\sum_{m=1}^M h_{\widehat{b}^{(s)}}^{(s)}\left( \widetilde{\Xbm}_{(n),k,m}^{(s)},\Xbm_{(n),\mathcal{M}^{(s)}}^0 \right) \right | \Xbm_{(n),\mathcal{M}^{(s)}}^0 \right\} \right] \\
	& + \Mean \left[\Mean \left\{h_{\widehat{b}^{(s)}}^{(s)}\left( \widetilde{\Xbm}_{(n),k,m}^{(s)},\Xbm_{(n),\mathcal{M}^{(s)}}^0 \right)-h_{\widehat{b}^{(s)}}^{(s)}\left( X_{(n),k}^0,\Xbm_{(n),\mathcal{M}^{(s)}}^0 \right) | \Xbm_{(n),\mathcal{M}^{(s)}}^0 \right\}\right]^2.
\end{align*}
By the boundedness of $\mathbb{H}^{(s)}$ and that $M$ is proportional to $NT$, the second line is of the order $O(N^{-1} T^{-1})$. The third line is of the order $O_p\{(NT)^{-2\kappa_2}\}$ under (C2). Without loss of generality, suppose $\kappa_2\le1$. It follows that $\Var(\eta_{\widehat{b}^{(s)},1,\tau}^{(s)})=O_p\{q^2 (NT)^{-2\kappa_2} \}$. Consequently, $\delta_1 = O_p\left\{ q^{1/2} (NT)^{-1/2-\kappa_2} \right\}$.

Putting together the bounds for $\delta_1$ and $\delta_2$, we have that,
\begin{eqnarray*}
	\left| \eta_{\widehat{b}^{(s)},1}^{(s)} \right| = O_p\left\{ q^{1/2} (NT)^{-1/2-\kappa_2} \right\},
\end{eqnarray*}
with probability at least $1-Q\beta(q)/q$. Since $\beta(q)=O(q^{-\kappa_3})$, set $q$ to be proportional to $\{(NT)\log(NT)\}^{1/(1+\kappa_3)}$. It then follows that $Q\beta(q)/q=O\{\log^{-1} (NT)\}\to 0$. In addition, since $\kappa_3>\{2\min(\kappa_1,\kappa_2)\}^{-1}-1$, we obtain $\left|\eta_{\widehat{b}^{(s)},1}^{(s)} \right| = o_p\{(NT)^{-1/2}\}$. This completes Step 1.

\smallskip
\noindent
\textbf{Step 2}.
This step is derived similarly as Step 1, and the details are omitted.

\smallskip
\noindent \textbf{Step 3}. Following similar arguments as in Step 1, we can show that
\begin{eqnarray*}
	\left| \eta_{\widehat{b}^{(s)},3}^{(s)} - \Mean \eta_{\widehat{b}^{(s)},3}^{(s)} \right| = o_p\left\{ (NT)^{-1/2} \right\}.
\end{eqnarray*}
It then suffices to show $\Mean \eta_{\widehat{b}^{(s)},3}^{(s)} = o_p\{(NT)^{-1/2}\}$, or equivalently, $\delta_3 = o_p\{(NT)^{-1/2}\}$, where
\begin{eqnarray*}\label{eqn1}
	\delta_3 & \equiv & \max_{b\in \{1,\ldots,B\}} \left| \Mean \left\{ g^{(s)}\left( \Xbf_{\mathcal{M}^{(s)}} \right) - \widehat{g}^{(s)}\left( \Xbf_{\mathcal{M}^{(s)}} \right) \right\} \right. \\
	& & \quad\quad \times \left. \Mean \left\{ h_b^{(s)}\left( \widetilde{\Xbm}_{k}^{(m)},\Xbf_{\mathcal{M}^{(s)}} \right) - h_b^{(s)}\left( X_{k},\Xbf_{\mathcal{M}^{(s)}} \right) | \; \Xbf_{\mathcal{M}^{(s)}} \right\} \right|.	
\end{eqnarray*}
By Cauchy-Schwarz inequality, we have that,
\begin{eqnarray*}
	\delta_3 & \le & \sqrt{\Mean |g^{(s)}(\Xbf_{\mathcal{M}^{(s)}})-\widehat{g}^{(s)}(\Xbf_{\mathcal{M}^{(s)}})|^2} \\
	& & \times \max_{b\in \{1,\ldots,B\}}\sqrt{\Mean \left|\Mean [\{h_b^{(s)}(\widetilde{\Xbm}_{k}^{(m)},\Xbf_{\mathcal{M}^{(s)}})-h_b^{(s)}(X_{k},\Xbf_{\mathcal{M}^{(s)}})\}|\Xbf_{\mathcal{M}^{(s)}}]\right|^2},
\end{eqnarray*}
where the first term on the right-hand-side is $O\{(NT)^{-\kappa_1}\}$ by condition (C2), and the second term is $O\{(NT)^{-\kappa_2}\}$ by condition (C2). Since $\kappa_1+\kappa_2>1/2$, we have $\delta_3 = o_p\{(NT)^{-1/2}\}$. This completes Step 3.

\smallskip
\noindent
\textbf{Step 4}.
In this step, we aim to establish \eqref{eqn:step4} for $\widehat{I}^{(s)}_{\widehat{b}^{(s)},\textrm{CF}}$ under the bidirectional asymptotic framework. Conditional on the data $O_s$, the index $\widehat{b}^{(s)}$ is fixed. We next show that \eqref{eqn:step4} holds under two scenarios, one with $N$ bounded, and the other with $N$ diverging.

\smallskip
\noindent
\textbf{Scenario 4.1}: $N$ is bounded and $T \to \infty$.
Condition (C3) implies that each $\{\Xbm_{i,t}\}_{t}$ is strong mixing. Since $X_j$ has the bounded fourth moment, and $\mathbb{H}$ is a bounded function class, it follows from \cite[Equation (1.12b)]{Rio2013} that $\Cov\left( I_{i,t,\widehat{b}^{(s)}}^{(s)*}, I_{i,t+q,\widehat{b}^{(s)}}^{(s)*} \right) = O(\beta^{1/2}(q))$, with respect to $q$. Since $\beta(q)=O(q^{-\kappa_3})$ and $\kappa_3>2$, it follows that $\Cov\left( I_{i,t,\widehat{b}^{(s)}}^{(s)*}, I_{i,t+q,\widehat{b}^{(s)}}^{(s)*} \right)$ decays at the rate of $q^{-\kappa_3^*}$ for some $\kappa_3^*>1$. Consequently,
\begin{eqnarray}\label{eqn:finite}
	\sum_{q=-\infty}^{+\infty}\Cov\left( I_{i,t,\widehat{b}^{(s)}}^{(s)*}, I_{i,t+q,\widehat{b}^{(s)}}^{(s)*} \right) < +\infty.
\end{eqnarray}
For each $i\in \mathcal{I}_{\ell}^c$, the process $T^{-1} \sum_{1\le t\le T} I_{i,t,\widehat{b}^{(s)}}^{(s)*}$ meets the requirements of Theorem 3 in \cite{Kourogenis2011}. Consequently, for each $i\in \mathcal{I}_{\ell}^c$,
\begin{eqnarray}\label{eqn:weakconvergence}
	\frac{\sum_{1\le t\le T} I_{i,t,\widehat{b}^{(s)}}^{(s)*}}{\sqrt{\Var(\sum_{1\le t\le T} I_{i,t,\widehat{b}^{(s)}}^{(s)*})}}\stackrel{d}{\to} N(0,1).
\end{eqnarray}
Since the processes $\sum_{1\le t\le T} I_{1,t,\widehat{b}^{(s)}}^{(s)*},\ldots,\sum_{1\le t\le T} I_{N,t,\widehat{b}^{(s)}}^{(s)*}$ are i.i.d., we have,
\begin{eqnarray*}
	\Mean \exp\left\{ iu \frac{\sum_{i\in \mathcal{I}_{\ell}^c}\sum_{1\le t\le T} I_{i,t,\widehat{b}^{(s)}}^{(s)*}}{\sqrt{N\Var\left( \sum_{1\le t\le T} I_{i,t,\widehat{b}^{(s)}}^{(s)*} \right)/2}} \right\} = \left[\Mean  \exp\left\{ iu \frac{\sum_{1\le t\le T} I_{1,t,\widehat{b}^{(s)}}^{(s)*}}{\sqrt{N\Var\left( \sum_{1\le t\le T} I_{1,t,\widehat{b}^{(s)}}^{(s)*} \right)/2}} \right\}\right]^{N/2}.
\end{eqnarray*}
Since $N$ is bounded, it follows from \eqref{eqn:weakconvergence} that
\begin{eqnarray*}
	\Mean \exp\left\{ iu \frac{\sum_{i\in \mathcal{I}_{\ell}^c}\sum_{1\le t\le T} I_{i,t,\widehat{b}^{(s)}}^{(s)*}}{\sqrt{N\Var\left( \sum_{1\le t\le T} I_{i,t,\widehat{b}^{(s)}}^{(s)*} \right)/2}} \right\}\stackrel{d}{\to} \left\{\exp\left(-\frac{u^2}{N}\right)\right\}^{N/2}=\exp(-u^2/2),
\end{eqnarray*}
for any $u$. This completes the proof for this scenario.

\smallskip
\noindent
\textbf{Scenario 4.2}: $N \to \infty$.
We apply the Lindeberg central limit theorem for triangle arrays to derive our results. It suffices to verify the Lindeberg's condition, i.e.,
\begin{align*}
	\frac{2}{N\Var\left( \sum_{1\le t\le T} I_{1,t,\widehat{b}^{(s)}}^{(s)*} \right)} & \sum_{i\in \mathcal{I}_{\ell}^c} \Mean \left(\sum_{1\le t\le T}I_{i,t,\widehat{b}^{(s)}}^{(s)*} \right)^2 \\
	\times \mathbb{I}\left\{ \left| \sum_{1\le t\le T}I_{i,t,\widehat{b}^{(s)}}^{(s)*} \right| \right. & \left. \ge \epsilon \sqrt{N\Var\left( \sum_{1\le t\le T} I_{1,t,\widehat{b}^{(s)}}^{(s)*} \right)/2}\right\}\to 0,
\end{align*}
for any $\epsilon>0$, where $\mathbb{I}\{\cdot\}$ denotes the indicator function.

Under the conditions of Theorem \ref{thm2}, we have that,
\begin{eqnarray}\label{eqn:step0}
	\Var\left( \sqrt{NT} \widehat{I}_{\widehat{b}^{(s)},\textrm{CF}}^{(s)*} | O_s \right) \ge \kappa_4/2,
\end{eqnarray}
with probability tending to $1$. By \eqref{eqn:step0}, and that $\sum_{1\le t\le T} I_{1,t,\widehat{b}^{(s)}}^{(s)*},\sum_{1\le t\le T}$ $I_{2,t,\widehat{b}^{(s)}}^{(s)*},\ldots$, $\sum_{1\le t\le T} I_{N,t,\widehat{b}^{(s)}}^{(s)*}$ are identically distributed, it suffices to show
\begin{eqnarray*}
	\frac{4}{\kappa_4T} \Mean \left( \sum_{1\le t\le T}I_{1,t,\widehat{b}^{(s)}}^{(s)*} \right)^2 \mathbb{I}\left( \left| \sum_{1\le t\le T}I_{1,t,\widehat{b}^{(s)}}^{(s)*} \right| \ge \epsilon \sqrt{\kappa_4NT/4} \right) \to 0,
\end{eqnarray*}
for any $\epsilon>0$, or equivalently,
\begin{eqnarray*}
	T^{-1} \Mean \left( \sum_{1\le t\le T}I_{1,t,\widehat{b}^{(s)}}^{(s)*} \right)^2 \mathbb{I}\left( \left| \sum_{1\le t\le T}I_{1,t,\widehat{b}^{(s)}}^{(s)*} \right|>\epsilon N^{1/2}T^{1/2} \right) \to 0.
\end{eqnarray*}

By \eqref{eqn:finite}, we have $\Mean \left(\sum_{1\le t\le T}I_{1,t,\widehat{b}^{(s)}}^{(s)*} \right)^2 = T\Mean \left(\sum_{1\le t\le T}I_{1,1,\widehat{b}^{(s)}}^{(s)*} \right)^2 + O(T) = O(T)$. By the dominated convergence theorem, it suffices to show
\begin{eqnarray*}
	T^{-1} \left( \sum_{1\le t\le T} I_{1,t,\widehat{b}^{(s)}}^{(s)*} \right)^2 \mathbb{I}\left( \left| I_{1,t,\widehat{b}^{(s)}}^{(s)*} \right| > \epsilon N^{1/2}T^{1/2} \right) = o_p(1),
\end{eqnarray*}
or equivalently,
\vspace{-0.01in}
\begin{eqnarray}\label{eqn:prob}
	\prob\left( \left| \sum_{1\le t\le T}I_{1,t,\widehat{b}^{(s)}}^{(s)*} \right| > \epsilon N^{1/2}T^{1/2} \right) \to 0.
\end{eqnarray}
By Chebyshev's inequality, \eqref{eqn:prob} holds, because
\begin{eqnarray*}
	\prob\left( \left| \sum_{1\le t\le T}I_{1,t,\widehat{b}^{(s)}}^{(s)*} \right| > \epsilon N^{1/2}T^{1/2} \right) \le \frac{\Mean \left| \sum_{1\le t\le T}I_{1,t,\widehat{b}^{(s)}}^{(s)*} \right|^2}{\epsilon^2 NT}=O(N^{-1})=o(1),
\end{eqnarray*}
as $N$ diverges to infinity. This completes Step 4.

\smallskip
\noindent
\textbf{Step 5}.
In this step, we establish the consistency of the batched mean estimator. We consider three scenarios, when $N$ is bounded and $T \to \infty$, when $T$ is bounded and $N \to \infty$, and when both $N, T \to \infty$.

\smallskip
\noindent
\textbf{Scenario 5.1}: $N$ is bounded and $T \to \infty$.
Note that
\begin{eqnarray*}
	& & \widehat{\sigma}_{\widehat{b}^{(s)},\textrm{CF}}^2  =  \frac{2K}{NT}\sum_{i\in \mathcal{I}_{\ell}^c}\sum_{k=1}^{T/K}\left\{ \frac{\sum_{t=(k-1)K+1}^{kK} \left( I_{i,t,\widehat{b}^{(s)}}^{(s)}-\widehat{I}_{\widehat{b}^{(s)},\textrm{CF}}^{(s)} \right) }{\sqrt{K}} \right\}^2 \\
	&=& \frac{2K}{NT}\sum_{i\in \mathcal{I}_{\ell}^c}\sum_{k=1}^{T/K}\left\{ \frac{\sum_{t=(k-1)K+1}^{kK} \left( I_{i,t,\widehat{b}^{(s)}}^{(s)}-\Mean \widehat{I}_{\widehat{b}^{(s)},\textrm{CF}}^{(s)} \right) }{\sqrt{K}} \right\}^2 - K \left\{ \widehat{I}_{\widehat{b}^{(s)},\textrm{CF}}^{(s)} - \Mean \left(\widehat{I}_{\widehat{b}^{(s)},\textrm{CF}}^{(s)} \right) \right\}^2 \\
	& \equiv & \delta_4 - \delta_5.
\end{eqnarray*}
Following similar arguments as in Step 4, we can show that
\begin{eqnarray*}
	\widehat{I}_{\widehat{b}^{(s)},\textrm{CF}}^{(s)}-\Mean \left(\widehat{I}_{\widehat{b}^{(s)},\textrm{CF}}^{(s)} \right) = O_p\{(NT)^{-1/2}\}.
\end{eqnarray*}
Since $K\ll NT$, we have $\delta_5 = o_p(1)$. Consequently, it suffices to show that
\begin{eqnarray}\label{eqn:key1}
	\begin{split}	
		\delta_4 \stackrel{P}{\to} \frac{NT}{2} \Var\left( \widehat{I}_{\widehat{b}^{(s)},\textrm{CF}}^{(s)} \right) = \frac{1}{T}\Var\left( \sum_{t=1}^T I_{i,t,\widehat{b}^{(s)}} \right).
	\end{split}	
\end{eqnarray}
Since $N$ is bounded and $K\gg T^{1/(1+\kappa_3)}$, we have $K\gg (NT)^{1/(1+\kappa_3)}$. Without loss of generality, suppose $T/K$ is divisible by 2. 
Following similar arguments as in Step 1, we approximate $\delta_4$ by
\begin{eqnarray*}
	&&\frac{K}{NT}\sum_{i\in \mathcal{I}_{\ell}^c}\sum_{k=1}^{T/(2K)}\underbrace{\left\{ \frac{\sum_{t=(2k-2)K+1}^{(2k-1)K} \left( I_{i,t,\widehat{b}^{(s)}}^{(s)0}-\Mean \widehat{I}_{\widehat{b}^{(s)},\textrm{CF}}^{(s)} \right) }{\sqrt{K}} \right\}^2}_{\phi_{i,k,1}^{(s)}}\\
	&+&\frac{K}{NT}\sum_{i\in \mathcal{I}_{\ell}^c}\sum_{k=1}^{T/(2K)}\underbrace{\left\{ \frac{\sum_{t=(2k-1)K+1}^{2kK} \left( I_{i,t,\widehat{b}^{(s)}}^{(s)0}-\Mean \widehat{I}_{\widehat{b}^{(s)},\textrm{CF}}^{(s)} \right) }{\sqrt{K}} \right\}^2}_{\phi_{i,k,2}^{(s)}},
\end{eqnarray*}
with probability tending to $1$, where $\left\{ I_{i,t,\widehat{b}^{(s)}}^{(s)0} \right\}_{i,t}$ denotes the version of $\left\{ I_{i,t,\widehat{b}^{(s)}}^{(s)} \right\}_{i,t}$ such that $\left\{ \phi_{i,k,m}^{(s)} \right\}_{i,k,m}$ are independent across different pairs $(i,k)$ for any $m=1,2$. By condition (C3), using the weak law of large numbers, $\delta_4$ converge in probability to
\begin{eqnarray}\label{eqn:I}
	\Mean \left\{ \frac{\sum_{t=1}^{K} \left( I_{1,t,\widehat{b}^{(s)}}^{(s)}-\Mean \widehat{I}_{\widehat{b}^{(s)},\textrm{CF}}^{(s)} \right) }{\sqrt{K}} \right\}^2=\Mean \left\{ \frac{\sum_{t=1}^{K} \left( I_{1,t,\widehat{b}^{(s)}}^{(s)}-\Mean I_{1,t,\widehat{b}^{(s)}}^{(s)} \right) }{\sqrt{K}} \right\}^2.
\end{eqnarray}
Similar to \eqref{eqn:finite}, we can show that both the right-hand-side of \eqref{eqn:I} and $T^{-1} \Var\left( \sum_{t=1}^T I_{i,t,\widehat{b}^{(s)}}^{(s)} \right)$ are bounded. In addition, their difference is asymptotically negligible as $K$ and $T$ increases to infinity. This yields \eqref{eqn:key1}, and completes the proof for this scenario.

\smallskip
\noindent
\textbf{Scenario 5.2}: $T$ is bounded and $N \to \infty$.
By condition (C4), we have $K=T$ under this setting. Then $\Mean \widehat{\sigma}_{\widehat{b}^{(s)},\textrm{CF}}^2$ is nearly unbiased to the variance of $\sqrt{(NT)/2}\widehat{I}_{\widehat{b}^{(s)},\textrm{CF}}^{(s)}$. The consistency follows from the law of large numbers. This completes the proof for this scenario.

\smallskip
\noindent
\textbf{Scenario 5.3}: Both $T$ and $N$ diverge to infinity.
It suffices to show \eqref{eqn:key1}. Since $N$ diverges to infinity, $\delta_4$ converges to
\begin{eqnarray*}
	\Mean \left\{ \frac{\sum_{t=1}^{K} \left( I_{1,t,\widehat{b}^{(s)}}^{(s)}-\Mean \widehat{I}_{\widehat{b}^{(s)},\textrm{CF}}^{(s)} \right) }{\sqrt{K}} \right\}^2=\Mean \left\{ \frac{\sum_{t=1}^{K} \left( I_{1,t,\widehat{b}^{(s)}}^{(s)}-\Mean I_{1,t,\widehat{b}^{(s)}}^{(s)} \right) }{\sqrt{K}} \right\}^2.
\end{eqnarray*}
Following similar arguments as in Scenario 5.1, we can show \eqref{eqn:key1} holds. This completes Step 5.

\smallskip
\noindent
\textbf{Step 6}. Putting together the results that $\eta_{\widehat{b}^{(s)},l}^{(s)} = o_p\{(NT)^{-1/2}\}$, $l=1,2,3$, we obtain that $\left| \widehat{I}_{\widehat{b}^{(s)},\textrm{CF}}^{(s)}-\widehat{I}_{\widehat{b}^{(s)},\textrm{CF}}^{(s)*} \right| = o_p\{(NT)^{-1/2}\}$. Following similar arguments, we can show that 
\[
\left| \Var\left( \sqrt{NT} \widehat{I}_{\widehat{b}^{(s)},\textrm{CF}}^{(s)*} | O_s \right) - \Var\left( \sqrt{NT} \widehat{I}_{\widehat{b}^{(s)},\textrm{CF}}^{(s)}|O_s \right) \right| = o_p(1). 
\]
By \eqref{eqn:step0}, we have that,
\begin{eqnarray*} \label{eqn:steps1-3}
	\frac{ \widehat{I}_{\widehat{b}^{(s)},\textrm{CF}}^{(s)*}- \widehat{I}_{\widehat{b}^{(s)},\textrm{CF}}^{(s)}}{\sqrt{\Var\left( \widehat{I}_{\widehat{b}^{(s)},\textrm{CF}}^{(s)*}|O_s \right)}}=o_p\{(NT)^{-1/2}\}.
\end{eqnarray*}
Note that, under $\mathcal{H}_0(j,k)$, $\Mean \left( \widehat{I}^{(s)*}_{\widehat{b}^{(s)},\textrm{CF}}|O_s \right) = 0$. By Step 4, we have that, conditional on $O_s$, \eqref{eqn:step4} holds. Since the limiting distribution is independent to the data $O_s$, \eqref{eqn:step4} also holds unconditionally. By Step 5, we have that $\left( \widehat{\sigma}_{\widehat{b}^{(s)},\textrm{CF}}^{(s)} \right)^2$ is consistent to the conditional variance of $\sqrt{(NT)/2} \widehat{I}_{\widehat{b}^{(s)},\textrm{CF}}^{(s)}$. As $\sqrt{(NT)/2} \widehat{I}_{\widehat{b}^{(s)},\textrm{CF}}^{(s)}$ and $\sqrt{(NT)/2} \widehat{I}_{\widehat{b}^{(s)},\textrm{CF}}^{(s)*}$ are asymptotically negligible, we can show that $\left( \widehat{\sigma}_{\widehat{b}^{(s)},\textrm{CF}}^{(s)} \right)^2$ is consistent to the conditional variance of $\sqrt{(NT)/2} \widehat{I}_{\widehat{b}^{(s)},\textrm{CF}}^{(s)*}$ as well. By Slutsky's theorem, we have that,
\begin{eqnarray*}
	\frac{\sqrt{(NT)/2} \widehat{I}_{\widehat{b}^{(s)},\textrm{CF}}^{(s)*}}{\widehat{\sigma}_{\widehat{b}^{(s)},\textrm{CF}}^{(s)}}\stackrel{d}{\to} N(0,1),
\end{eqnarray*}
or equivalently, $\widehat{T}^{(s)}_{\widehat{b}^{(s)},\textrm{CF}}\stackrel{d}{\to} N(0,1)$. This completes the proof of Theorem \ref{thm2}.
\eop

\subsection{Proof of Theorem \ref{thm3}}
\label{sec:proofthm3}

We first introduce the notion of the VC type class \citep[][Definition 2.1]{Cherno2014}. Specifically, let $\mathcal{F}$ denote a class of measurable functions, with a measurable envelope function $F$ such that $\sup_{f\in \mathcal{F}}|f| \le F$. For any probability measure $Q$, let $e_Q$ denote a semi-metric on $\mathcal{F}$ such that $e_Q(f_1,f_2)=\|f_1-f_2\|_{Q,2} = \sqrt{\int |f_1-f_2|^2 dQ}$. An $\epsilon$-net of the space $(\mathcal{F}, e_Q)$ is a subset $\mathcal{F}_{\epsilon}$ of $\mathcal{F}$, such that for every $f\in \mathcal{F}$, there exists some $f_{\epsilon}\in \mathcal{F}_{\epsilon}$ satisfying $e_Q(f,f_{\epsilon}) < \epsilon$. We say that $\mathcal{F}$ is a VC type class with envelope $F$, if there exist constants $c_0 > 0, c_1 \ge 1$, such that $\sup_Q \mathbb{N}\left( \mathcal{F},e_Q, \epsilon\|F\|_{Q,2} \right) \le (c_0 / \epsilon)^{c_1}$, for all $0 < \epsilon \le 1$, where the supremum is taken over all finitely discrete probability measures on the support of $\mathcal{F}$, and $\mathbb{N}\left( \mathcal{F},e_Q, \epsilon\|F\|_{Q,2} \right)$ is the infimum of the cardinality of $\epsilon\|F\|_{Q,2}$-nets of $\mathcal{F}$. We refer to $c_1$ as the VC index of $\mathcal{F}$.

We next present the proof. Throughout the proof, we assume the indices of the data subsets $\mathcal{I}_s$ and $\mathcal{I}_s^c$ are fixed, and show the $p$-value converges to $1$ in probability, given $\mathcal{I}_s$ and $\mathcal{I}_s^c$. As such, unconditionally, the $p$-value converges to $1$ in probability as well. We begin with a definition,
\begin{eqnarray*}
	\widehat{I}_{b,\textrm{NCF}}^{(s)*} = 2(NT)^{-1}\sum_{i\in \mathcal{I}_{\ell}} \sum_{1\le t\le T}I_{i,t,b}^{(s)*},
\end{eqnarray*}
where $I_{i,t,b}^{(s)*}$ is as defined in the proof of Theorem \ref{thm2}.  Note that $g^{(s)}$ depends on $s$ only through the set $\widehat{\textrm{AC}}_j^{(s)}$. Thus, we use the notation $g_{j,k,\mathcal{M}}$ to denote $g^{(s)}$. For a given set $\mathcal{M}$, define
\begin{eqnarray*}
	\zeta_{b,1}^{(s)}(\mathcal{M}) & = & \frac{2}{NT}\sum_{i\in \mathcal{I}_{\ell}}\sum_{1\le t\le T} \left\{ \Xbm_{i,t,j}-g_{j,k,\mathcal{M}}(\Xbm_{i,t,\mathcal{M}}) \right\} \\
	& & \times \left[ \frac{1}{M}\sum_{m=1}^M h_b^{(s)}\left( \widetilde{\Xbm}_{i,t,k}^{(s,m)},\Xbm_{i,t,\mathcal{M}} \right) - \Mean \left\{ h_b^{(s)}\left( \Xbm_{i,t,k},\Xbm_{i,t,\mathcal{M}} \right) | \; \Xbm_{i,t,\mathcal{M}} \right\} \right],\\
	\zeta_{b,2}^{(s)}(\mathcal{M}) & = & \frac{2}{NT}\sum_{i\in \mathcal{I}_{\ell}}\sum_{1\le t\le T} \left\{ g_{j,k,\mathcal{M}}(\Xbm_{i,t,\mathcal{M}})-\widehat{g}^{(s)}(\Xbm_{i,t,\mathcal{M}}) \right\} \\
	& & \times \left[ h_b^{(s)}(\Xbm_{i,t,k},\Xbm_{i,t,\mathcal{M}})-\Mean \left\{ h_b^{(s)}(\Xbm_{i,t,k},\Xbm_{i,t,\mathcal{M}}) | \Xbm_{i,t,\mathcal{M}} \right\} \right],
\end{eqnarray*}
\begin{eqnarray*}
	\zeta_{b,3}^{(s)}(\mathcal{M}) & = & \frac{2}{NT}\sum_{i\in \mathcal{I}_{\ell}}\sum_{1\le t\le T} \left\{ g_{j,k,\mathcal{M}}(\Xbm_{i,t,\mathcal{M}})-\widehat{g}^{(s)}(\Xbm_{i,t,\mathcal{M}}) \right\} \\
	& & \times \left[ \frac{1}{M}\sum_{m=1}^M h_b^{(s)}\left( \widetilde{\Xbm}_{i,t,k}^{(s,m)},\Xbm_{i,t,\mathcal{M}} \right) - \Mean \left\{h_b^{(s)}\left( \Xbm_{i,t,k},\Xbm_{i,t,\mathcal{M}} \right) | \; \Xbm_{i,t,\mathcal{M}} \right\} \right].
\end{eqnarray*}
We next divide the proof of this theorem into 5 steps. In Steps 1 to 3, we show that $\max_{\mathcal{M}\in \mathbb{M}}\max_b$ $|\zeta_{b,l}^{(s)}(\mathcal{M}-\{k\})|=O_p\{(NT)^{-1/2}\log (NT)\}$ for $l=1,2,3$, respectively, where $\mathbb{M}$ denotes the class of subsets $\mathcal{M}$ that meets the requirements of Proposition \ref{prop1}. In Step 4, we show that
\begin{eqnarray}\label{eqn:thm3}
	\left| I\left( j,k|\widehat{\textrm{AC}}_j^{(s)};h^{(s)}_{\widehat{b}^{(s)}} \right) \right| \gg N^{-1/2} T^{-1/2},
\end{eqnarray}
with probability approaching one. In Step 5, we put all the above results together to complete the proof.

\smallskip
\noindent 
\textbf{Step 1}.
It suffices to show $\max_b \left| \zeta_{b,1}^{(s)}(\mathcal{M}-\{k\}) \right| = O_p\{(NT)^{-1/2}\log (NT)\}$ for any $\mathcal{M}\in \mathbb{M}$. To simplify the presentation, when there is no confusion, we write $\zeta_{b,l}^{(s)}(\mathcal{M}-\{k\})$ and $g_{j,k,\mathcal{M}-\{k\}}$  as $\zeta_{b,l}^{(s)}$ and $g_{j,k}$, respectively.

To bound $\max_b  |\zeta_{b,1}^{(s)}|$, we apply Lemma \ref{lemma:aux} (see Section \ref{sec:aux-lemma}). Note that $\widetilde{\Xbm}_{i,t,k,m}^{(s)}$ can be written as $\mathbb{G}^{(s)}\left( \Xbm_{i,t,\mathcal{M}-\{k\}}, Z_{j,k}^{(m)} \right)$. 
Note that the generator $\mathbb{G}^{(s)}$ belongs to a VC type class $\{f:f\in \mathcal{F}\}$ with a bounded envelop function $F$. Define the function,
\begin{eqnarray*}
	\tau_{b,f}(\Xbm_{i,t},Z_{i,t}) & = & \{\Xbm_{i,t,j}-\tsb{g}_{j,k}(\Xbm_{i,t,\mathcal{M}-\{k\}})\} \times \Mean \big\{ \cos(\omega_b f(\Xbm_{i,t,\mathcal{M}-\{k\}},Z_{i,t})) \\
	& & - \cos(\omega_b \Xbm_{i,t,k})|\Xbm_{i,t,\mathcal{M}-\{k\}} \big\}, \; \textrm{ for } 1 \le b \le B/2, \\
	\tau_{b,f}(\Xbm_{i,t},Z_{i,t}) & = & \{\Xbm_{i,t,j}-\tsb{g}_{j,k}(\Xbm_{i,t,\mathcal{M}-\{k\}})\} \times \Mean \big\{ \sin(\omega_b f(\Xbm_{i,t,\mathcal{M}-\{k\}},Z_{i,t})) \\
	& & - \sin(\omega_b \Xbm_{i,t,k})|\Xbm_{i,t,\mathcal{M}-\{k\}} \big\}, \; \textrm{ for } B/2 < b \le B.
\end{eqnarray*}
where $\{Z_{i,t}\}_{i,t}$ are i.i.d., and are independent of the observed data. Therefore,
\begin{eqnarray*}
	\max_b \left| \zeta_{b,1}^{(s)} \right| \le \max_b \sup_{f\in \mathcal{F}} \{2/(NT)\} \, \left| \sum_{i\in \mathcal{I}_{\ell}} \sum_{1\le t\le T} \tau_{b,f}(\Xbm_{i,t},Z_{i,t}) \right|.
\end{eqnarray*}

By Lemma A.6 of \cite{Cherno2014}, for each $b$, we can show the class of functions $\{\tau_{b,f}:f\in \mathcal{F} \}$ corresponds to a VC type class with envelop function uniformly bounded by $O(1) |\omega_b|$, where $O(1)$ denotes some positive constant. In addition, we have $\sup_{b,f} \Var(\tau_{b,f})=O\{(NT)^{-2\kappa_2}\}$ under the given conditions. By setting $q=\kappa \log (NT)$ with some proper choice of $\kappa$, it follows from the auxiliary Lemma \ref{lemma:aux} given in Section \ref{sec:aux-lemma}, and the given condition on the VC index that, we have, with probability at least $1-o\{(NT)^{-\kappa_7}\}$,
\vspace{-0.01in}
\begin{eqnarray*}
	\sup_{f\in \mathcal{F}} \frac{2}{NT}\left|\sum_{i\in \mathcal{I}_{\ell}} \sum_{1\le t\le T} \tau_{b,f}(\Xbm_{i,t},Z_{i,t})\right| & \le & O(1) \left[\frac{\omega^* \{\log (NT)+\log \omega^*\}}{NT}\right.\\
	& & \left.+\frac{\log (NT)+\sqrt{\log (NT)}\sqrt{\log \omega^*}}{\sqrt{NT}}\right].
\end{eqnarray*}
where $\omega^*=\max_{1\le b\le B} |\omega_b|$.

By Bonferroni's inequality and the condition that $B=O\{(NT)^{\kappa_7}\}$,
\begin{eqnarray*}\label{eqn2}
	& & \max_b \sup_{f\in \mathcal{F}} \frac{2}{NT}\left|\sum_{i\in \mathcal{I}_{\ell}} \sum_{1\le t\le T} \tau_{b,f}(\Xbm_{i,t},Z_{i,t})\right| \\
	& \le & O(1) \left[\frac{\omega^* \{\log (NT)+\log \omega^*\}}{NT}+\frac{\log (NT)+\sqrt{\log (NT)}\sqrt{\log \omega^*}}{\sqrt{NT}}\right] \\
	& \le & (NT)^{-1/2}\log (NT).
\end{eqnarray*}
The last inequality is due to the fact that, each $\omega_b$ is standard normal, and $B=O\{(NT)^{\kappa_7}\}$, therefore, $\omega^*=O_p\{\sqrt{\log (NT)}\}$. This yields that $\max_b |\zeta_{b,1}^{(s)}|=O_p\{(NT)^{-1/2}\log (NT)\}$, which completes Step 1.

\smallskip
\noindent
\textbf{Step 2}.
This step is derived similarly as Step 1, and the details are omitted

\smallskip
\noindent
\textbf{Step 3}.
Similar to Step 1, it suffices to bound $\max_b \left| \zeta_{b,3}^{(s)}(\mathcal{M}-\{k\}) \right|$ for each $\mathcal{M} \in \mathbb{M}$. By Cauchy-Schwarz inequality and following similar arguments as in the proof of Theorem 3 of \cite{CausalMARL}, we have, up to some logarithmic terms,
\begin{eqnarray*}
	\sqrt{\sum_{i\in \mathcal{I}_{\ell},1\le t\le T} \Mean \left| g_{j,k}\left( \Xbm_{i,t,\mathcal{M}-\{k\}} \right) - \widehat{g}_{i,k}^{(s)}\left( \Xbm_{i,t,\mathcal{M}-\{k\}} \right)\right|^2} \le O\{(NT)^{1/2-2\kappa_1}\}, \\
	\max_b \sqrt{\sum_{i,t} \Mean \left| \left\{h_b^{(s)}\left( \widetilde{\Xbm}_{i,t,k,m}^{(s)},\Xbm_{i,t,\mathcal{M}-\{k\}} \right) - h_b^{(s)}\left( \Xbm_{i,t,k},\Xbm_{i,t,\mathcal{M}-\{k\}} \right) \right\} | \; \Xbm_{i,t,\mathcal{M}-\{k\}} \right|^2} \\
	\le O\{(NT)^{1/2-2\kappa_2}\}.
\end{eqnarray*}
Under the condition that $\kappa_1+\kappa_2>1/2$, we obtain that $\max_b \left| \zeta_{b,3}^{(s)}(\mathcal{M}-\{k\}) \right|$ $= o_p\{(NT)^{-1/2}\}$ for each $\mathcal{M}\in \mathbb{M}$, which completes Step 3.

\smallskip
\noindent
\textbf{Step 4}.
Based on the results from Steps 1-3, we obtain that 
\[
\max_b \Big| \widehat{I}^{(s)}_{b,\textrm{NCF}}-\widehat{I}^{(s)*}_{b,\textrm{NCF}} \Big| \le \max_{\mathcal{M}\in \mathbb{M}}\max_b  |\zeta_{b,l}^{(s)}(\mathcal{M}-\{k\})|=O_p\{(NT)^{-1/2}\log (NT)\}. 
\]
In the proof of Theorem \ref{thm2}, we have shown that $\min_b \Var\left( \sqrt{NT} \widehat{I}_{b,\textrm{CF}}^{(s)*}|O_s \right) \ge \kappa_4/2$. Since $\Var\left( \sqrt{NT} \widehat{I}_{b,\textrm{CF}}^{(s)*}|O_s \right)$ depends on $O_s$ only though $\mathcal{M}^{(s)}$, we obtain that, with probability approaching one,
\begin{eqnarray}\label{eqn3}
	\min_b \Var(\sqrt{NT} \widehat{I}_{b,\textrm{CF}}^{(s)*}|\mathcal{M}^{(s)})\ge \kappa_4/2.
\end{eqnarray}
Following similar arguments as in the proof of the first three steps, we can show that 
\[
\max_b \left| \left( \widehat{\sigma}^{(s)}_{b,\textrm{NCF}} \right)^2 - \Var\left( \sqrt{NT/2} \widehat{I}_{b,\textrm{CF}}^{(s)*}|\mathcal{M}^{(s)} \right) \right| = o_p(1).
\]
Therefore,
$\min_b \widehat{\sigma}^{(s)}_{b,\textrm{NCF}}\ge \sqrt{\kappa_4}/4$. It then follows that,
\begin{eqnarray}\label{eqn4}\qquad
	\max_b\left|\frac{ \widehat{I}_{b,\textrm{NCF}}^{(s)*}}{\sqrt{NT\Var\left( \widehat{I}_{b,\textrm{CF}}^{(s)*}|\mathcal{M}^{(s)} \right)/2}} - \frac{ \widehat{I}_{b,\textrm{NCF}}^{(s)}}{\widehat{\sigma}^{(s)}_{b,\textrm{NCF}}}\right|=O_p\left\{\frac{\log (NT)}{\sqrt{NT}}\right\}.
\end{eqnarray}
Following similar arguments as in Step 1, we can show that $\max_b \bigg| \widehat{I}_{b,\textrm{NCF}}^{(s)*} -$ $I\left( j,k|\widehat{\textrm{AC}}_j^{(s)};h_b^{(s)} \right) \bigg| = O_p\{(NT)^{-1/2}\log (NT)\}$. This together with \eqref{eqn3} and \eqref{eqn4} yields that,
\begin{eqnarray}\label{eqn5}\qquad
	\max_b\left|\frac{ I\left( j,k|\widehat{\textrm{AC}}_j^{(s)};h_b^{(s)} \right)}{\sqrt{NT\Var\left( \widehat{I}_{b,\textrm{CF}}^{(s)*}|\mathcal{M}^{(s)} \right)/2}}-\frac{ \widehat{I}_{b,\textrm{NCF}}^{(s)}}{\widehat{\sigma}^{(s)}_{b,\textrm{NCF}}}\right|=O_p\left\{\frac{\log (NT)}{\sqrt{NT}}\right\}.
\end{eqnarray}

Next, since $\Delta(\mathbb{H})\gg (NT)^{-1/2}\log (NT)$, there exists some $\omega_0$, such that one of the following two inequalities hold, $\min_{\mathcal{M}\in \mathbb{M}}I(j,k|\mathcal{M};\cos(\omega_0 \cdot ))\gg (NT)^{-1/2}\log (NT)$, or \\$\min_{\mathcal{M}\in \mathbb{M}}I(j,k|\mathcal{M}; \sin(\omega_0 \cdot ))\gg (NT)^{-1/2}\log (NT)$. Without loss of generality, suppose the former holds.

Note that the objective function $\min_{\mathcal{M}\in \mathbb{M}}I(j,k|\mathcal{M};\cos(\omega \cdot))$ is Lipschitz continuous in $\omega$, and any $\omega$ within the interval $[\omega_0-(NT)^{-1/2}\log (NT), \omega_0+(NT)^{-1/2}\log (NT)]$ satisfies that
\begin{eqnarray*}
	\min_{\mathcal{M}\in \mathbb{M}}I(j,k|\mathcal{M};\cos(\omega \cdot))\gg (NT)^{-1/2}\log (NT).
\end{eqnarray*}
Since each $\omega_b$ is normally distributed, the probability that $\omega_b$ falls into this interval is lower bounded by $c (NT)^{-1/2}\log (NT)$ for some constant $c>0$. Since we randomly generate $B/2$ many $\omega$, the probability that at least one of the $\omega$ falls into this interval is lower bounded by
\begin{eqnarray*}
	1-\{1-c (NT)^{-1/2}\log (NT)\}^{B/2}\ge 1-\exp\{-cB (NT)^{-1/2}\log (NT)/2\}.
\end{eqnarray*}
The above probability tends to $1$ under the condition that $B=\kappa_6 (NT)^{\kappa_7}$ for some $\kappa_7\ge 1/2$. Consequently, we obtain that,
\begin{eqnarray*}
	\max_{b} \min_{\mathcal{M}\in \mathbb{M}}I(j,k|\mathcal{M};h_b^{(s)})\gg (NT)^{-1/2}\log (NT).
\end{eqnarray*}
This together with \eqref{eqn3} yields that
\begin{eqnarray*}
	\max_b \left|\frac{ I\left( j,k|\widehat{\textrm{AC}}_j^{(s)};h_b^{(s)} \right)}{\sqrt{NT\Var\left( \widehat{I}_{b,\textrm{CF}}^{(s)*}|\mathcal{M}^{(s)} \right)/2}}\right|\gg (NT)^{-1/2}\log (NT).
\end{eqnarray*}
By \eqref{eqn5}, we have that,
\begin{eqnarray*}
	\max_b \left|\frac{ \widehat{I}_{b,\textrm{NCF}}^{(s)}}{\widehat{\sigma}^{(s)}_{b,\textrm{NCF}}}\right|\gg (NT)^{-1/2}\log (NT).
\end{eqnarray*}
By definition, we have that,
\begin{eqnarray*}
	\left|\frac{ \widehat{I}_{\widehat{b}^{(s)},\textrm{NCF}}^{(s)}}{\widehat{\sigma}^{(s)}_{\widehat{b}^{(s)},\textrm{NCF}}}\right|\gg (NT)^{-1/2}\log (NT).
\end{eqnarray*}
Using  \eqref{eqn5} again, we obtain that,
\begin{eqnarray*}
	\left|\frac{ I\left( j,k|\widehat{\textrm{AC}}_j^{(s)};h_{\widehat{b}^{(s)}}^{(s)} \right)}{\sqrt{NT\Var\left( \widehat{I}_{\widehat{b}^{(s)},\textrm{CF}}^{(s)*}|\mathcal{M}^{(s)} \right)/2}}\right|\gg (NT)^{-1/2}\log (NT).
\end{eqnarray*}
This together with \eqref{eqn3} yields \eqref{eqn:thm3}. This completes Step 4.

\smallskip
\noindent
\textbf{Step 5}.
Following similar arguments as in the proof of Steps 1-3 in Theorem \ref{thm2}, we can show that $\left| \Mean \left(\widehat{I}_{\widehat{b}^{(s)},\textrm{CF}}^{(s)}-\widehat{I}_{\widehat{b}^{(s)},\textrm{CF}}^{(s)*} | O_s \right) \right|=O_p\{(NT)^{-1/2}\}$. By \eqref{eqn:thm3}, we have $\left| \Mean \left( \widehat{I}_{\widehat{b}^{(s)},\textrm{CF}}^{(s)*} | O_s \right) \right| \gg N^{-1/2}T^{-1/2}$ with probability approaching one.
Following similar arguments as in the proof of Theorem \ref{thm2}, we have that $\sqrt{NT}\left\{ \widehat{I}_{\widehat{b}^{(s)},\textrm{CF}}^{(s)} - \Mean \left( \widehat{I}_{\widehat{b}^{(s)},\textrm{CF}}^{(s)}|O_s \right) \right\} = O_p\{(NT)^{-1/2}\}$. Therefore, $\sqrt{NT} \left| \widehat{I}_{\widehat{b}^{(s)},\textrm{CF}}^{(s)} \right|$ diverges to infinity with probability approaching one. Consequently, we obtain that $p^{(s)}(j,k)\stackrel{p}{\to} 0$ for each $s$. This completes the proof of Theorem \ref{thm3}.
\eop

\subsection{Proof of Theorem \ref{thm4}}\label{sec:proofthm4}

Under the acyclicity constraint in \eqref{eqn:DAG}, we have $\widehat{\thetabf}^{(s)}=\widetilde{\thetabf}^{(s)}(\widehat{\pi}^{(s)})$ for some ordering $\widehat{\pi}^{(s)}$. We aim to show $\widehat{\pi}^{(s)}\in \Pi^*$ with probability approaching one.

For any ordering $\pi$, define the objective function,
\begin{eqnarray*}
	\mathcal{L}(\pi)=\sum_{j=0}^{d-1} \inf_{f_j}\Mean \left\{ X_{j+1}-f_j\left( \Xbf_{\{\pi_1,\ldots,\pi_{j} \}} \right) \right\}^2,
\end{eqnarray*}
where the minimum is taken over all square integrable functions, and the function $f_0$ equals zero almost surely. It is straightforward to show that $\mathcal{L}(\pi)=\sum_{j=1}^d \mathcal{L}_j(\pi)$, where
\begin{eqnarray*}
	\mathcal{L}_j(\pi) = \Mean \left\{ X_{j}-\Mean\left( X_{j} | \Xbf_{\{\pi_1,\ldots,\pi_{j-1} \}} \right) \right\}^2.
\end{eqnarray*}
Let $\widehat{\mathcal{L}}(\pi)=\sum_{j=1}^d \widehat{\mathcal{L}}_j(\pi)$, where $\widehat{\mathcal{L}}_j(\pi)$ is the penalized least squares objective,
\begin{eqnarray*}
	\min_{\substack{\thetabf_j=(\Abf_j^{(1)},\ldots,\Abf_j^{(h)})\\ \scriptsize{\textrm{supp}}(\Abf_j^{(1)})\in \{\pi_1,\ldots,\pi_{j-1}\} } } \frac{2}{NT}\sum_{i\in \mathcal{I}_{\ell}} \sum_{1\le t\le T}\{\Xbm_{i,t,j}-\textrm{MLP}(\Xbm_{i,t};\thetabf_j)\}^2+\lambda \|\Abf_j^{(1)}\|_{1,1}.
\end{eqnarray*}

Note that, any ordering $\pi$ that minimizes the objective function $\mathcal{L}(\pi)$ belongs to $\Pi^*$. As such, there exists some $\epsilon>0$, such that
\begin{eqnarray}\label{eqn:eps}
	\mathcal{L}(\pi^*)\le \min_{\pi\notin \Pi^*} \mathcal{L}(\pi)-\epsilon,\,\,\,\,\forall \pi^*\in \Pi^*.
\end{eqnarray}
We next divide the proof of this theorem into 2 steps. In Step 1, we show that $\widehat{\mathcal{L}}(\pi^*)$ converges to $\mathcal{L}(\pi^*)$ for all $\pi^*\in \Pi^*$. In Step 2, we show that $ \widehat{\mathcal{L}}(\pi)\ge \mathcal{L}(\pi)+o_p(1)$ for all $\pi\notin \Pi^*$, which ultimately  leads to the conclusion of this theorem. Note that the DAG dimension $d$ is fixed in our proof.

\smallskip
\noindent
\textbf{Step 1}. It suffices to show $\widehat{\mathcal{L}}_j(\pi^*)=\mathcal{L}_j(\pi^*)+o_p(1)$, or equivalently, $|\widehat{\mathcal{L}}_j(\pi^*)-\mathcal{L}_j(\pi^*)| \le \epsilon$ for all $j = 1,\ldots,d$, $\pi^*\in \Pi^*$, and any sufficiently small $\epsilon>0$.

Fix an $0<\epsilon< 1$. Since $f_j$ is continuous, it follows from Stone-Weierstrass theorem that there exists a multivariate polynomial function $f_j^*$ such that the absolute value of the residual $f_j-f_j^*$ is uniformly bounded by $\epsilon/6$. Since $\pi^*\in \Pi^*$, $f_j^*(\Xbf)$ can be written as a function of $\Xbf_{\{\pi_1,\ldots,\pi_{j-1}\}}$.

By Theorem 1 of \cite{yarotsky2017error}, there exists a feedforward neural network with a bounded number of hidden units that uniformly approximates $f_j^*$, with the approximation error uniformly bounded by $\epsilon/6$ in absolute value. By Lemma 1 of \cite{farrell2018deep}, such a feedforward network can be embedded into an MLP with a bounded number of hidden units. Since we allow $H$ and $L$ to diverge, such an MLP can be further embedded into an MLP with $L-1$ layers and the widths of all layers being proportional to $H$. Denote this MLP by MLP$^*$, let $A_j^{(1)*},\ldots,A_j^{(L-1)*}$ denote the weight matrices at each layer, and $b_j^{(1)*},\cdots,b_j^{(L-1)*}$ the corresponding bias vectors. We can embed MLP$^*$ into another MLP, with $L$ layers, by setting $\Abf_j^{(l)}=\Abf_j^{(l-1)*}$ and $b_j^{(l)}=b_j^{(l-1)*}$ for $l=2,\ldots,L$, $\Abf_j^{(1)}$ such that its submatrix formed by columns in $\{\pi_1,\cdots,\pi_{j-1}\}$ and rows in $\{1,\cdots,j-1\}$ is set to an identity matrix and other entries are set to zero, and $\bbf_j^{(1)}$ to a zero vector. The resulting MLP satisfies $\|\Abf_j^{(1)}\|_{1,1}=j-1$, which is finite.  Therefore, $f_j$ can be approximated by an MLP with finite $\|\Abf_j^{(1)}\|_{1,1}=j-1$ such that the approximation error is uniformly bounded by $\epsilon/3$ in absolute value. In addition, its weight matrix in the first layer $\Abf_j^{(1)}$ satisfies that $\|\Abf_j^{(j)}\|_{1,1}=j-1$. In other words, there exists some $\thetabf_j$,  such that
\begin{eqnarray}\label{eqn:eq6}
	\left| \Mean(X_{j}|\Xbf_{\{\pi_1,\ldots,\pi_{j-1} \}}-\textrm{MLP}(\Xbf;\thetabf_j) \right| \le \epsilon/3,
\end{eqnarray}
almost surely, and that
\begin{eqnarray}\label{eqn:eq7}
	\textrm{supp}(\Abf_j^{(1)})\in \{\pi_1,\ldots,\pi_{j-1}\}\,\,\,\,\hbox{and}\,\,\,\,\|\Abf_j^{(1)}\|_{1,1}=j-1.
\end{eqnarray}
It follows from \eqref{eqn:eq6} that
\begin{eqnarray*}
	& & \left| \Mean \left\{ X_{j}-\textrm{MLP}(\Xbf;\thetabf_j) \right\}^2-\mathcal{L}_j(\pi) \right| \\
	& \le & \left| \Mean \left( X_{j}|\Xbf_{\{\pi_1,\ldots,\pi_{j-1} \}} \right) - \textrm{MLP}(\Xbf;\thetabf_j) \right|^2 \\
	& & + 2 \left| X_j-\Mean\left( X_j|\Xbf_{\{\pi_1,\ldots,\pi_{j-1}\}} \right) \right| \left| \Mean\left( X_{j}|\Xbf_{\{\pi_1,\ldots,\pi_{j-1} \}} \right) - \textrm{MLP}(\Xbf;\thetabf_j) \right| \\
	& \le & \epsilon^2/9+2(\epsilon/3)(1+\epsilon/3) < \epsilon.
\end{eqnarray*}
This together with \eqref{eqn:eq7} and the condition $\lambda\to 0$ yields that,
\begin{eqnarray*}
	\inf_{\substack{\thetabf_j=(\Abf_j^{(1)},\ldots,A_j^{(h)})\\ \scriptsize{\textrm{supp}}(\Abf_j^{(1)})\in \{\pi_1,\ldots,\pi_{j-1}\} } } \left[\Mean \sum_{1\le t\le T} \{ X_{j}-\textrm{MLP}(\Xbf;\thetabf_j) \}^2+\lambda \|\Abf_j^{(1)}\|_{1,1}\right]-\mathcal{L}_j(\pi)<\epsilon.
\end{eqnarray*}
By definition, we have that,
\begin{eqnarray*}
	\mathcal{L}_j(\pi)\le \inf_{\substack{\thetabf_j=(\Abf_j^{(1)},\ldots,A_j^{(h)})\\ \scriptsize{\textrm{supp}}(\Abf_j^{(1)})\in \{\pi_1,\ldots,\pi_{j-1}\} } } \left[\Mean \sum_{1\le t\le T}\{X_{j}-\textrm{MLP}(\Xbf;\thetabf_j)\}^2+\lambda \|\Abf_j^{(1)}\|_{1,1}\right].
\end{eqnarray*}
It follows that,
\begin{eqnarray*}
	\left|\inf_{\substack{\thetabf_j=(\Abf_j^{(1)},\ldots,A_j^{(h)})\\ \scriptsize{\textrm{supp}}(\Abf_j^{(1)})\in \{\pi_1,\ldots,\pi_{j-1}\} } } \left[\Mean \sum_{1\le t\le T}\{X_{j}-\textrm{MLP}(\Xbf;\thetabf_j)\}^2+\lambda \|\Abf_j^{(1)}\|_{1,1}\right]-\mathcal{L}_j(\pi)\right|<\epsilon.
\end{eqnarray*}
To show $|\widehat{\mathcal{L}}_j(\pi^*)-\mathcal{L}_j(\pi^*)|\le \epsilon$, it suffices to show that,
\begin{eqnarray}\nonumber
	\left|\inf_{\substack{\thetabf_j=(\Abf_j^{(1)},\ldots,A_j^{(h)})\\ \scriptsize{\textrm{supp}}(\Abf_j^{(1)})\in \{\pi_1,\ldots,\pi_{j-1}\} } } \left[\Mean \sum_{1\le t\le T}\{X_{j}-\textrm{MLP}(\Xbf;\thetabf_j)\}^2+\lambda \|\Abf_j^{(1)}\|_{1,1}\right]-\widehat{\mathcal{L}}_j(\pi)\right|\\\nonumber
	=o_p(1).
\end{eqnarray}
Under the conditions of the theorem, we can further restrict the parameter space to the class of $\thetabf_j$, such that $\textrm{MLP}(\cdot;\thetabf_j)$ is bounded by some constant. As such, the above is upper bounded by
\begin{eqnarray}\label{eqn:eq8}
	\begin{split}
		& \sup_{\substack{\thetabf_j=(\Abf_j^{(1)},\ldots,A_j^{(h)})\\ \scriptsize{\textrm{supp}}(\Abf_j^{(1)})\in \{\pi_1,\ldots,\pi_{j-1}\} } } \left|\Mean \sum_{1\le t\le T}\{X_{j}-\textrm{MLP}(\Xbf;\thetabf_j)\}^2\right.\\
		& -\left.\frac{2}{NT}\sum_{i\in \mathcal{I}_{\ell}} \sum_{1\le t\le T}\{\Xbm_{i,t,j}-\textrm{MLP}(\Xbm_{i,t};\thetabf_j^{(s)}(\pi^*))\}^2\right|,
	\end{split}
\end{eqnarray}
where the supremum is taken over all $\thetabf_j$ such that $\textrm{MLP}(\cdot;\thetabf_j)$ is bounded by some constant. It then suffices to show that \eqref{eqn:eq8} is $o_p(1)$. Following Step 1 of Theorem 2, we can first approximate \eqref{eqn:eq8} by a sum of independent random variables. This allows us to upper bounded \eqref{eqn:eq8} by $O\{\log (NT)\}$ many Radamacher complexity terms, under the exponential $\beta$-mixing condition. Following similar arguments as in Section A.2.2 of \cite{liang2018well}, each of these Radamacher complexity terms can be upper bounded by $O\{(NT)^{-\kappa_7}\}$ for some $\kappa_7>0$, under the given conditions on $L$ and $H$. This completes Step 1.

\smallskip
\noindent
\textbf{Step 2}.
Following similar arguments as in Step 1, we can show that
\begin{eqnarray*}
	\widehat{\mathcal{L}}_j(\pi)\ge \min_{\substack{\thetabf_j=(\Abf_j^{(1)},\ldots,A_j^{(h)})\\ \scriptsize{\textrm{supp}}(\Abf_j^{(1)})\le \{\pi_1,\ldots,\pi_{j-1}\} }} \Mean \{X_{j}-\textrm{MLP}(\Xbf;\thetabf_j)\}^2+\lambda\|\Abf_j^{(1)}\|_{1,1}-o_p(1),
\end{eqnarray*}
for any $\pi$ and $j = 1, \ldots, d$. Since the penalty term is non-negative, and the first term on the right-hand-side is lower bounded by $\mathcal{L}_j(\pi) = \Mean \left\{ X_{j+1} \right.$ $\left. -\Mean\left( X_{j+1}|\Xbf_{\{\pi_1,\ldots,\pi_j\}} \right)\right\}^2$, we obtain that,
\begin{eqnarray*}
	\widehat{\mathcal{L}}_j(\pi) \ge \mathcal{L}_j(\pi) - o_p(1),
\end{eqnarray*}
for any $\pi$ and $j = 1, \ldots, d$.

Since $d$ is fixed, so is the number of orderings. In view of \eqref{eqn:eps}, we obtain,
\begin{eqnarray*}
	\widehat{\mathcal{L}}(\pi^*)\le \min_{\pi\notin \Pi^*} \widehat{\mathcal{L}}(\pi)-\epsilon/2, \;\; \textrm{ for any } \; \pi^*\in \Pi^*,
\end{eqnarray*}
with probability approaching one. Note that $\widehat{\pi}^{(s)}$ minimizes the empirical objective function $\widehat{\mathcal{L}}(\pi)$. We thus obtain that $\widehat{\pi}^{(s)} \in \Pi^*$ with probability approaching one. This completes the proof of Theorem \ref{thm4}.
\eop

\subsection{An auxiliary lemma}
\label{sec:aux-lemma}

We present a useful lemma that is needed in Step 1 of the proof of Theorem \ref{thm3}. We first briefly introduce the setup. Let $\{Z_t:t\ge 0\}$ be a stationary $\beta$-mixing process with the $\beta$-mixing coefficient $\{\beta(q):q\ge 0\}$. Let $\mathcal{F}$ be a pointwise measurable class of functions that take $Z_t$ as input, and has a measurable envelope function $F$. For any $f\in \mathcal{F}$, suppose $\Mean \{f(Z_0)\} = 0$. Let $\sigma^2>0$ be a positive constant, such that $\sup_{f\in \mathcal{F}} \Mean \{f^2(Z_0)\} \le \sigma^2 \le \Mean \{F^2(Z_0)\}$. In the next lemma, we provide an exponential inequality for the empirical process $\sup_{f\in \mathcal{F}} \left| \sum_{t=0}^{T-1} f(Z_t) \right|$.

\begin{lemma}\label{lemma:aux}
	Suppose the envelop function is uniformly bounded by some constant $C>0$. In addition, suppose $\mathcal{F}$ belongs to the class of VC-type class such that $\sup_Q N(\mathcal{F}, e_Q, \varepsilon \|F\|_{Q,2})\le (A/\varepsilon)^{\nu}$ for some $A\ge e,\nu\ge 1$. Then there exist some constants $c_1,c_2>0$, such that
	\begin{eqnarray*}
		\prob\left(\sup_{f\in \mathcal{F}}\left|\sum_{t=0}^{T-1} f(Z_t)\right| > c_1\sqrt{\nu q\sigma^2T \log \left(\frac{AC}{\sigma}\right)} + c_1 \nu C \log \left(\frac{AC}{\sigma}\right) + c_1 q\tau + Cq \right) \\
		\le c_2 q \exp\left(-\frac{\tau^2q}{c_2T\sigma^2}\right) + c_2q \exp\left(-\frac{\tau}{c_2 C}\right)+\frac{T\beta(q)}{q},
	\end{eqnarray*}
	for any $\tau>0$ and $1\le q<T/2$.
\end{lemma}

\noindent
\textit{Proof}: We divide the proof of this lemma into three steps. In Step 1, we use Berbee's coupling lemma \citep[see Lemma 4.1 in][]{Dedecker2002} to approximate $\sup_{f\in \mathcal{F}}|\sum_{t=0}^{T-1} f(Z_t)|$ by the sum of i.i.d.\ variables. In Step 2, we apply the tail inequality in Lemma 1 of \cite{Adam2008} to bound the deviation between the empirical process and its mean. In Step 3, we apply the maximal inequality in Corollary 5.1 of \cite{Cherno2014} to bound the expectation of the empirical process.

\smallskip
\noindent
\textbf{Step 1}. 
Following the discussion below Lemma 4.1 of \cite{Dedecker2002},  we can construct a sequence of random variables $\{Z_{t}^0:t\ge 0\}$, such that
\begin{eqnarray}\label{eqn:step1eq1}
	\sup_{f\in \mathcal{F}}\left|\sum_{t=0}^{T-1} f(Z_t)\right|=\sup_{f\in \mathcal{F}}\left|\sum_{t=0}^{T-1} f(Z_t^0)\right|,
\end{eqnarray}
with probability at least $1-T\beta(q)/q$, and that the sequences $\{U_{2i}^0:i\ge 0\}$ and $\{U_{2i+1}^0:i\ge 0\}$ are i.i.d., with  $U_i^0=(Z_{iq}^0,Z_{iq+1}^0,\cdots,Z_{iq+q-1}^0)$. 

Recall that $\mathcal{I}_r=\{q\floor{T/q},q\floor{T/q} +1, \cdots,T-1\}$, we have
\begin{eqnarray*}
	\sup_{f\in \mathcal{F}}\left|\sum_{t=0}^{T-1} f(Z_t^0)\right|\le \sum_{j=0}^{q-1}\sup_{f\in \mathcal{F}} \left|\sum_{t=0}^{\floor{T/q}} f(Z_{tq+j}^0)\right|+\sup_{f\in \mathcal{F}}\left|\sum_{t\in \mathcal{I}_r} f(Z_t^0)\right|.
\end{eqnarray*}
Under the boundedness assumption on $F$, the second term on the right-hand-side is bounded from above by $Mq$. Without loss of generality, suppose $\floor{T/q}$ is an even number. The first term on the right-hand-side can be bounded from above by $\sum_{j=0}^{2q-1}\sup_{f\in \mathcal{F}} |\sum_{t=0}^{\floor{T/(2q)}} f(Z_{2tq+j}^0)|$. Therefore, 
\begin{eqnarray*}
	\sup_{f\in \mathcal{F}}\left|\sum_{t=0}^{T-1} f(Z_t^0)\right|\le \sum_{j=0}^{2q-1}\sup_{f\in \mathcal{F}} \left|\sum_{t=0}^{\floor{T/(2q)}} f(Z_{2tq+j}^0)\right|+Mq.
\end{eqnarray*}
This, together with \eqref{eqn:step1eq1}, yields that,
\begin{eqnarray}\label{eqn:step1eq2}
	\prob\left(\sup_{f\in \mathcal{F}}\left|\sum_{t=0}^{T-1} f(Z_t)\right|>2\tau q+Mq\right)\le \prob\left(\sum_{j=0}^{2q-1}\sup_{f\in \mathcal{F}} \left|\sum_{t=0}^{\floor{T/(2q)}} f(Z_{2tq+j}^0)\right|>2\tau q\right)+\frac{T\beta(q)}{q},
\end{eqnarray}
for any $\tau>0$. By Bonferroni's inequality, we obtain that,
\begin{eqnarray*}
	\prob\left(\sum_{j=0}^{2q-1}\sup_{f\in \mathcal{F}} \left|\sum_{t=0}^{\floor{T/(2q)}} f(Z_{2tq+j}^0)\right|>2\tau q\right)\le \sum_{j=0}^{2q-1} \prob\left(\sup_{f\in \mathcal{F}} \left|\sum_{t=0}^{\floor{T/(2q)}} f(Z_{2tq+j}^0)\right|>\tau \right),
\end{eqnarray*}
for any $\tau>0$. Since the process is stationary, we obtain that,
\begin{eqnarray*}
	\prob\left(\sum_{j=0}^{2q-1}\sup_{f\in \mathcal{F}} \left|\sum_{t=0}^{\floor{T/(2q)}} f(Z_{2tq+j}^0)\right|>2\tau q\right)\le 2q \prob\left(\sup_{f\in \mathcal{F}} \left|\sum_{t=0}^{\floor{T/(2q)}} f(Z_{2tq}^0)\right|>\tau \right).
\end{eqnarray*}
Combining this with \eqref{eqn:step1eq2} yields that,
\begin{align}\label{eqn:step1eq3}
	\begin{split}
		\prob\left(\sup_{f\in \mathcal{F}}\left|\sum_{t=0}^{T-1} f(Z_t)\right|>2\tau q+Mq\right)\le 2q \prob\left(\sup_{f\in \mathcal{F}} \left|\sum_{t=0}^{\floor{T/(2q)}} f(Z_{2tq}^0)\right|>\tau \right)+\frac{T\beta(q)}{q}.
	\end{split}
\end{align}
By construction, $\{Z_{2tq}^0:t\ge 0\}$ are i.i.d. This completes Step 1.

\smallskip
\noindent
\textbf{Step 2}. 
Next, we relate the empirical process $\sup_{f\in \mathcal{F}} |\sum_{t=0}^{\floor{T/(2q)}} f(Z_{2tq}^0)|$ to its expectation. Without loss of generality, suppose $T=kq$ for some integer $k>0$. Set the constants $\eta$ and $\delta$ in Lemma 1 of \cite{Adam2008} to 1, we have that, 
\begin{eqnarray*}
	\prob\left(\sup_{f\in \mathcal{F}} \left|\sum_{t=0}^{\floor{T/(2q)}} f(Z_{2tq}^0)\right|>2\Mean \sup_{f\in \mathcal{F}} \left|\sum_{t=0}^{\floor{T/(2q)}} f(Z_{2tq}^0)\right| +\tau \right)\\
	\le 4\exp\left(-\frac{\tau^2}{2T\sigma^2/q}\right)+\exp\left(-\frac{\tau}{CM}\right),
\end{eqnarray*}
for some constant $C>0$. Combining this with \eqref{eqn:step1eq3}, we obtain that, 
\vspace{-0.01in}
\begin{align}\label{eqn:step2}
	\begin{split}
		\prob\left(\sup_{f\in \mathcal{F}}\left|\sum_{t=0}^{T-1} f(Z_t)\right|>4q\Mean \sup_{f\in \mathcal{F}} \left|\sum_{t=0}^{\floor{T/(2q)}} f(Z_{2tq}^0)\right|+2\tau q+Mq\right)\\
		\le 8q\exp\left(-\frac{\tau^2}{2T\sigma^2/q}\right)+2q\exp\left(-\frac{\tau}{CM}\right)+\frac{T\beta(q)}{q},
	\end{split}
\end{align}
for any $\tau>0$. This completes Step 2.

\smallskip
\noindent
\textbf{Step 3}. It remains to bound $\Mean \sup_{f\in \mathcal{F}} |\sum_{t=0}^{\floor{T/(2q)}} f(Z_{2tq}^0)|$. By Corollary 5.1 of \cite{Cherno2014}, we have that, 
\begin{eqnarray*}
	\Mean \sup_{f\in \mathcal{F}} \left|\sum_{t=0}^{\floor{T/(2q)}} f(Z_{2tq}^0)\right|\preceq \sqrt{\frac{\nu \sigma^2T}{q} \log \left(\frac{AM}{\sigma}\right)}+\nu M \log \left(\frac{AM}{\sigma}\right).
\end{eqnarray*}
Combining this with \eqref{eqn:step2}, we obtain that, 
\begin{eqnarray*}
	\prob\left(\sup_{f\in \mathcal{F}}\left|\sum_{t=0}^{T-1} f(Z_t)\right|>c\sqrt{\nu q\sigma^2T \log \left(\frac{AM}{\sigma}\right)}+c\nu M \log \left(\frac{AM}{\sigma}\right)+c q\tau+Mq\right)\\
	\le Cq\exp\left(-\frac{\tau^2q}{CT\sigma^2}\right)+Cq\exp\left(-\frac{\tau}{CM}\right)+\frac{T\beta(q)}{q},
\end{eqnarray*}
for some constants $c,C>0$, and any $\tau>0,1\le q<T/2$. This completes the proof of Lemma \ref{lemma:aux}. 
\eop

\end{document}